\pdfoutput=1
\documentclass[preprint,12pt,review]{elsarticle}
\usepackage{dsfont}
\usepackage{amsmath}
\usepackage{amsthm}
\usepackage{amsfonts}
\usepackage{amssymb}
\usepackage{mathrsfs}
\usepackage{picture}
\usepackage{graphicx}
\usepackage{times}
\usepackage{balance}
\usepackage{booktabs}
\usepackage{multirow}
\usepackage{rotating}
\usepackage{comment}
\usepackage{indentfirst}
\usepackage{lineno}
\usepackage{color}
\usepackage{url}
\usepackage[implicit]{hyperref}
\usepackage{pdfpages}
\usepackage{indentfirst}

\newcommand{\ignore}[1]{}

\journal{Neurocomputing}

\begin{document}
\begin{frontmatter}


\title{Visual Tracking via Shallow and Deep Collaborative Model}

\author{Bohan Zhuang\corref{cor1}}
\cortext[cor1]{Corresponding author}
\cortext[cor1]{\emph{Email Address}: bohan.zhuang@adelaide.edu.au; \,\emph{Tel./Fax}: 0061431977667}



\author{Lijun Wang}

\author{Huchuan Lu}
\address{School of Information and Communication Engineering,
	Dalian University of Technology, Dalian, 116023, China}


\begin{abstract}
	

%
In this paper, we propose a robust tracking method based on the collaboration of a generative model and a discriminative classifier, where features are learned by shallow and deep architectures, respectively.
For the generative model, we introduce a block-based incremental learning scheme, in which a local binary mask is constructed to deal with occlusion.
The similarity degrees between the local patches and their corresponding subspace are integrated to formulate a more accurate global appearance model.
%
In the discriminative model, we exploit the advances of deep learning architectures to learn generic features which are robust to both background clutters and foreground appearance variations.
To this end, we first construct a discriminative training set from auxiliary video sequences.
A deep classification neural network is then trained offline on this training set.
%
Through online fine-tuning, both the hierarchical feature extractor and the classifier can be adapted to the appearance change of the target for effective online tracking.
The collaboration of these two models achieves a good balance in handling occlusion and target appearance change, which are two contradictory challenging factors in visual tracking.
Both quantitative and qualitative evaluations against several state-of-the-art algorithms on challenging image sequences demonstrate the accuracy and the robustness of the proposed tracker.
\end{abstract}

\end{frontmatter}

\section{Introduction}
\begin{figure}[tbsp]
\begin{tabular}{c@{}c@{}c}
\includegraphics[width=0.33\linewidth, height=0.19\linewidth]{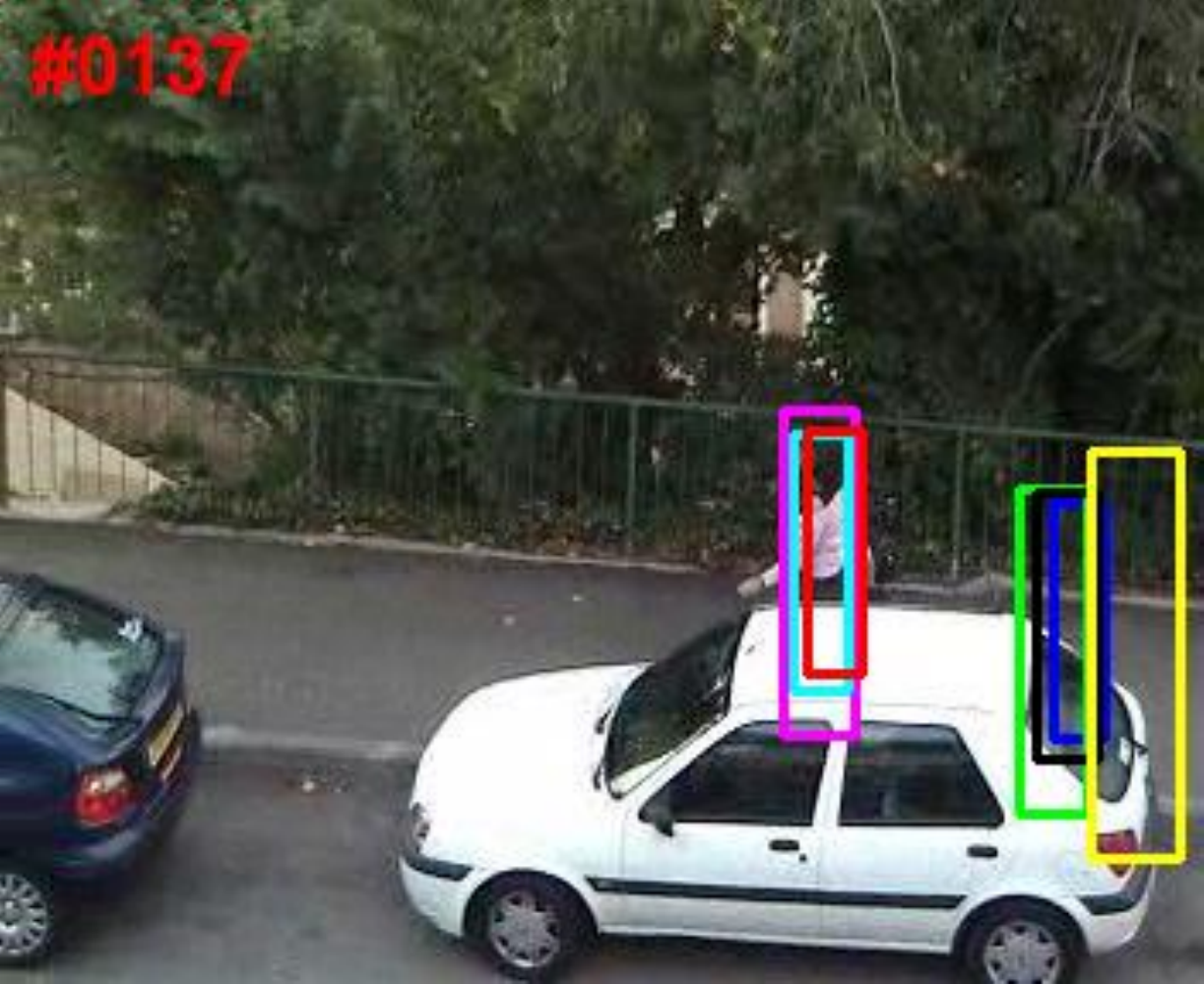}
&
\includegraphics[width=0.33\linewidth, height=0.19\linewidth]{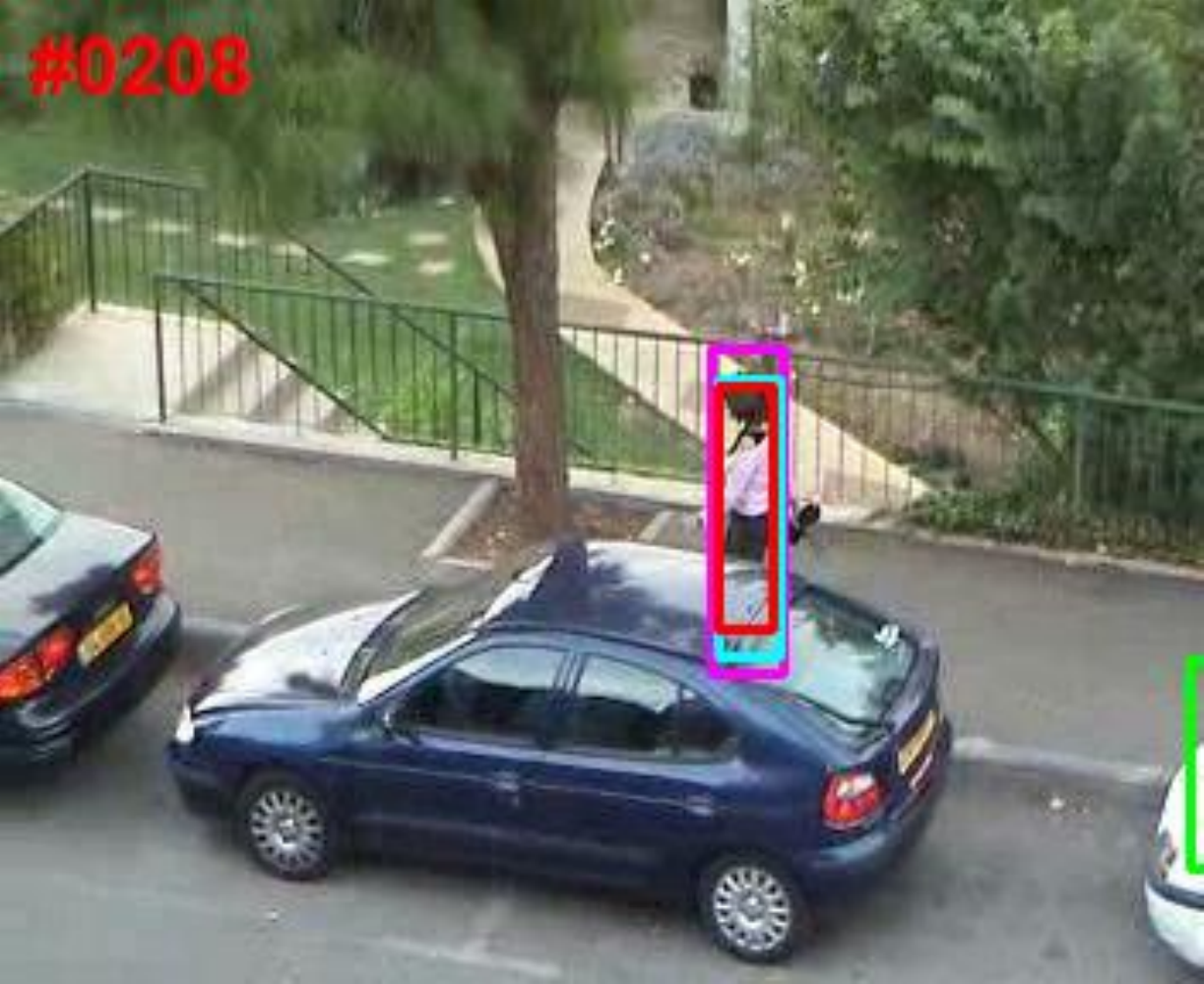}
&
\includegraphics[width=0.33\linewidth, height=0.19\linewidth]{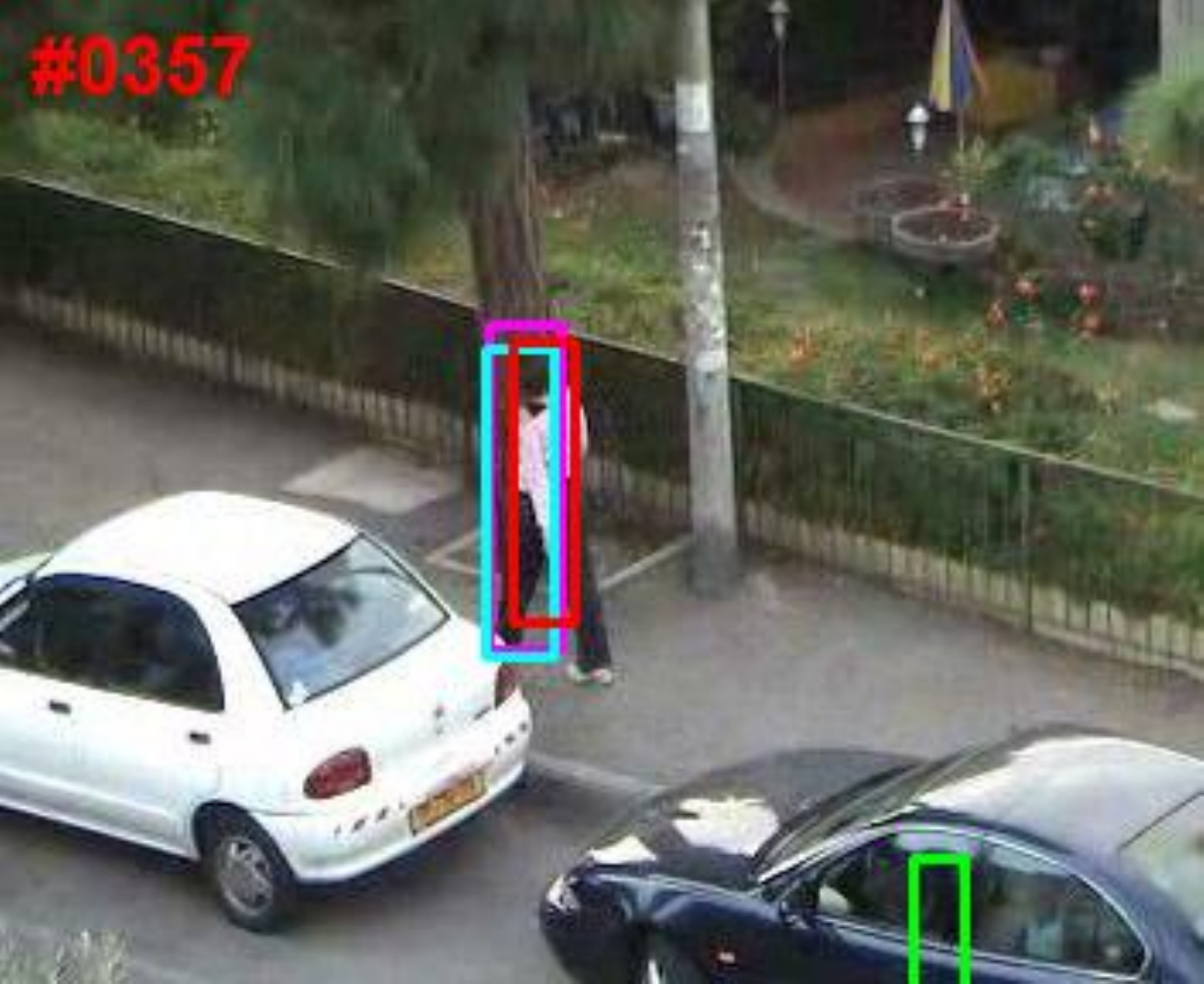}
\\
\end{tabular}

\begin{tabular}{c@{}c@{}c}
\includegraphics[width=0.33\linewidth, height=0.19\linewidth]{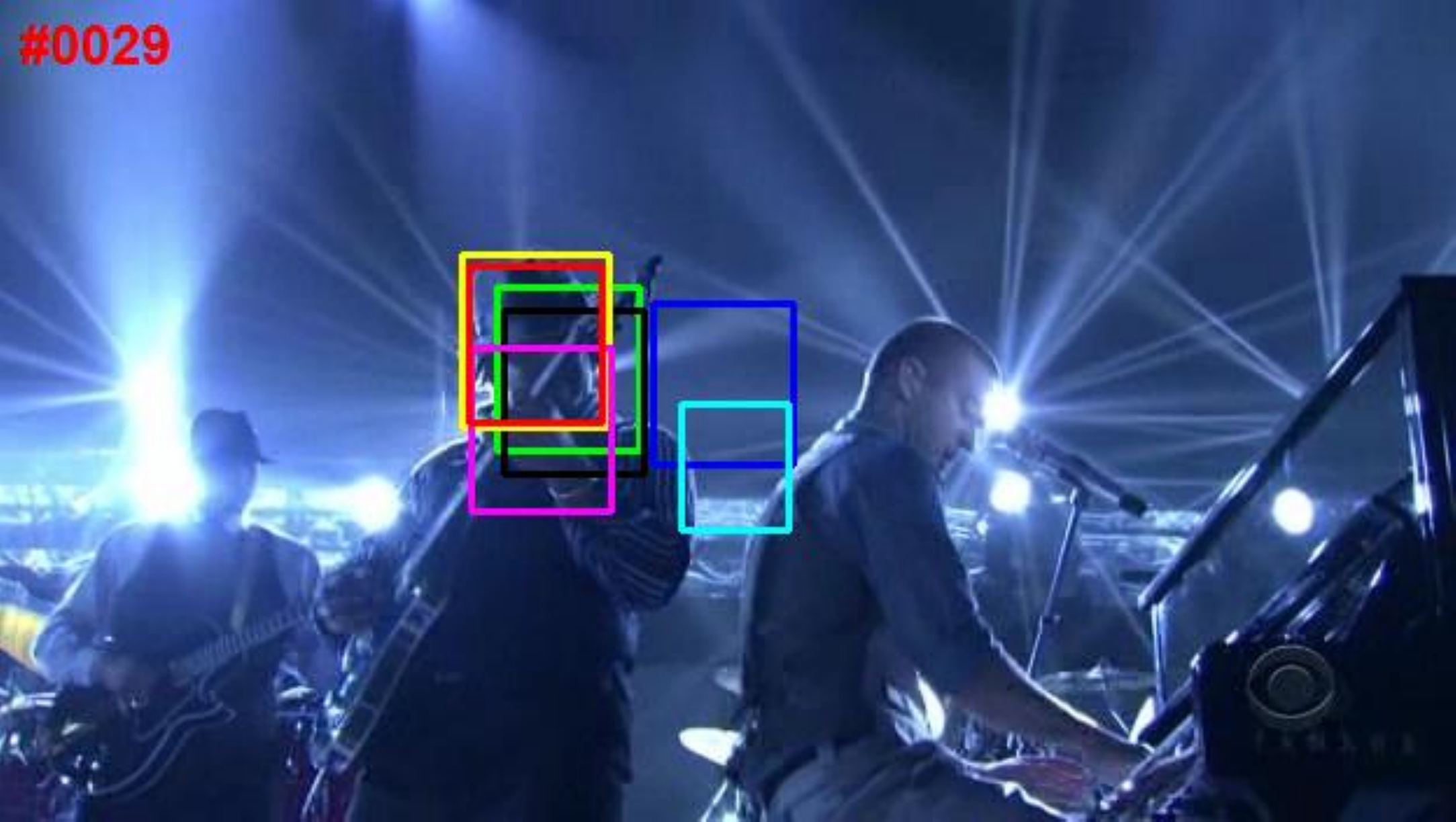}
&
\includegraphics[width=0.33\linewidth, height=0.19\linewidth]{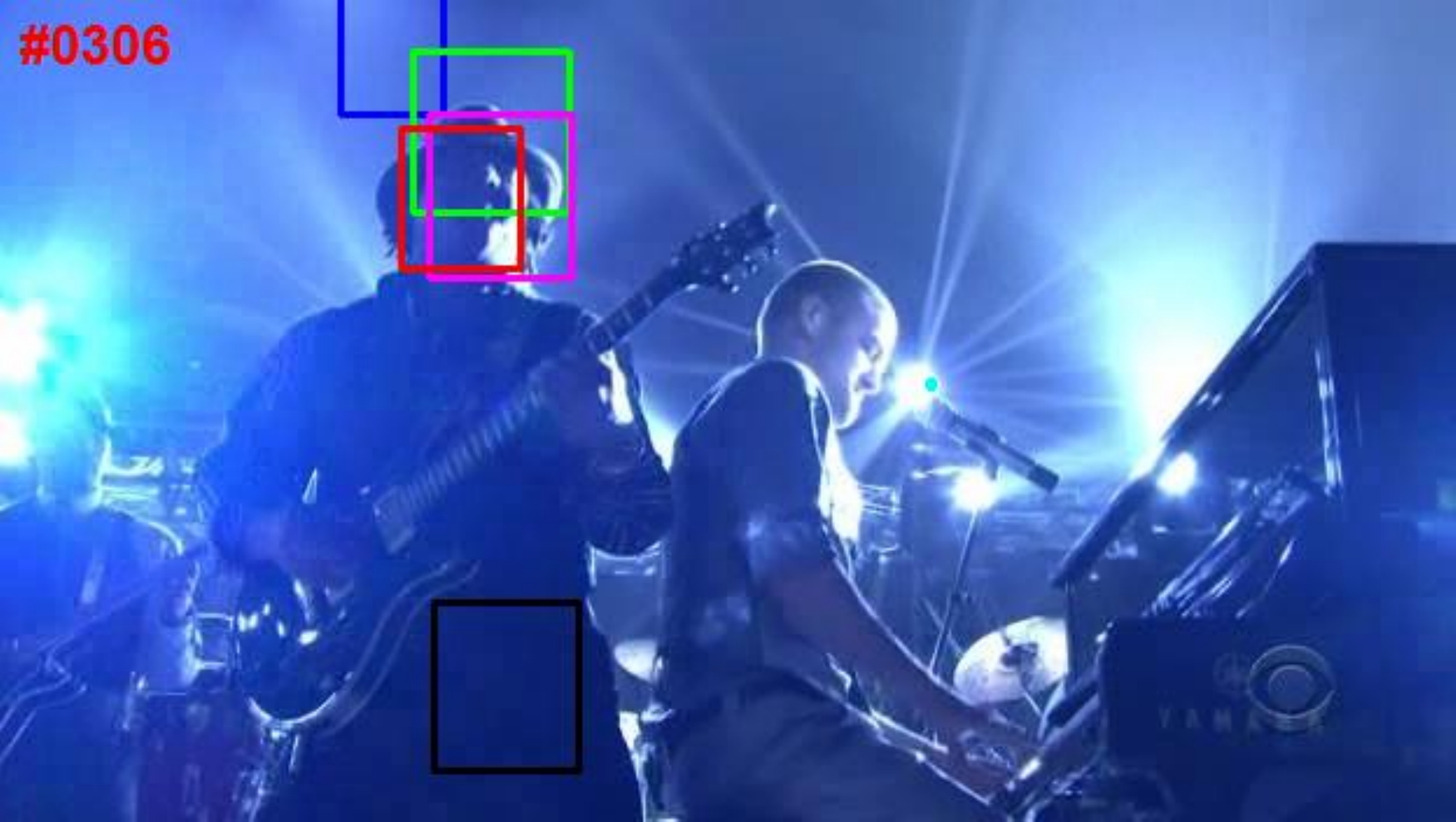}
&
\includegraphics[width=0.33\linewidth, height=0.19\linewidth]{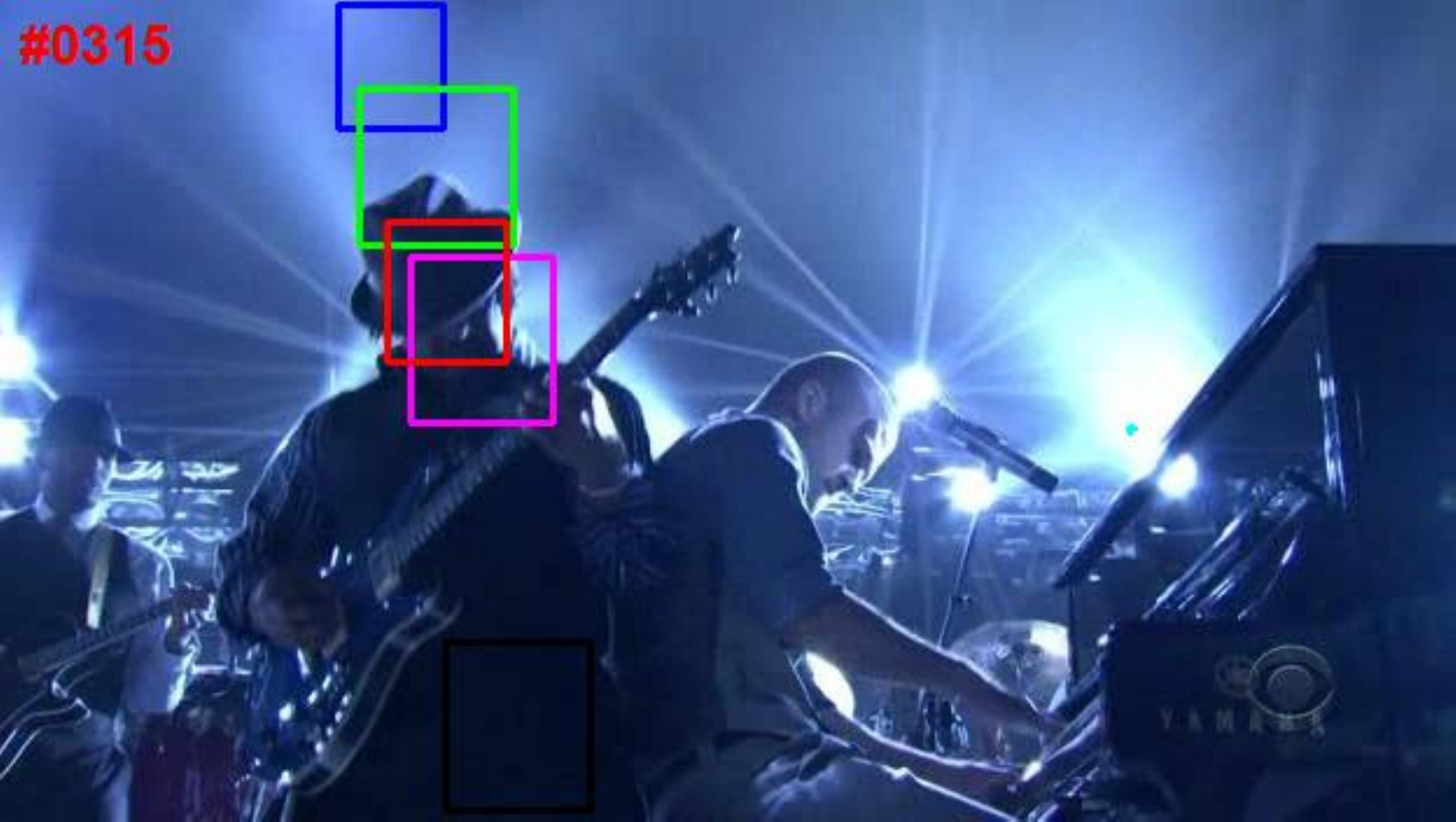}
\\
\end{tabular}

\begin{tabular}{c@{}c@{}c}
\includegraphics[width=0.33\linewidth, height=0.19\linewidth]{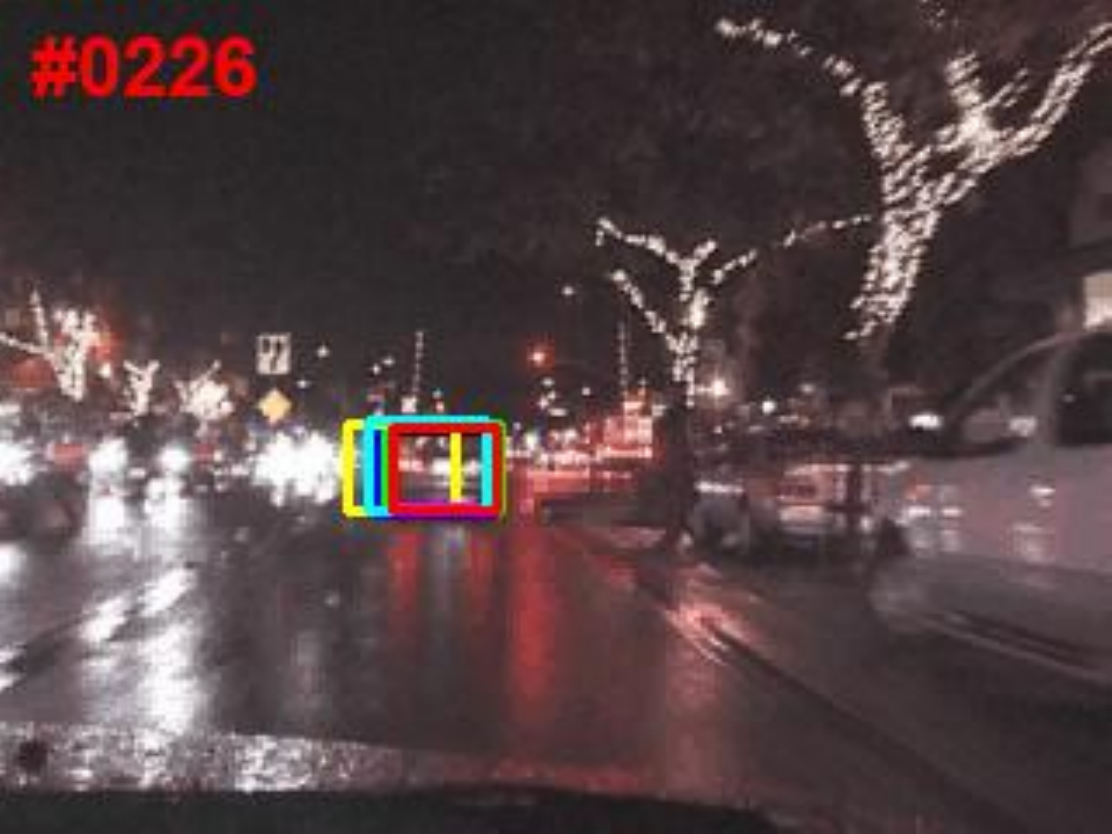}
&
\includegraphics[width=0.33\linewidth, height=0.19\linewidth]{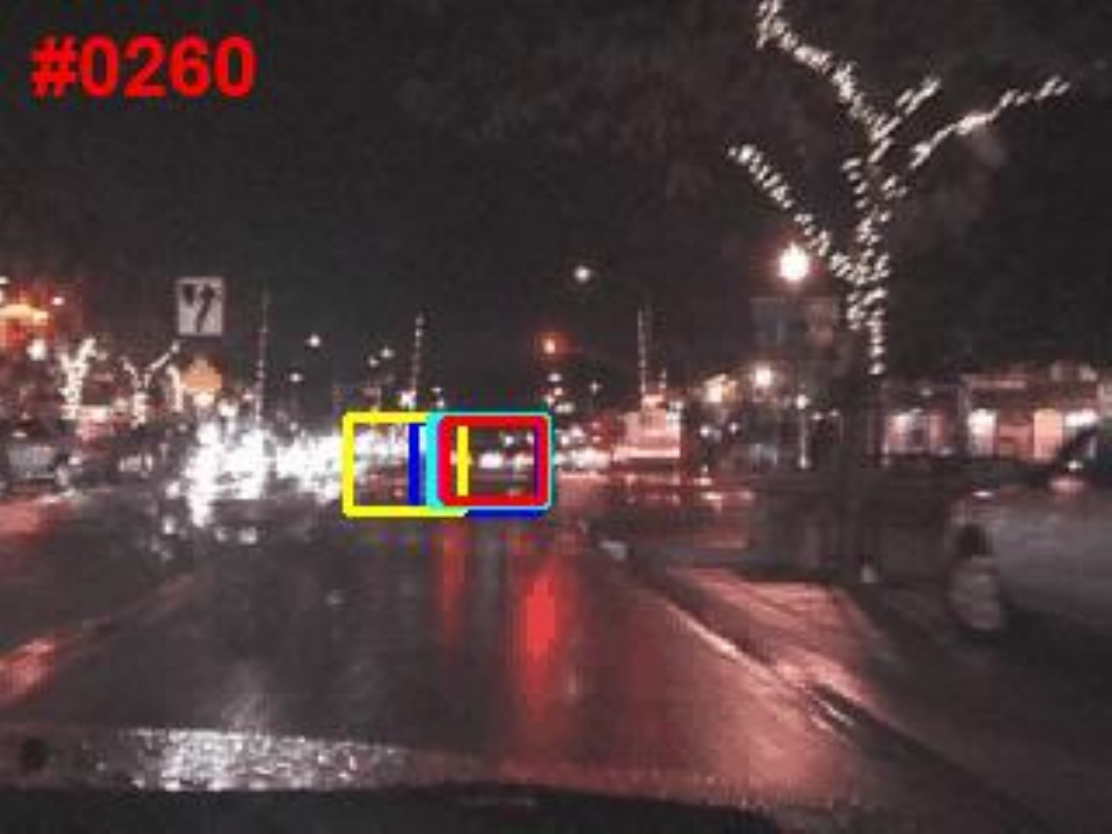}
&
\includegraphics[width=0.33\linewidth, height=0.19\linewidth]{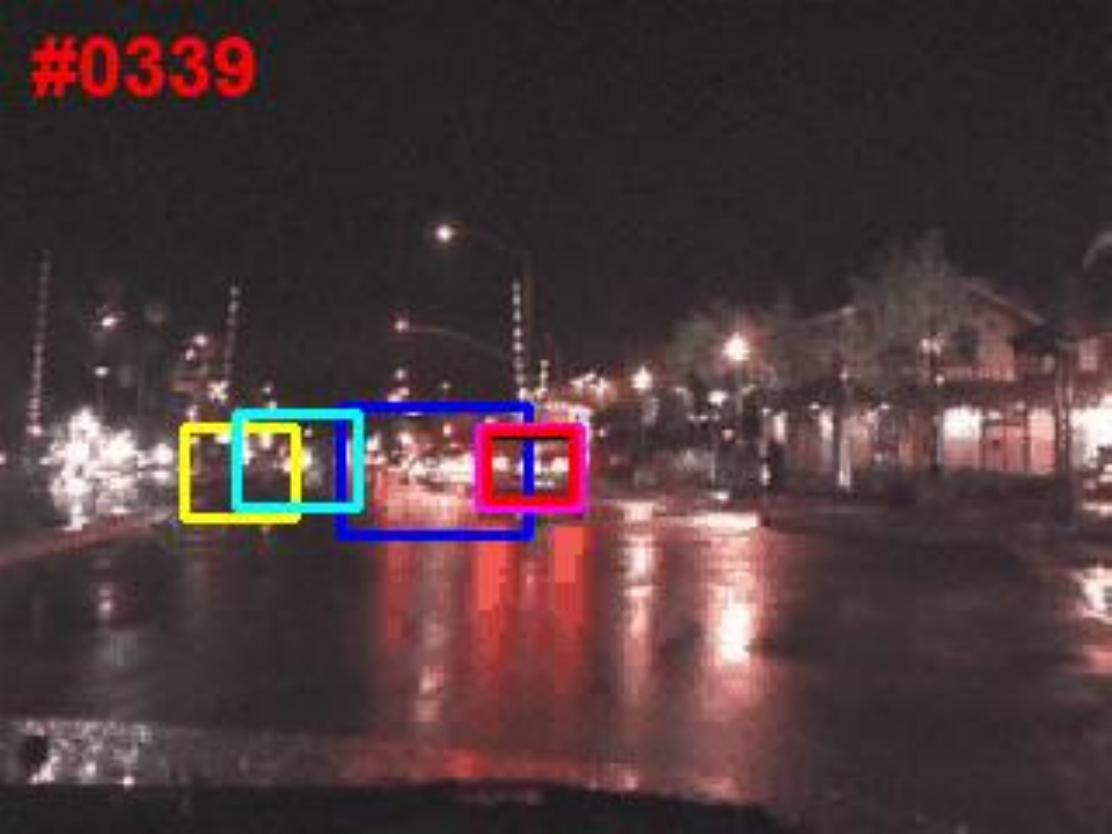}
\\
\end{tabular}

\begin{tabular}{c@{}c@{}c}
\includegraphics[width=0.33\linewidth, height=0.19\linewidth]{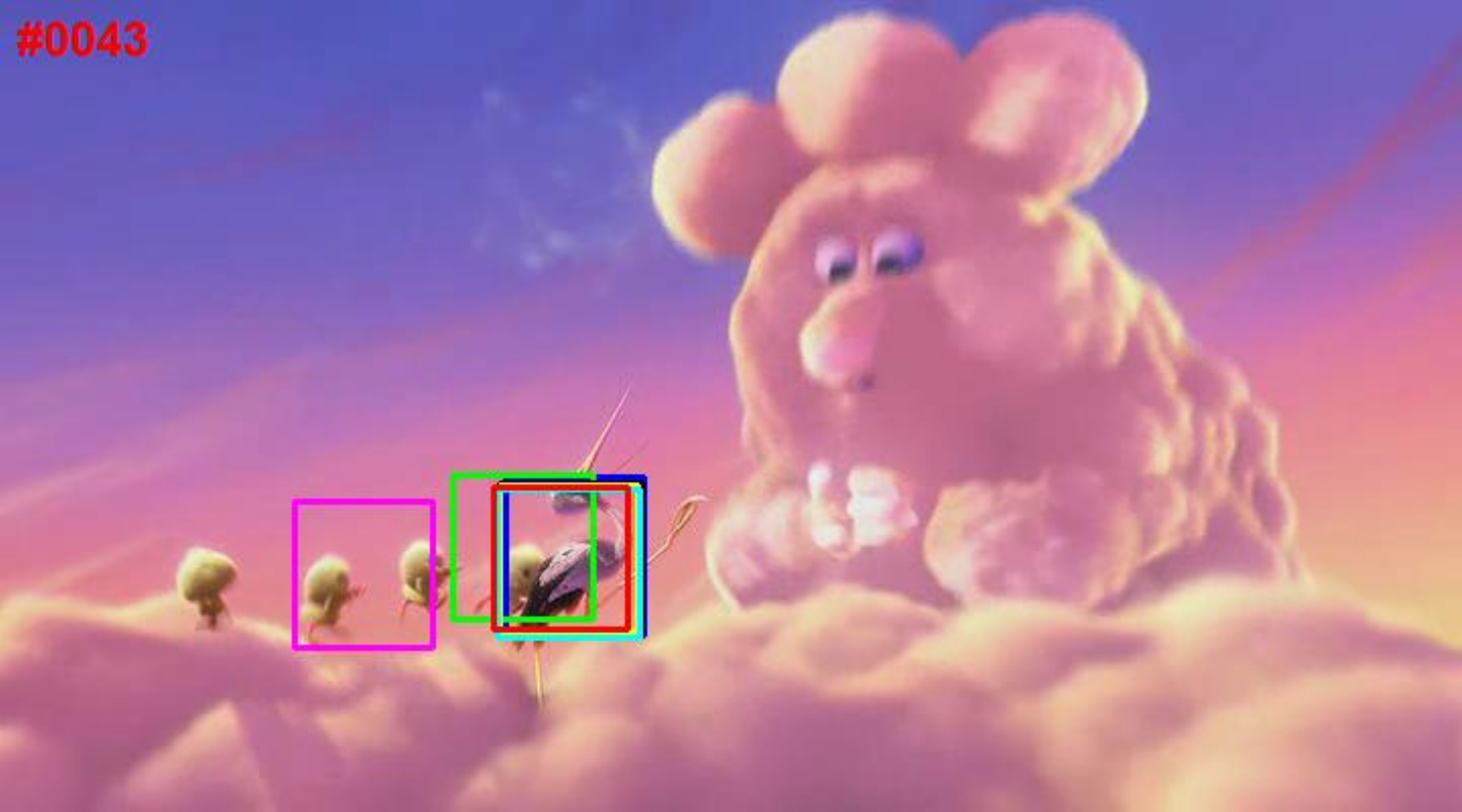}
&
\includegraphics[width=0.33\linewidth, height=0.19\linewidth]{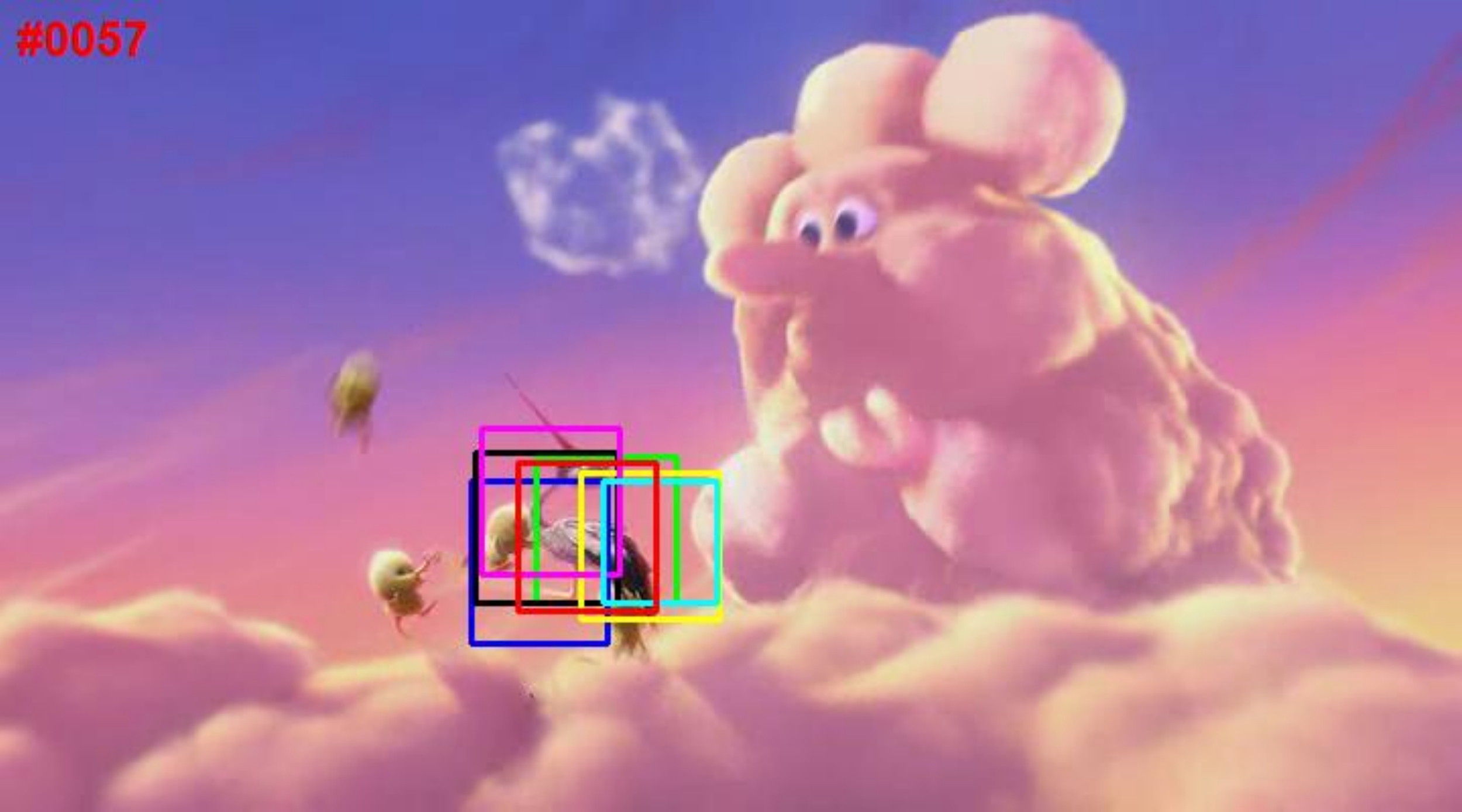}
&
\includegraphics[width=0.33\linewidth, height=0.19\linewidth]{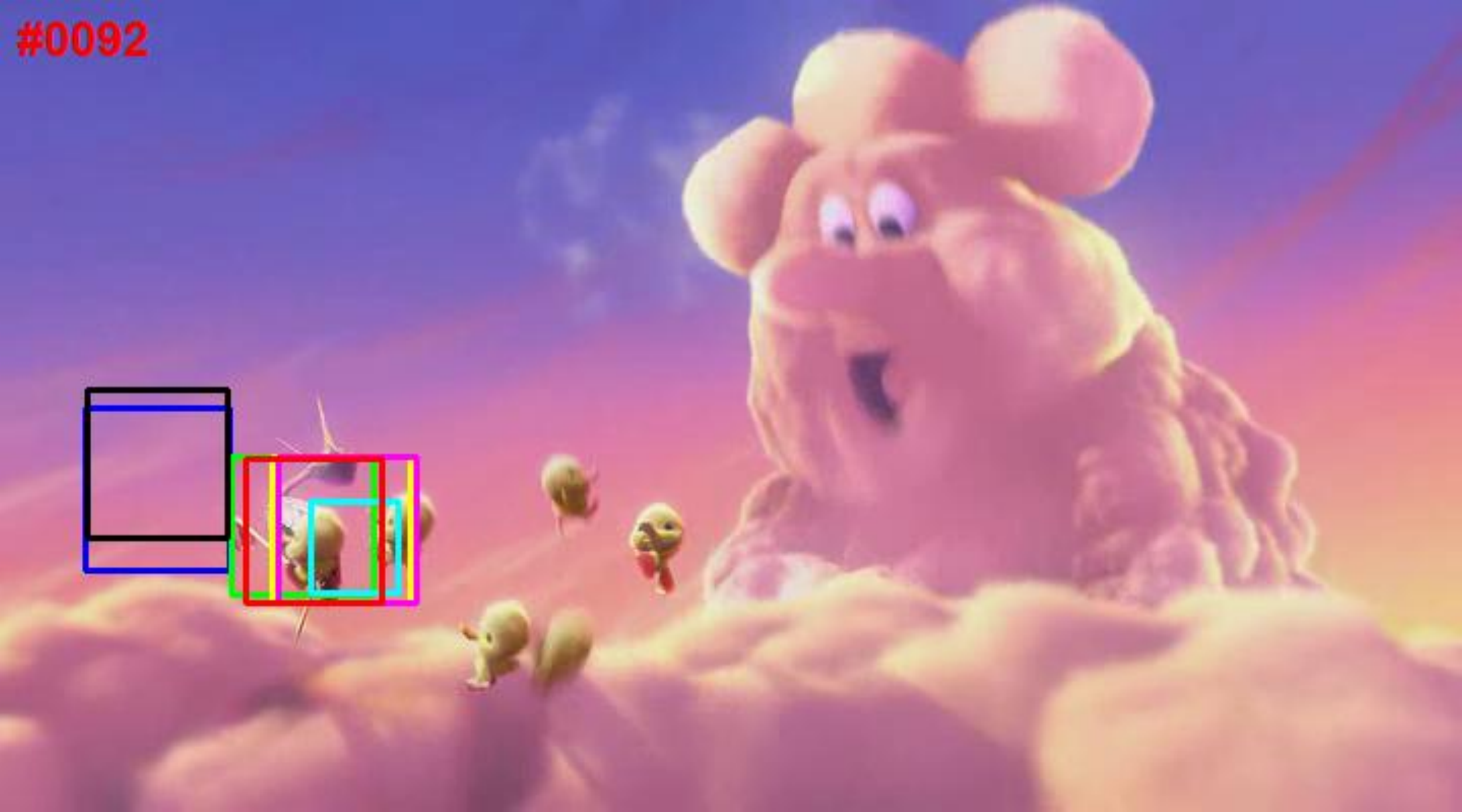}
\\
\end{tabular}

\begin{tabular}{c@{}c@{}c}
\includegraphics[width=0.33\linewidth, height=0.19\linewidth]{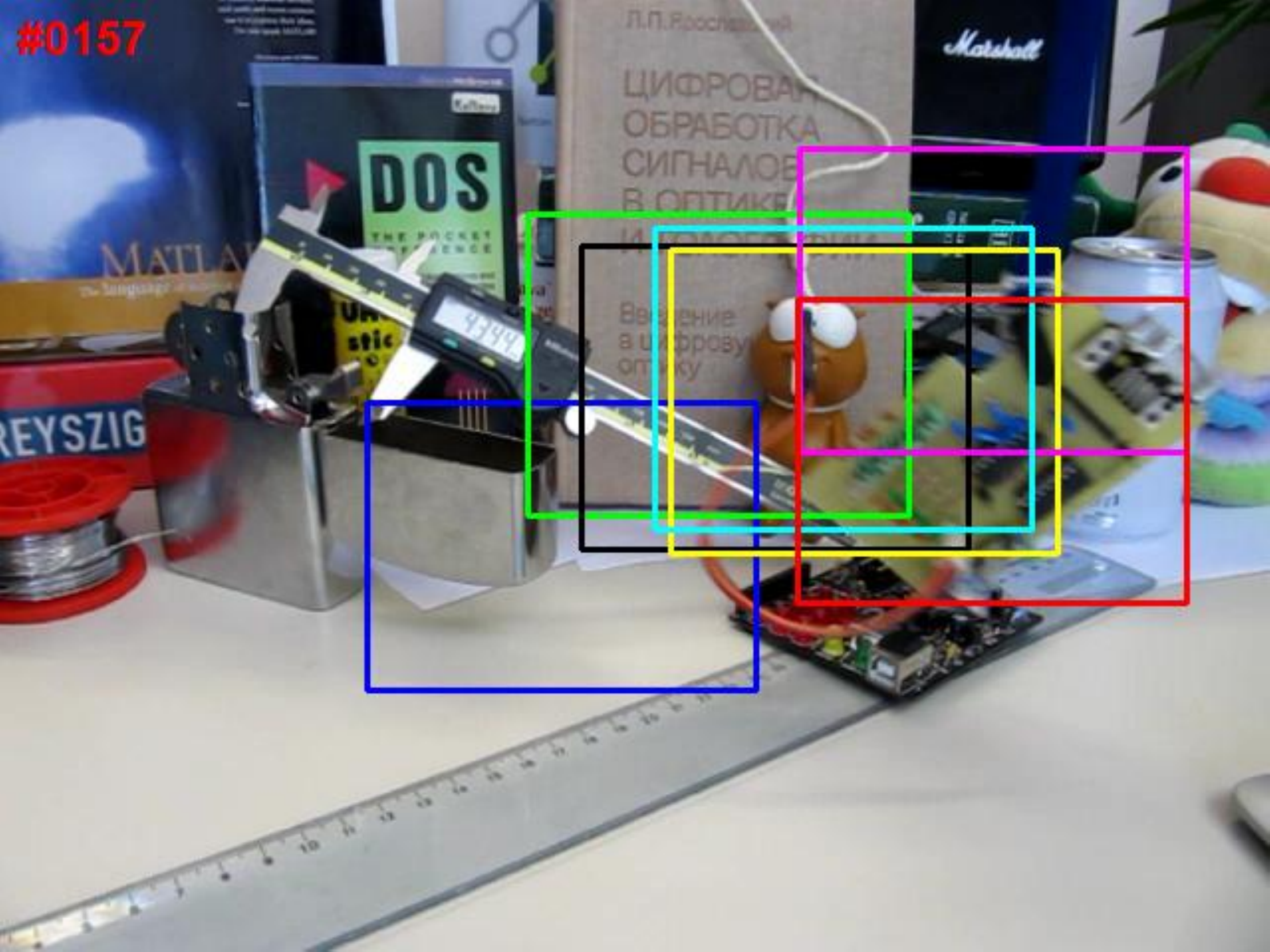}
&
\includegraphics[width=0.33\linewidth, height=0.19\linewidth]{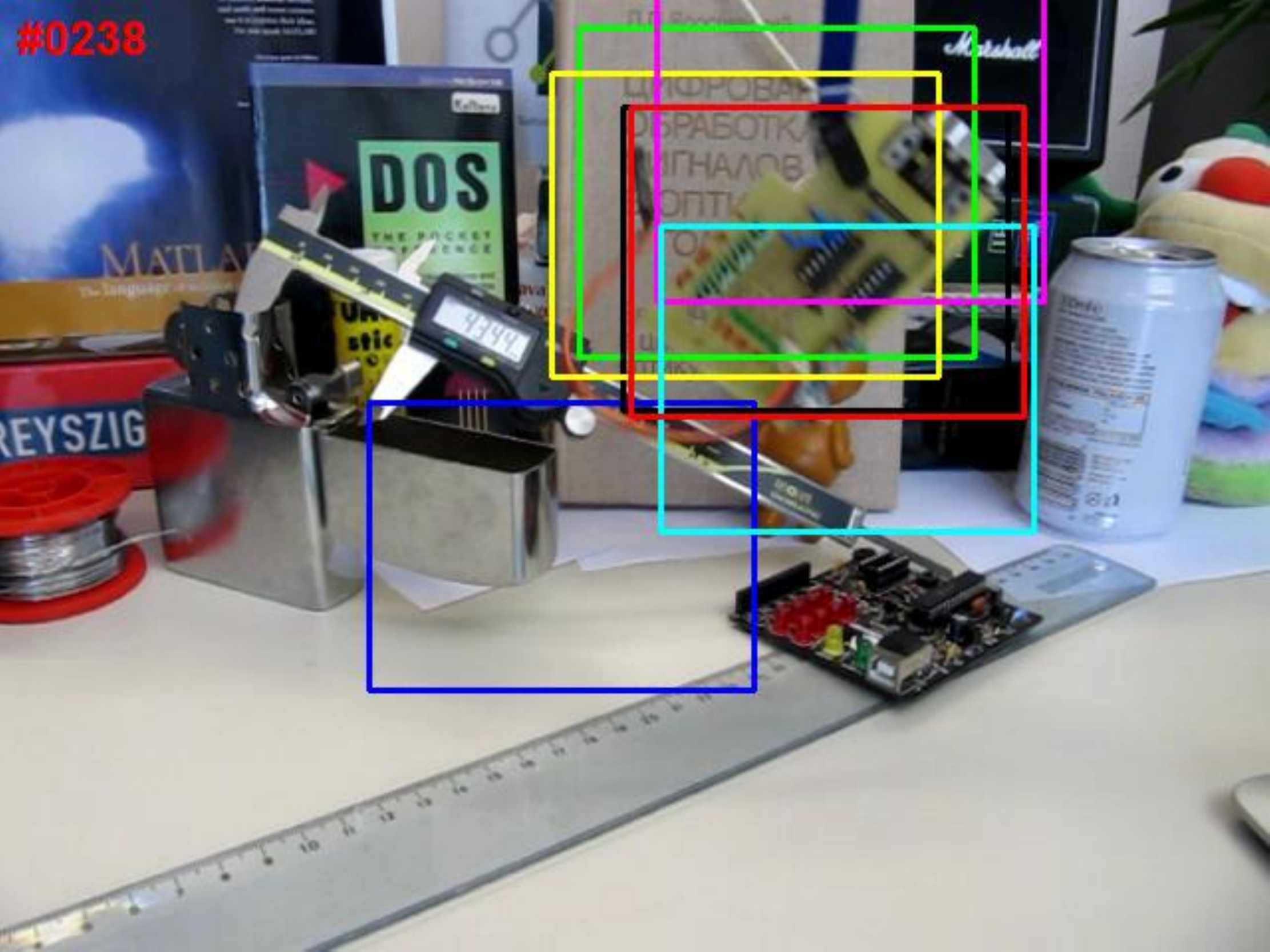}
&
\includegraphics[width=0.33\linewidth, height=0.19\linewidth]{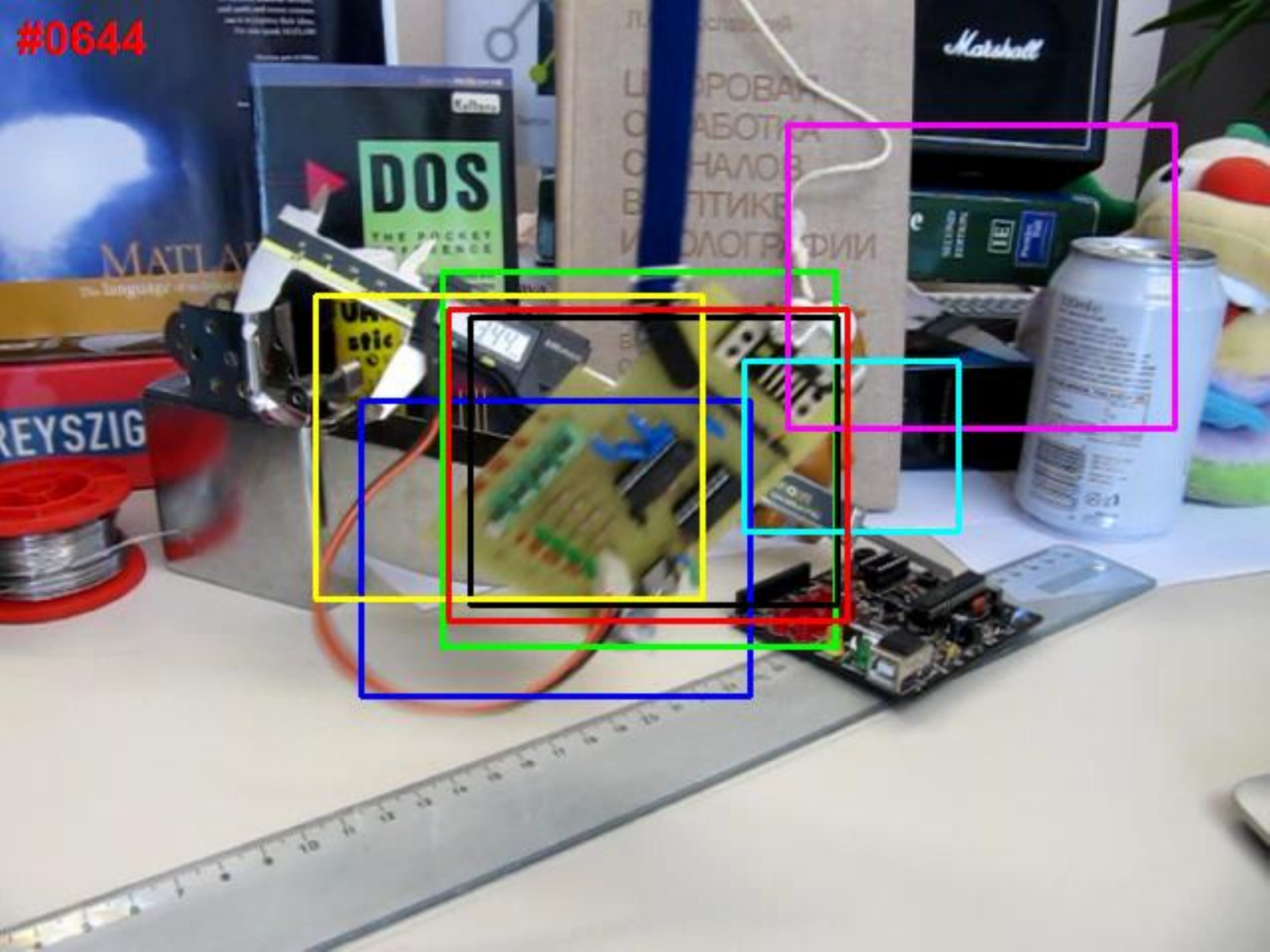}
\\
\end{tabular}

\caption{Challenges during tracking in real-world environments, including heavy occlusion (\emph{woman}), abrupt motion (\emph{shaking}), illumination change (\emph{carDark}), pose variation (\emph{bird}) and complex background (\emph{board}). We use blue, green, black, yellow, magenta, cyan and red rectangles to represent the tracking results of the IVT~\cite{ross2008incremental}, ASLA~\cite{jia2012visual}, OSPT~\cite{wang2013online}, CT~\cite{zhang2012real}, Struck~\cite{hare2011struck}, DLT~\cite{wang2013learning} and the proposed method, respectively.}
\label{fig:introductiongraphics}
\end{figure}

Visual tracking has long been playing a critical role in numerous vision applications such as military surveillance, human-computer interaction, activity recognition and behavior analysis.
 %
 %
The research in designing a robust tracker, which can well handle the challenging factors such as occlusion, illumination variation, rotation, motion blur, shape deformation and background clutter (See Figure~\ref{fig:introductiongraphics}), is very attractive.

Current trackers can mainly be categorized into either generative or discriminative approaches.
Generative trackers treat the tracking process as finding the candidate most similar to the target object.
These methods are mostly based on templates (like ~\cite{barnard2014robust, adam2006Frag, jia2012visual, meixue_L1_2009, zhuang2014visual, zhang2013gpu, suau2012real}, etc), subspace (like ~\cite{wang2013online, ross2008incremental}) or inference methods (like ~\cite{xie2013tracking, sabirin2012moving}).
%
%
%
Mei et al.~\cite{meixue_L1_2009} formulate tracking as a sparse coding problem where the target is sparsely represented by the target templates as well as the trivial ones.
And in~\cite{zhang2013gpu}, 3D articulated body pose tracking from multiple cameras are proposed to better deal with self-occlusions and pose variations.
In~\cite{sabirin2012moving}, Sabirin propose a novel spatio-temporal graphical models to simultaneously detect and track moving objects for video surveillance.

On the other hand, the discriminative trackers equate tracking as a binary classification problem in order to distinguish the target from the background (like those in ~\cite{avidan2007ensemble, kalal2010pn, babenko2009visual, grabner2008semi, chu2013tracking}).
In~\cite{chu2013tracking}, Chu et al. utilize projected gradient to facilitate multiple kernels in finding the best match during tracking under predefined constraints.
And further in~\cite{avidan2007ensemble}, a set of weak classifiers are combined into a strong one for robust visual tracking.
Kalal et al.~\cite{kalal2010pn} propose to train a binary classifier from labeled and unlabeled examples which are iteratively corrected by employing positive and negative constraints.
Furthermore, several trackers~\cite{zhong2012robust, yu2008online, liu2009robust, dinh2011co} are proposed to enjoy the advantages of both generative and discriminative models with good performance.
%
Motivated by this observation, we propose a novel collaborative model, where the generative model employs the shallow feature learning strategy to account for occlusion and the discriminative model adopts the deep feature learning strategy to effectively separate the foreground from the background.

In terms of feature learning, we use deep models to refer to networks that have more than one layer of hidden nodes, and use shallow models to refer to the rest feature learning methods with shallow architectures.
Some discriminative tracking methods focus on feature representation by utilizing shallow models.
The compressive tracker~\cite{zhang2012real} employs a sparse random measurement matrix to extract the data independent features for the appearance model and separates the object from the background using a naive Bayes classifier.
In~\cite{grabner2006real}, Grabner et al. propose an online AdaBoost feature selection algorithm to adapt the classifier to the appearance change of the target.
Collins et al.~\cite{collins2005online} use a feature ranking mechanism to adaptively select the top-ranked discriminative features from multiple feature spaces for tracking.
In~\cite{grabner2007learning}, keypoint descriptors in the region of the interested object are learned online with background information being considered in the meantime.
%
%
%
However, due to the difficulty in representing complex functions using limited samples and the restricted capability in generalizing complicated classification problems, the performance of shallow models in tracking scenarios is not satisfactory.

On the other hand, deep learning has been successfully introduced to several computer vision applications, such as image classification~\cite{krizhevsky2012imagenet}, face recognition~\cite{sun2013hybrid} and object-class segmentation~\cite{schulz2012learning}.
Its aim is to replace the hand-crafted features with the high-level and robust features learned from raw pixel values ~\cite{zhuang2016fast,  hinton2006reducing, le2013building, coates2012emergence, marc2007sparse}.
The deep feature learning strategy demonstrates a strong capability to extract essential characteristics from massive  auxiliary data by layer-wisely training a deep nonlinear network.
The rich invariant features learned in this way can be further employed in classification and prediction problems, and are empirically shown by our experiments to improve the accuracy and robustness of visual trackers.
%



The main contributions of our work are summarized as follows:
\begin{itemize}

\item { We combine the shallow feature learning and deep feature learning strategy in one collaborative tracker for the first time.
The shallow feature learning method, acting as a generative model, mainly accounts for occlusion, whereas the deep classification neural network, serving as a discriminative model, covers the appearance change of the target.
Thus the integration of these two models enables our tracker to well handle occlusion and appearance change at the same time.
}

\item{We construct a training set containing both positive and negative image patches sampled from natural images.
    The deep network is first trained offline on this training set and is further fine-tuned online in the tracking process.
    Because of the large discriminative training set and the effective deep architecture, the offline training makes the learned generic features more robust against appearance variations and background clutter, while the online fine-tuning adapts the generic features to the specific target.
}
\item{We propose a block-based local PCA model with online update scheme to alleviate the negative effects caused by the occluded target regions.}
\end{itemize}

\section{Related work and context}
%
Tracking methods which employ online subspace learning (~\cite{ ross2008incremental, wang2010incremental, wang2007object, hu2011incremental}) have demonstrated strong robustness in dealing with in-plane rotation, illumination variation and pose change.
Ross et al.~\cite{ross2008incremental} propose an incremental learning algorithm for tracking, where drifting problem is addressed to some extent by online learning a subspace representation of the target appearance.
%
%
However, it cannot deal with partial occlusion very well due to the employed global appearance model.
%
On the other hand, part-based trackers~\cite{adam2006Frag, liu2011robust} are more flexible in detecting and handling noise caused by occlusion and non-rigid appearance change.
Based on these observations, we propose a part-based PCA representation scheme as the shallow online feature learning model, to handle appearance change as well as occlusion.

%
Recently, there has been much interest in unsupervised learning of hierarchical generative models such as deep belief networks.
In~\cite{hinton2006fast}, Hinton et al. derive an efficient training approach for deep belief networks (DBN) which consists of a greedy layer-wise pre-training stage and a fine-tuning stage using a contrastive version of the wake-sleep algorithm.
Following~\cite{hinton2006fast}, several improvements on DBN have been successfully applied in numerous applications~\cite{lee2009convolutional,lee2009unsupervised,lee2007sparse}.
Since the features of an object could change significantly due to various factors, a good tracker desires the generic features which can capture the essential characteristic of the foreground object.
%
Therefore, we employ a DBN, which is trained offline using our self-built auxiliary training set, to handle drastic appearance change of the tracked object.
In~\cite{wang2013learning}, Wang et al. train a denoising autoencoder offline to learn generic image features and add a classification layer on top which is trained online for tracking.
This work introduces deep learning into visual tracking applications for the first time and achieves good tracking results with low computational cost.
%
%
The stacked denoising autoencoder is trained to recover the original images from their corrupted versions which are obtained by adding small Gaussian noise to the original images.
However, the Gaussian noise assumption does not hold for visual tracking where the noise is mainly caused by occlusion and appearance change.
%
%
In contrast, our work explicitly detects occlusion using the proposed part-based PCA model and excludes the negative effects of occlusion block-wisely by employing an occlusion mask, thus facilitating a more accurate appearance model.
Furthermore, we add a classification layer on top of the deep belief network and train this discriminative model together offline rather than just simply train a generative deep model to extract hierarchical features.
Also, we construct a large training set composed by both positive foreground samples and negative background samples to train the framework offline, which can motivate our deep classification network to learn generic and discriminative features that can separate the foreground from the background.

\section{Bayesian interference framework}
In this paper, object tracking is carried out within the Bayesian inference framework.
%
Given the observation set of the target ${\bf{Y}_{1:t}} = \{ {\bf{y}_1},{\bf{y}_2},...,{\bf{y}_t}\}$ up to the $t$-th frame, object tracking is formulated as finding the most likely state of the target at time $t$ by using the maximum a posteriori estimation,
\begin{equation}
{{\bf{x}}_t} = \mathop {\arg \max }\limits_{{\bf{x}}_t^i} p({\bf{x}}_t^i|{{\bf{Y}}_{1:t}})
\label{eq:MAP}
\end{equation}
where $\mathbf{x}_t^i$ represents the state of the $i$-th sample in the $t$-th frame.
The posterior probability $p({\bf{x}_t}|{\bf{Y}_{1:t}})$ can be inferred by the Bayesian theory as follows

\begin{equation}
p({{\bf{x}}_t}|{{\bf{Y}}_t}) \propto p({{\bf{y}}_t}|{{\bf{x}}_t})\int {p({{\bf{x}}_t}} |{{\bf{x}}_{t - 1}})p({{\bf{x}}_{t - 1}}|{{\bf{Y}}_{t - 1}})d{{\bf{x}}_{t - 1}}
\end{equation}
where ${p({\bf{x}_t}} |{\bf{x}_{t - 1}})$ denotes the motion model and $p({\bf{y}_t}|{\bf{x}_t})$ denotes the observation model.
The motion model ${p({\bf{x}_t}} |{\bf{x}_{t - 1}})$ describes the temporal relationship of the target states between two consecutive frames using an affine transformation and is approximated by a Gaussian distribution as $p({\bf{x}_t}|{\bf{x}_{t - 1}}) = N({\bf{x}_t};{\bf{x}_{t - 1}},\Psi )$, where $\Psi$ is a diagonal covariance matrix whose elements are the variances of the affine parameters.
The appearance model $p({\bf{y}_t}|{\bf{x}_t})$ is defined as the similarity between the candidates and the target observations.
%

\begin{figure*}[tbp]
\centering
\includegraphics[width=1.00\linewidth,height=0.25\linewidth]{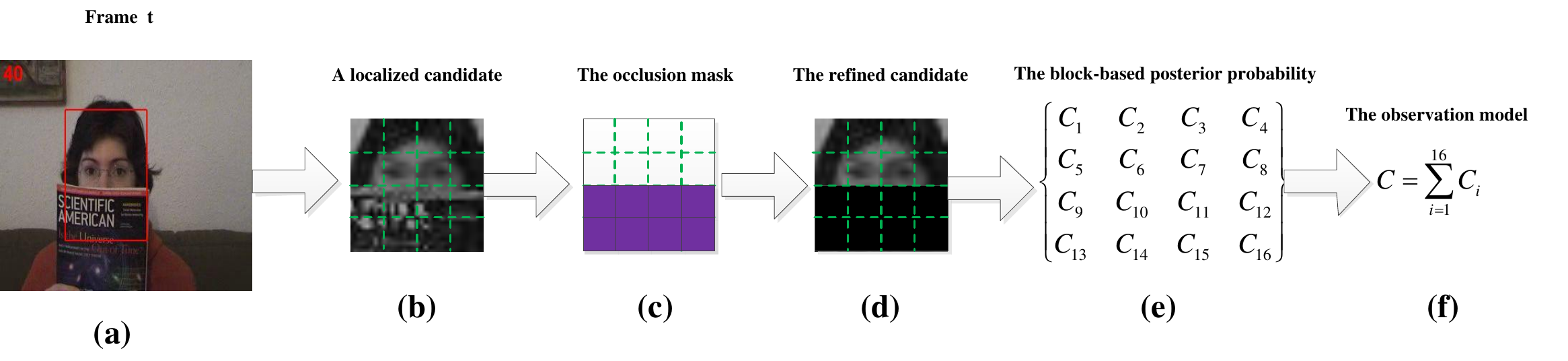}
\vspace{-8mm}
\caption{The local generative model. (a) A frame at time t with a candidate is picked as example. (b) The localized candidate. We divide the candidate into $4 \times 4$ rectangle blocks. (c) The block-wisely occlusion mask obtained from the previous frame. (d) The refined candidate by applying the occlusion mask to the original candidate. (e) The block-based posterior probability. (f) The integral posterior probability.}

\label{fig:generativemodel}
\end{figure*}

\section{The proposed algorithm}
\subsection{The shallow generative model}
\subsubsection{Local model}

In practical occasion, the tracking algorithms based on a global subspace model are able to handle many challenging factors (such as illumination change, scale change, motion blur, etc.).
Let $\bf{y}$ denotes an observation vector, $\bf{u}$ is the center vector of the subspace, then ${{\bf{y}}^*} = {\bf{y}} - {\bf{u}}$ indicates the corresponding observation normalized by subtracting the mean of the subspace.
The PCA model can be represented as
\begin{equation}
{\bf{y}^*} = {\bf{Uz}} + {\bf{e}}
\end{equation}
where $\bf{U}$ is the subspace composed by column basis vectors, $\bf{z}$ indicates the corresponding coefficient vector and $\bf{e}$ is the noise term.
The PCA model assumes that the noise is subject to Gaussian distribution with small variances.
Therefore, $\bf{z}$ can be approximated by $\hat{\bf{z}}  = {{\bf{U}}^ \top }{{\bf{y}}^*}$, and the reconstruction error can be computed as $||{{\bf{y}}^*} - {\bf{U}}{{\bf{U}}^ \top }{{\bf{y}}^*}||_2^2$.
Then, the global model can be generated from the subspace governed by
\begin{equation}
p({{\bf{y}}}|{{\bf{x}}}) \propto \exp ( - {\left\| {{{\bf{y}}^*} - {\bf{U}}{{\bf{U}}^ \top }{{\bf{y}}^*}} \right\|^2})
\label{eq:IVT}
\end{equation}
where ${{\bf{y}}}$ is the the observation image patch corresponding to the state ${{\bf{x}}}$.
Eq.~\ref{eq:IVT} can appropriately describe the negative exponential distance from the corresponding image patch to the subspace, spanned by $\bf{U}$, formed by previous target particles.
However, in practical visual tracking problem, this model is not able to handle partial occlusion well as Gaussian noise with small variances cannot model the outliers and fails to take the contiguous spatial distribution of occlusion into account.
It is shown that analyzing the local image patches can promote the recognition performance against partial occlusion whose spatial support is unknown but contiguous like~\cite{martinez2002recognizing, ahonen2006face}.
Inspired by this, we construct a block-based local model with incremental learning.
%
Given an observation image patch, it is first divided into sub-blocks of $n \times n$ pixels ($4 \times 4$ blocks in our experiments).
%
The generative score ${{C_{i}}} $ of the $i$-th block of the given candidate indicates the similarity between the $i$-th block and its corresponding subspace formed by previous target particles, and is defined as:
\begin{equation}
{C_i} = \exp ( - {\left\| {({\mathbf{\pi}_i} - {{\bf{u}}_i}) - {{\bf{U}}_i}{\bf{U}}_i^ \top ({\bf{\pi}}_i - {{
\bf{u}}_i})} \right\|^2})
\end{equation}
where ${\bf{\pi}}_i$ denotes the $i$-th block of the observation and its subspace as well as its corresponding center are denoted as ${{\bf{U}}_{i}}$ and ${{\bf{u}}_{i}}$ , respectively.
The global generative score of the observation is then defined as the sum of the generative scores of its blocks
\begin{equation}\label{eq:model}
  G = \sum \limits_{i=1}^{n^2} C_i
\end{equation}
It is worth noticing that when partial occlusion occurs, the generative scores corresponding to the occluded parts will be suppressed to be small, i.e. the occlusion bears little similarity to the target subspace and thus contributing little to the global generative score.
%
Thus, our generative model defined in Eq.~\ref{eq:model} can render a more accurate posterior probability by alleviating the influence of the partial occlusion.
\subsubsection{The refined local model}
To further avoid the negative impact of the occluded appearance, the proposed method introduces an occlusion mask $\mathbf{M}=[M_i]_{n^2 \times 1}$, where $M_i$ indicates whether the $i$-th block has been occluded based on its generative score:
%
%
\begin{equation}   \label{eq:mask}
{M_{i}} = \left\{ \begin{array}{l}
0{\kern 1pt} {\kern 1pt} {\kern 1pt} {\kern 1pt} {\kern 1pt} {\kern 1pt} if{\kern 1pt} {\kern 1pt} {\kern 1pt} {C_{i}} \le \delta \\
1{\kern 1pt} {\kern 1pt} {\kern 1pt} {\kern 1pt} {\kern 1pt} {\kern 1pt} if{\kern 1pt} {\kern 1pt} {\kern 1pt} {\kern 1pt} {C_{i}} > \delta
\end{array} \right.
\end{equation}
where $\delta$ is a fixed threshold.
Then the final generative score for the $i$-th block $\hat{C}_{i}$ is computed as:
\begin{equation}  \label{eq:refinedmask}
\begin{array}{l}
{\hat{C}_i} = {M_i}\exp ( - {\left\| {({\bf{\pi}}_i - {{\bf{u}}_i}) - {{\bf{U}}_i}{\bf{U}}_i^{\top}({\bf{\pi}}_i - {{\bf{u}}_i})} \right\|^2})\\
i = 1,2,\ldots,{n^2}.
\end{array}
\end{equation}
Taking occlusion into consideration, the final global generative score of the observation is defined as
\begin{equation} \label{eq:refinedmodel}
  \hat{G} = \sum \limits_{i=1}^{n^2} \hat{C}_i
\end{equation}
An example in Figure~\ref{fig:generativemodel} illustrates the benefit of the occlusion mask.
By adding the occlusion mask, we can observe that the observation likelihoods of the occluded blocks of the candidate are set to be zeros, which means we completely remove the negative influence from the partial occlusion.
Hence, only the parts bear enough similarity to the target subspace account for the generative score, which assigns more weights on the foreground parts and leads a more accurate and more convincing generative model.
\ignore{
Due to temporal contiguity of video clips, the occlusion masks between two consecutive frames remain almost the same.
Thus, the occlusion mask formed in frame $t$ is then used in frame ${t+1}$ to calculate $C_{i}$.
}
\subsubsection{Online update}
An effective online learning scheme is an essential component for our generative local model, as the appearance of the tracked target and its surrounding background may change during the tracking process.
The subspace is incrementally learned by implementing the IPCA method~\cite{ross2008incremental} every 5 frames, in which 16 eigenvectors are maintained in each frame.
In order to detect the severe occlusion, we define an occlusion rate $o$ for each frame according to its corresponding occlusion mask $\mathbf{M}$:
\begin{equation}
  o=\sum \limits_{i=1}^{n^2} M_i \end{equation}
When sever occlusion happens, the occlusion rate $o$ will fall below a predefined threshold $\chi $, and we give up local model updating.
%
%
%
%

\section{The deep discriminative model}
\subsection{Offline training process}
\subsubsection{Training set preparation}
%
\begin{figure}[tbp]
\centering
\resizebox{\textwidth}{!}{
\includegraphics{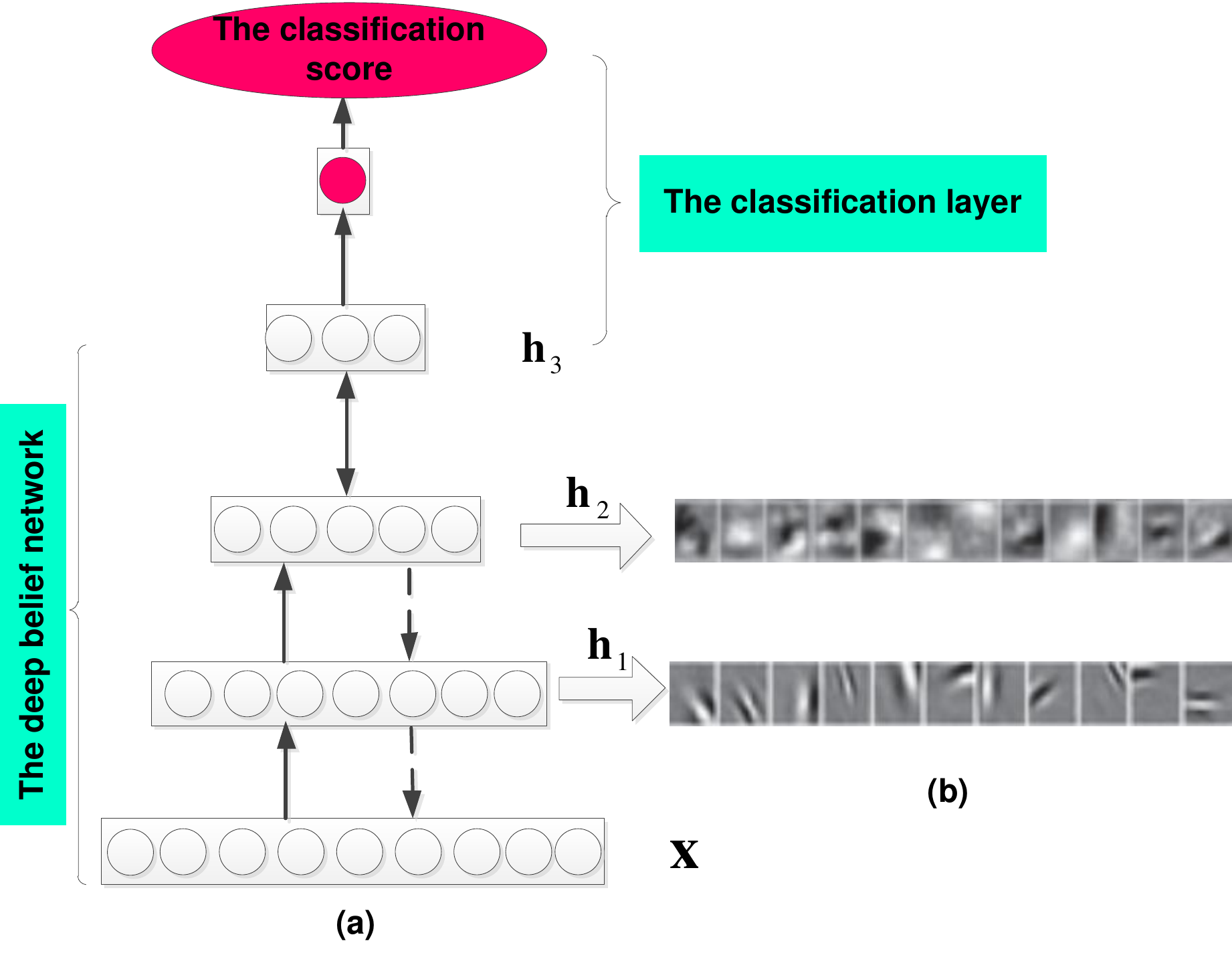}}
\vspace{-8mm}
\caption{This figure illustrates the structure of our deep discriminative model and some layer bases. (a) The deep belief network and the classifier. x is the input layer, ${{\bf{h}}_1}$, ${{\bf{h}}_2}$ represent the hidden layers and ${{\bf{h}}_3}$ is an associate memory layer.  ${{\bf{h}}_2}$ and ${{\bf{h}}_3}$ have undirected connections, while other layers are all directed connected. We add a sigmoid classifier on top of the layer ${{\bf{h}}_3}$ for classification. (b) The visualized bases (filters) of the two bottom hidden layers. We can observe that each second hidden layer base is visualized as the weighted combination of the first hidden layer bases.}
\label{fig:DBN}
\end{figure}

In order to train the deep network offline, we build a training set consisting of 100 thousand gray-scale image patches, with 50 thousand positive samples and 50 thousand negative samples.
These training image patches are sampled from a set of auxiliary video sequences which cover numerous scenes in the real world. Some of the videos are collected from the existing tracking videos (do not belong to the benchmark dataset~\cite{wu2013online}) that have ground-truth bounding boxes. A large part of the videos are self-collected in natural and the ground-truth is first generated from the output by combining the state-of-the-art tracking algorithms TLD~\cite{kalal2010pn} and CT~\cite{zhang2012real}. Then we manually rectify the bad-estimated ground-truth. 
%
The positive training patches are drawn according to the ground-truth of the video sequences, whereas the negative training patches are randomly sampled from an annular region around the target location.
Note that the negative image patches sampled in this way cover both the background and parts of the target object, which serves to incorporate discriminative information into the raw features.
%
%
Both the positive and negative training patches are resized into $32 \times 32$ pixels and are vectorized into $1024$-d column vectors with each elements belonging to $[0,1]$ after normalization.
%
%
%
%
%
%

\subsubsection{Greedy layer-wise unsupervised pretraining}
Since the appearance of an object could change significantly, we propose to learn to extract a deep hierarchical representation of the image data by employing the deep belief networks (DBN).
%
This generative model consists of a stack of Restricted Boltzmann Machines (RBM) and can be pre-trained in a greedy layer-wise unsupervised manner~\cite{hinton2006fast}.
Each layer comprises a set of binary or real-valued units, which represent features that capture higher order correlations in the original input data.
Two adjacent layers are fully connected, but units in the same layer are conditionally independent.
Once training a layer of the network, a DBN can be viewed as an RBM that defines a prior over the top layer of hidden units.
The parameters $\xi$ learned by an RBM define two conditional probabilities, $p({\bf{v}}|{\bf{h}},\xi )$ and the prior distribution over hidden vectors, $p({\bf{h}}|\xi)$.
Then a visible vector ${\bf{v}}$ can be generated with the probability:
\begin{equation}
p({\bf{v}}) = \sum\limits_{\bf{h}} {p({\bf{h}}|\xi )} p({\bf{v}}|{\bf{h}},\xi )
\end{equation}
After learning, the parameters $\xi$ are frozen and the hidden unit values are inferred. The probability $p({\bf{v}}|{\bf{h}},\xi )$ is kept, and $p({\bf{h}}|\xi )$ will be redefined by an improved model that treats the previous layer's activations as the training data for the next higher layer.
It has been shown that the variational lower bound on the training data likelihood will increase by iteratively performing this greedy algorithm.
This greedy layer-wise training approach, as being illustrated in Figure~\ref{fig:DBN}, has been shown to provide a good initialization for parameters for the multilayer network.

\subsubsection{Supervised offline training}
After initializing the weights of the DBN layer-wisely using the unsupervised method, we add a sigmoid layer as a classification layer and further fine-tune the whole network based on the classification error.
One noteworthy advantage of our training approach is that by making the deep network learn to separate the target from the background rather than just recovery corrupted images like~\cite{wang2013learning}, we can incorporate more discriminative information into the learned generic features.
To make a trade-off between the inference speed and the final performance, we choose the network architecture 1024-256-64-16-1 (from lowest to highest) and the logistic function is used as the non-linear function in the whole network.
Given the training set $\mathbf{S}=[\mathbf{s}_1,\ldots,\mathbf{s}_K]$ and the corresponding label set $\mathbf{L}=[l_1,\ldots,l_K]^{\top}$, the network is trained by minimizing the sparsity constrained euclidean loss with a weight decay as follows
\begin{equation}\label{eq:loss}
\begin{split}
  J(\Theta;\mathbf{S},\mathbf{L})=& \sum\limits_{k=1}^{K}\|f_{\Theta}(\mathbf{s}_k)-l_k\|_2^2+\gamma\sum\limits_{m}^{4}\|\mathbf{W}^m\|_F^2
  \\&+\eta\sum_{m=1}^{4}\sum_{i=1}^{n_m} \mathrm{KL}(\rho\|\hat{\rho}^m_i)
  \end{split}
\end{equation}
where $\mathbf{W}^m$ is the weight matrix of the $m$-th layer; ${n_m}$ is the number of hidden units in the m-th layer;
$\Theta$ denotes the network parameter set including the weights $\mathbf{W}^m$ and the bias $\mathbf{b}^m$;
$f_{\Theta}(\mathbf{s}_k)$ is the label predicted by the network given the input $\mathbf{s}_k$ and the parameter $\Theta$;
$\gamma$ and $\eta$ are trade-off parameters for the weight decay term and the sparse constraint, respectively;
$\mathrm{KL}(\rho\|\hat{\rho}^m_i)$ is the Kullback-Leibler (KL) divergence between the sparsity parameter $\rho$ and the average activation $\hat{\rho}^m_i$ of the $i$-th hidden unit in the $m$-th layer, and is used to impose sparsity on the activation of hidden units.

The minimization of Eq.~\ref{eq:loss} is conducted using stochastic gradient descent.
%
%
%
\begin{figure*}[tbp]
\centering
\resizebox{\textwidth}{!}{
\includegraphics{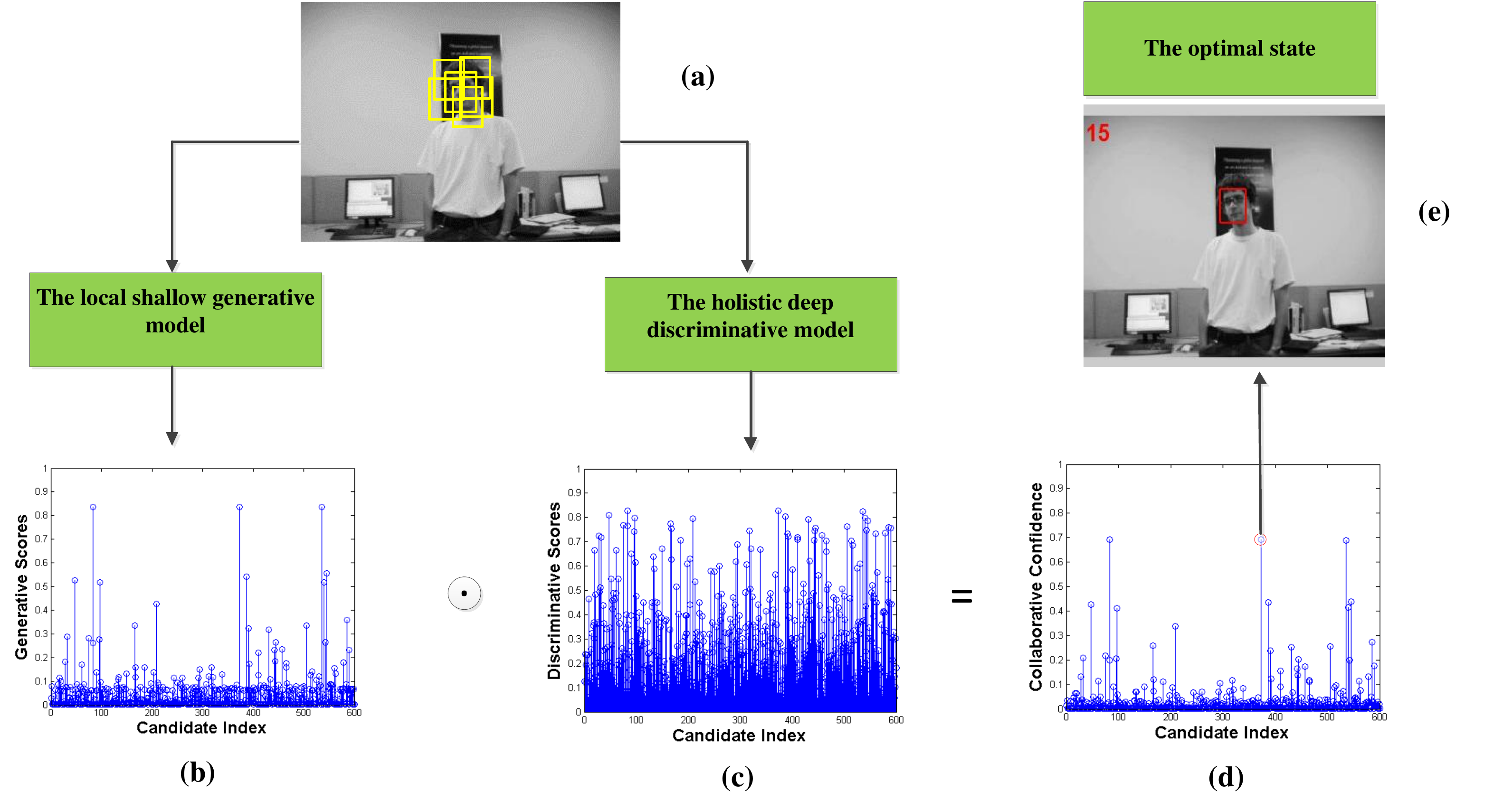}}
\vspace{-8mm}
\caption{This figure intuitively illustrates how to get the collaborative scores for all candidates and choose the best state based on it. (a) A sampled frame with candidates. (b) The score vector generated by the block-based local model and it indicates the degree of similarity to the target subspace for all candidates.  The notion $ \odot $ is the Hadamard product (element-wise product). (c) The score vector generated by the deep discriminative model and it denotes the likelihood to be the foreground for all candidates. (d) The final collaborative score vector. (e) The optimal state corresponding to the candidate that scores the highest.}
\label{fig:collaborative model}
\end{figure*}
The weights $\mathbf{W}^m$ in the $m$-th layer are updated as
\begin{equation} \label{eq:weightupdate1}
{\Delta _{i + 1}} = 0.9\cdot{\Delta _i} - 0.002\cdot\varepsilon \cdot{\bf{W}}_i^m - \varepsilon \frac{{\partial J}}{{\partial {\bf{W}}_i^m}}
\end{equation}
\begin{equation} \label{eq:weightupdate2}
\begin{array}{l}
{\bf{W}}_{i + 1}^m = {\Delta _{i + 1}} + {\bf{W}}_i^m\\
\end{array}
\end{equation}
where $i$ is the index of iterations; $\Delta $ is the momentum variable; $\varepsilon$ is the learning rate and the derivative $\frac{{\partial J}}{{\partial {{\bf{W}}^m}}} = {{\bf{h}}^{m - 1}}{({{\bf{e}}^m})^T} $  is computed as the outer product of the back-propagation error ${{\bf{e}}^m}$ and the output of the previous layer ${{\bf{h}}^{m - 1}}$.
For the output layer of the network, the back-propagation error ${{\bf{e}}^4}$ is computed as
\begin{equation}
{{\bf{e}}^4} = diag({\bf{L}}  - {\hat{\bf{L}}})diag(\hat{\bf{L}})({\bf{I}} - \hat{\bf{L}})\mathbb{}
\end{equation}
where $diag( \cdot )$ is the diagonal matrix, ${\bf{I}} \in \mathbb{R}^{K\times 1}$
denotes a column vector whose entries are all $1$ and $\hat{\mathbf{L}} = [f_{\Theta}(\mathbf{s}_1),\ldots,f_{\Theta}(\mathbf{s}_K)]^{\top}$ is the output of the deep network.
For the $m$-th hidden layer with the logistic function, the back propagation error can be computed as
\begin{equation}
{{\bf{e}}^m} = diag({({\bf{W}}^{m +1})^{\top}}{{\bf{e}}^{m + 1}})diag({{\bf{h}}^m})({\bf{I}} - {{\bf{h}}^m})
\end{equation}
where ${{\bf{W}}^{m + 1}}$ and ${{\bf{e}}^{m + 1}}$ are the weight and the error of the $(m+1)$-th layer; ${{\bf{h}}^m}$ is the output of the $m$-th layer.
\subsection{The online tracking process}

When a new frame arrives, we draw candidates according to the particle filter approach and get the discriminative score for each candidate by making candidates pass forward the network.
%
It is worth noticing that our computational complexity is very low as the forward propagation in each layer just need one time multiplication between the weights and the hierarchical features.
We set a threshold $\tau$ to decide whether the deep network should be fine-tuned online.
If the maximum confidence of all candidates is below the threshold, it indicates that the hierarchical features extracted cannot adapt to the appearance change of the target, and we fine-tune this network again.
In each frame, new foreground bounding boxes are sampled to update the positive training set in order to capture new target appearance.
If the network should be fine-tuned, we resample the negative examples from the background at a short distance from the current tracking result.
Then the online sampled positive and negative training samples are used to fine-tune the parameters.

\section{The Collaborative model}
We propose a collaborative model using both generative and discriminative model within the particle filter framework.
The discriminative score based on the hierarchical features extracted by the deep classification neural network from the holistic candidates and the similarity score based on the local PCA generative model conjunctively contribute to the robustness and effectiveness of our tracker.
Given the observation $\mathbf{y}_t^k$ of the $k$-th candidate state $\mathbf{x}_t^k$ in the $t$-th frame , the collaborative score of this candidate is defined as
\begin{equation}  \label{eq:collaborative model}
{\phi_t^k} = {\hat{G}_t^k}{f_{\Theta}(\mathbf{y}_t^k)}
\end{equation}
where $\hat{G}_t^k$ is the global generative score for the $k$-th candidate computed as Eq.~\ref{eq:refinedmask} and Eq.~\ref{eq:refinedmodel}, and $f_{\Theta}(\mathbf{y}_t^k)$ is the discriminative score generated by the deep network.
We give a summary of this collaborative model in Figure~\ref{fig:collaborative model}.
The multiplicative formula is more appropriate than the alternative additive scheme in our collaborative framework.
In the discriminative model, the classification scores related to negative candidates are suppressed to be small while those corresponding to positive candidates are given higher values which are close to the fine-tuning threshold.
%
%
By adding the appearance similarity, the best candidate can be decided among the screened positive candidates which only account for a small amount of the total candidates.
In this sense, a good candidate must have two high scores regarding the similarity level and the classification confidence respectively in order to get a high observation likelihood.
Otherwise, one small value will contribute to the suppressed final confidence.
We can observe that some collaborative scores are similar to each other.
This result is rational because we sample numerous candidates, and inevitably, some of them share the similar features, which leads to the similar responses to the collaborative model.
The observation likelihood of ${{\bf{y}}_t^k}$ is then computed within the Bayesian framework by
\begin{equation}
p({\mathbf{y}_t^k}|{\mathbf{x}_t^k}) \propto {\phi_t^k}
\end{equation}
Finally, the target state ${{\bf{x}}_t}$ can be obtained by maximizing
\begin{equation}
{{\mathbf{x}}_t} = \arg \max_{\mathbf{x}_t^k} p({\mathbf{y}_t^k}|{\mathbf{x}_t^k})
\end{equation}
\section{Experiments}
The proposed algorithm is implemented in MATLAB and runs at 7 frames per second on a 2.5 GHz i5-2450M Core PC with 4GB memory.
The parameters except for the initial calibration are fixed in all test sequences and they are summarized as follows.
We sample 5 positive training samples per frame and the positive training set has totally 50 samples.
With regard to the negative training samples, we sample 100 patches if DBN should be fine-tuned online.
The thresholds $ \tau $ and $\chi$ are all set to 0.8 and the particle filter uses 600 particles.
For offline training and online tuning, the momentum and the learning rate are set to 0.9 and 0.002 with the mini-batch size of 100 and 50, respectively.
The six affine parameters are set to [6,6,.01,.000,.000,.000].
Each image observation is normalized to $32 \times 32$ pixels and we extract $4 \times 4$ occlusion masks within the target region.
We present some representative results in this section and more results in the supplementary material.
All the MATLAB source code, dataset and supplementary material will be made available on
our website ({\small {\url{https://bitbucket.org/jingruixiaozhuang/pr2016}}}).
\subsection{Qualitative Evaluation}
We use twenty-five challenging sequences for evaluation.
The main challenging factors of these videos include occlusion, motion blur, pose variation, background clutter, illumination change, deformation and in-plane or out-of-plane rotation.
The proposed approach is compared with thirteen state-of-the-art algorithms, including IVT~\cite{ross2008incremental}, L1APG~\cite{bao2012real},  MTT~\cite{zhang2012robust}, ASLA~\cite{jia2012visual}, OSPT~\cite{wang2013online}, CT~\cite{zhang2012real}, MIL~\cite{babenko2009visual}, Frag~\cite{adam2006Frag}, TLD~\cite{kalal2010pn}, Struck~\cite{hare2011struck}, DFT~\cite{sevilla2012distribution}, CSK~\cite{henriques2012exploiting}, DLT~\cite{wang2013learning}.
We also add the last column "ours-D" to present the results of the proposed method with the discriminative model only to see how much the generative model contribute to the final performance. 
For most of the videos, we directly use the results provided by the benchmark~\cite{wu2013online} for fair evaluation.
Two criteria, the center location error as well as the overlap rate, are employed in our paper for the purpose of assessing the performance of the proposed tracker.
Table~\ref{tab:centererror} reports the average center location errors in pixels. where a smaller average center error means a more accurate tracking result.
Given the ground truth ${R_G}$ and the corresponding tracking result ${R_T}$ of each frame, we can evaluate the overlap rate by the PASCAL VOC~\cite{everingham2010pascal} criterion.
The score is defined as $score = \frac{{area({R_T} \cap {R_G})}}{{area({R_T} \cup {R_G})}} $.
Table~\ref{tab:overlaprate} reports the overlap rate, where larger average scores mean more accurate results.
Since dealing with occlusion is an important part of our paper, we provide detailed analysis in section \ref{QE}. 

\begin{table*}[htp]
\caption{ Comparison results in terms of average center error (in pixels). The best three results are shown in red, blue, and green fonts.}
\centering
\resizebox{\textwidth}{!}{
\begin{tabular}{|c|c|c|c|c|c|c|c|c|c|c|c|c|c|c|c|}
\hline \textbf{sequences} &IVT &ASLA &OSPT &CXT &CPF &CT &MIL &FRAG &TLD &Struck &DFT &CSK &DLT &Our &Our-D\\
\hline
\emph{bird}   &128.0  &127.6  &35.7  &126.8  &118.9 &\textcolor{green}{\textbf{13.4}}  &120.4  &28.3 &30.3  &53.2 &\textcolor{blue}{\textbf{10.7}} &18.3 &19.4  &\textcolor{red}{\textbf{8.8}}  &15.9\\
\emph{board}   &157.3  &70.1  &\textcolor{green}{\textbf{41.0}}  &121.2  &97.2 &56.8  &70.4 &\textcolor{blue}{\textbf{31.9}}  &44.8   &83.6  &98.4 &86.3 &65.5 &\textcolor{red}{\textbf{14.1}}  &40.6\\
\emph{boy}  &91.8  &106.1  &86.3  &7.4  &4.6  &9.0 &12.8  &40.5  &4.5  &\textcolor{green}{\textbf{3.8}}  &7.6 &20.1  &\textcolor{red}{\textbf{2.1}} &4.1 &\textcolor{blue}{\textbf{2.6}}\\
\emph{car4}   &\textcolor{blue}{\textbf{2.0}}  &\textcolor{red}{\textbf{1.7}} &67.5 &58.1 &38.7 &86.0  &50.8  &131.5  &12.9  &8.7 &37.0  &19.1  &\textcolor{green}{\textbf{2.5}}  &2.9  &3.3 \\
\emph{carDark}   &8.1  &\textcolor{red}{\textbf{1.0}}  &\textcolor{green}{\textbf{1.3}} &16.5  &57.2  &119.2  &43.5  &36.5  &27.5 &\textcolor{red}{\textbf{1.0}} &58.8  &3.2  &18.8  &\textcolor{blue}{\textbf{1.2}} &9.4 \\
\emph{caviar2}   &\textcolor{blue}{\textbf{1.6}}  &\textcolor{red}{\textbf{1.3}}  &\textcolor{green}{\textbf{2.1}}  &7.2  &54.3 &73.1 &63.1 &3.3 &21.7 &8.0 &24.5 &9.2 &59.9 &4.9 &45.0  \\
\emph{crossing}   &2.8 &\textcolor{blue}{\textbf{1.5}}  &8.6 &23.4 &9.8  &3.6  &3.2  &38.6  &24.3 &2.8 &26.1 &9.0 &\textcolor{green}{\textbf{1.6}} &\textcolor{red}{\textbf{1.4}}  &22.8\\
\emph{david2}   &\textcolor{blue}{\textbf{1.4}}   &\textcolor{blue}{\textbf{1.4}}  &\textcolor{red}{\textbf{1.3}}  &\textcolor{red}{\textbf{1.3}}  &5.3 &76.7  &10.9  &56.9  &5.0  &\textcolor{green}{\textbf{1.5}}  &2.9  &2.3  &1.9  &1.6  &1.7\\
\emph{dog1}   &\textcolor{red}{\textbf{3.5}}  &5.1  &5.0  &4.9  &7.6 &7.0  &7.8  &11.9  &\textcolor{green}{\textbf{4.2}}  &5.7  &11.0  &\textcolor{blue}{\textbf{3.8}}   &4.4  &4.8  &5.0\\
\emph{doll}   &32.4   &11.8  &6.6 &4.7  &8.6 &21.8  &16.7  &13.7  &6.0  &8.9 &7.3 &44.7 &\textcolor{green}{\textbf{5.5}}  &\textcolor{red}{\textbf{2.6}}  &\textcolor{blue}{\textbf{4.6}}\\
\emph{dudek}   &\textcolor{blue}{\textbf{9.5}}  &15.0  &\textcolor{green}{\textbf{11.5}}  &12.8  &76.4 &26.5 &17.7 &82.7 &17.9 &11.4 &10.3 &13.4  &\textcolor{red}{\textbf{8.1}}  &12.7  &13.3 \\
\emph{faceocc1}   &17.9  &78.2  &\textcolor{blue}{\textbf{12.3}} &25.3  &28.8  &25.8  &29.9  &\textcolor{red}{\textbf{11.0}} &27.4 &18.8  &20.2  &11.9  &20.1 &\textcolor{green}{\textbf{13.0}} &21.0\\
\emph{faceocc2}  &\textcolor{green}{\textbf{7.1}}  &19.4  &11.7 &6.3  &21.0   &18.9  &13.6  &16.0  &12.3  &\textcolor{blue}{\textbf{6.0}} &8.3 &\textcolor{red}{\textbf{5.9}}  &11.3  &8.2 &13.3\\
\emph{fish}   &\textcolor{green}{\textbf{5.1}}  &\textcolor{blue}{\textbf{3.4}}  &\textcolor{blue}{\textbf{3.4}}   &6.2  &40.5 &10.7  &24.1  &21.6  &6.4  &\textcolor{blue}{\textbf{3.4}}  &16.8 &41.2 &24.1 &\textcolor{red}{\textbf{2.9}}  &23.5\\
\emph{football}   &14.6  &15.2  &\textcolor{blue}{\textbf{5.7}} &\textcolor{green}{\textbf{7.4}}  &12.8  &12.0  &12.1  &\textcolor{red}{\textbf{5.4}}  &14.3  &17.3  &9.3  &16.2  &192.6  &\textcolor{red}{\textbf{5.4}}  &11.2\\
\emph{freeman1}   &11.6  &105.2  &\textcolor{red}{\textbf{7.6}}  &26.8  &12.2 &118.7  &11.2 &\textcolor{blue}{\textbf{10.1}} &39.7 &14.3 &\textcolor{green}{\textbf{10.3}}  &125.5  &103.3  &\textcolor{red}{\textbf{7.6}} &11.2\\
\emph{girl}   &22.6  &\textcolor{green}{\textbf{3.1}}  &19.5  &11.0  &18.9  &18.9 &13.7 &20.7  &9.8  &\textcolor{red}{\textbf{2.6}}   &24.0  &19.3  &10.6  &\textcolor{blue}{\textbf{3.0}} &5.2\\
\emph{mhyang}   &\textcolor{red}{\textbf{1.8}}  &\textcolor{blue}{\textbf{2.0}}  &\textcolor{green}{\textbf{2.1}} &4.0 &13.0 &13.3  &20.4 &12.5 &9.5 &2.6 &4.4 &3.6 &2.7 &2.3 &5.4\\
\emph{mountainBike}   &\textcolor{blue}{\textbf{7.4}}  &\textcolor{green}{\textbf{8.8}}  &158.3  &178.8 &211.0 &214.3  &73.0  &206.7  &101.9 &8.6  &9.8  &\textcolor{red}{\textbf{6.5}}  &14.4  &11.9 &11.6\\
\emph{singer1}   &11.7  &3.4  &45.8 &11.4  &6.5  &15.5 &16.4 &88.9 &8.0 &14.5 &\textcolor{green}{\textbf{4.2}}  &14.0 &\textcolor{blue}{\textbf{3.7}} &5.2  &\textcolor{red}{\textbf{3.6}} \\
\emph{singer2}   &175.0  &174.9  &150.0 &163.6  &50.7  &127.3  &\textcolor{green}{\textbf{22.5}}  &88.6  &\textcolor{red}{\textbf{3.3}}  &174.3  &43.7 &185.5 &172.5  &\textcolor{blue}{\textbf{10.8}} &182.7\\
\emph{shaking}   &85.3  &22.7  &111.5 &129.2 &180.7  &80.0   &24.0   &192.1  &NAN &30.7  &\textcolor{red}{\textbf{9.0}} &17.2  &195.0  &\textcolor{blue}{\textbf{10.8}} &\textcolor{blue}{\textbf{16.6}}\\
\emph{walking}   &\textcolor{red}{\textbf{1.8}}  &\textcolor{blue}{\textbf{2.1}}  &\textcolor{green}{\textbf{2.9}}  &205.7  &4.3 &6.9 &3.4 &9.3 &10.2 &4.6 &5.9 &7.2 &16.6 &7.5  &7.6\\
\emph{walking2}   &3.0  &37.9  &2.9 &34.7  &53.4 &58.5  &60.6 &57.5 &37.3 &11.2 &46.2 &17.9 &\textcolor{green}{\textbf{2.4}} &\textcolor{red}{\textbf{2.1}} &\textcolor{blue}{\textbf{2.3}}\\
\emph{woman}  &176.9  &140.3  &248.3  &72.5  &124.6   &114.5   &125.3 &111.9  &18.8 &\textcolor{red}{\textbf{4.2}}  &118.9 &207.3 &\textcolor{green}{\textbf{5.6}} &\textcolor{blue}{\textbf{5.5}} &9.5\\
\hline
\end{tabular}}
\label{tab:centererror}
\end{table*}
\begin{table*}[htp]
\caption{Comparison results in terms of average overlap rate (in pixels). The best three results are shown in red, blue, and green fonts.}
\centering
\resizebox{\textwidth}{!}{
\begin{tabular}{|c|c|c|c|c|c|c|c|c|c|c|c|c|c|c|c|c|}
\hline \textbf{sequences} &IVT &ASLA &OSPT &CXT &CPF &CT &MIL &FRAG  &TLD &Struck &DFT &CSK  &DLT &Our &Our-D \\
\hline
\emph{bird}   &0.09  &0.08  &0.47  &0.09  &0.11 &\textcolor{green}{\textbf{0.58}} &0.12 &0.54 &0.50  &0.32 &\textcolor{blue}{\textbf{0.75}} &\textcolor{green}{\textbf{0.58}} &0.50  &\textcolor{red}{\textbf{0.76}}  &0.59\\
\emph{board}   &0.15  &0.46  &\textcolor{green}{\textbf{0.68}}  &0.18  &0.30 &0.41 &0.20 &0.73 &0.49 &0.42 &0.34 &0.41 &0.47 &\textcolor{red}{\textbf{0.85}} &\textcolor{blue}{\textbf{0.76}}  \\
\emph{boy}  &0.25  &0.37  &0.52  &0.55   &0.71  &0.66 &0.50  &0.39  &0.67 &0.77 &0.63 &0.66  &\textcolor{red}{\textbf{0.83}}  &\textcolor{green}{\textbf{0.73}} &\textcolor{blue}{\textbf{0.82}}\\
\emph{car4}   &\textcolor{blue}{\textbf{0.88}}   &0.77 &0.25 &0.31 &0.17 &0.48  &0.26   &0.19  &0.63  &0.50  &0.36  &0.48  &\textcolor{red}{\textbf{0.89}}  &\textcolor{green}{\textbf{0.86}}  &0.85\\
\emph{carDark}   &0.67   &\textcolor{blue}{\textbf{0.88}}  &0.79  &0.57  &0.08 &0.75  &0.20  &0.31  &0.45  &\textcolor{red}{\textbf{0.90}} &0.38  &0.75  &0.57  &\textcolor{green}{\textbf{0.83}}  &0.81 \\
\emph{caviar2}   &\textcolor{blue}{\textbf{0.82}}  &\textcolor{red}{\textbf{0.88}}  &\textcolor{green}{\textbf{0.80}}  &0.64  &0.36 &0.55 &0.42 &0.74 &0.51 &0.58 &0.42 &0.55 &0.35 &0.71  &0.65 \\
\emph{crossing}   &0.28  &\textcolor{red}{\textbf{0.77}}  &0.60 &0.27 &0.54 &0.49  &\textcolor{blue}{\textbf{0.74}}  &0.31 &0.41 &0.69 &0.32 &0.49 &\textcolor{green}{\textbf{0.70}} &0.69 &0.30 \\
\emph{david2}   &0.66   &\textcolor{red}{\textbf{0.90}}  &0.80  &\textcolor{blue}{\textbf{0.89}}  &0.53 &0.81 &0.46 &0.24 &0.70 &\textcolor{green}{\textbf{0.87}} &0.70 &0.81  &0.83  &0.83  &0.82\\
\emph{dog1}   &\textcolor{blue}{\textbf{0.74}} &0.71 &0.63  &\textcolor{red}{\textbf{0.80}}  &0.72 &0.55 &0.54  &0.54  &0.59  &0.55 &0.60  &0.55 &0.69 &\textcolor{green}{\textbf{0.73}} &0.70 \\
\emph{doll}   &0.42  &\textcolor{red}{\textbf{0.81}}  &0.61 &0.75 &0.73   &0.32  &0.47  &0.54  &0.58  &0.54 &0.65 &0.32  &\textcolor{red}{\textbf{0.81}}  &\textcolor{blue}{\textbf{0.77}} &\textcolor{green}{\textbf{0.76}} \\
\emph{dudek}   &\textcolor{green}{\textbf{0.75}}   &0.74  &0.74 &0.73  &0.50 &0.72 &0.71 &0.54 &0.65 &0.73 &\textcolor{blue}{\textbf{0.80}} &0.72 &\textcolor{red}{\textbf{0.81}} &\textcolor{green}{\textbf{0.75}}  &0.73 \\
\emph{faceocc1}   &0.73   &0.32  &\textcolor{blue}{\textbf{0.81}} &0.64  &0.53  &\textcolor{green}{\textbf{0.80}}  &0.60  &\textcolor{red}{\textbf{0.82}} &0.59 &0.73  &0.69 &\textcolor{green}{\textbf{0.80} } &0.69  &\textcolor{green}{\textbf{0.80}} &0.69\\
\emph{faceocc2}  &0.73   &0.65  &0.72 &0.75 &0.43 &\textcolor{blue}{\textbf{0.78}}   &0.68  &0.65  &0.62 &\textcolor{red}{\textbf{0.79}} &0.74  &\textcolor{blue}{\textbf{0.78}}  &0.66 &\textcolor{green}{\textbf{0.76}} &0.63 \\
\emph{fish}   &0.78  &\textcolor{blue}{\textbf{0.86}}  &\textcolor{green}{\textbf{0.85}} &0.79  &0.19 &0.21  &0.46  &0.55  &0.81 &\textcolor{blue}{\textbf{0.86}} &0.56  &0.21  &0.45  &\textcolor{red}{\textbf{0.89}} &0.45\\
\emph{football}   &0.56  &0.54  &\textcolor{red}{\textbf{0.70}}  &0.55  &\textcolor{green}{\textbf{0.64}}  &0.56  &0.59 &\textcolor{red}{\textbf{0.70}} &0.50  &0.55  &\textcolor{blue}{\textbf{0.66}} &0.56  &0.25  &\textcolor{red}{\textbf{0.70}} &0.47 \\
\emph{freeman1}   &0.42   &0.27 &\textcolor{green}{\textbf{0.49}} &0.35 &0.37 &0.24 &0.35 &0.38 &0.29 &0.34  &0.36 &0.24  &0.30  &\textcolor{blue}{\textbf{0.51}} &\textcolor{red}{\textbf{0.54}} \\
\emph{girl}    &0.16   &\textcolor{blue}{\textbf{0.70}}  &0.38  &0.56  &0.44 &0.38  &0.41  &0.46 &0.58  &\textcolor{red}{\textbf{0.75}} &0.29 &0.38  &0.52  &\textcolor{green}{\textbf{0.64}} &0.56\\
\emph{mhyang}   &0.78  &\textcolor{red}{\textbf{0.90}}  &0.81 &0.85  &0.35 &0.80 &0.51 &0.65 &0.64 &0.82 &0.73 &0.80 &\textcolor{blue}{\textbf{0.89}}&\textcolor{green}{\textbf{0.88}}  &0.81\\
\emph{mountainBike}   &\textcolor{red}{\textbf{0.74}}   &\textcolor{red}{\textbf{0.73}}  &0.28 &0.23  &0.11 &\textcolor{green}{\textbf{0.71}}  &0.46  &0.13  &0.20 &\textcolor{green}{\textbf{0.71}}  &0.70  &\textcolor{green}{\textbf{0.71}}  &0.46 &0.65 &0.53 \\
\emph{singer1}   &0.57   &\textcolor{green}{\textbf{0.80}}  &0.36 &0.50  &0.45 &0.36 &0.36 &0.21 &0.73 &0.37 &0.49  &0.36  &\textcolor{red}{\textbf{0.85}}  &0.76 &\textcolor{blue}{\textbf{0.81}} \\
\emph{singer2}   &0.04  &0.04  &0.06  &0.07  &0.22 &0.04  &\textcolor{blue}{\textbf{0.52}}  &0.20  &0.03 &0.04 &\textcolor{green}{\textbf{0.42 }} &0.04  &0.04  &\textcolor{red}{\textbf{0.69}}  &0.03\\
\emph{shaking}   &0.03   &0.46  &0.05 &0.12  &0.13 &0.58  &0.43 &0.08  &NAN   &0.35   &\textcolor{red}{\textbf{0.71}}  &\textcolor{green}{\textbf{0.58}}  &0.01 &\textcolor{blue}{\textbf{0.61}} &0.57 \\
\emph{walking}   &\textcolor{blue}{\textbf{0.73}}   &\textcolor{green}{\textbf{0.74}}  &\textcolor{red}{\textbf{0.75}}   &0.17 &0.65 &0.55  &0.55  &0.55 &0.46 &0.59 &0.57 &0.55 &0.40 &0.44 &0.42 \\
\emph{walking2}  &\textcolor{green}{\textbf{0.76}}   &0.36  &\textcolor{blue}{\textbf{0.79}} &0.37  &0.32 &0.47 &0.29 &0.28  &0.29 &0.53 &0.33 &0.47 &\textcolor{red}{\textbf{0.80}} &\textcolor{red}{\textbf{0.80}} &\textcolor{blue}{\textbf{0.79}}\\
\emph{woman}  &0.15   &0.15  &0.16  &0.20  &0.07 &0.20  &0.16  &0.15  &0.13  &\textcolor{red}{\textbf{0.73}}  &0.15 &0.20  &\textcolor{green}{\textbf{0.61}} &0.54 &\textcolor{blue}{\textbf{0.63}} \\
\hline
\end{tabular}}
\label{tab:overlaprate}
\end{table*}
\subsection{Qualitative Evaluation} \label{QE}
\begin{figure*}[tbp]
\centering
\begin{tabular}{c@{}c@{}c@{}c@{}c@{}c}
\centering
\includegraphics[width=0.15\linewidth, height=0.08\linewidth]{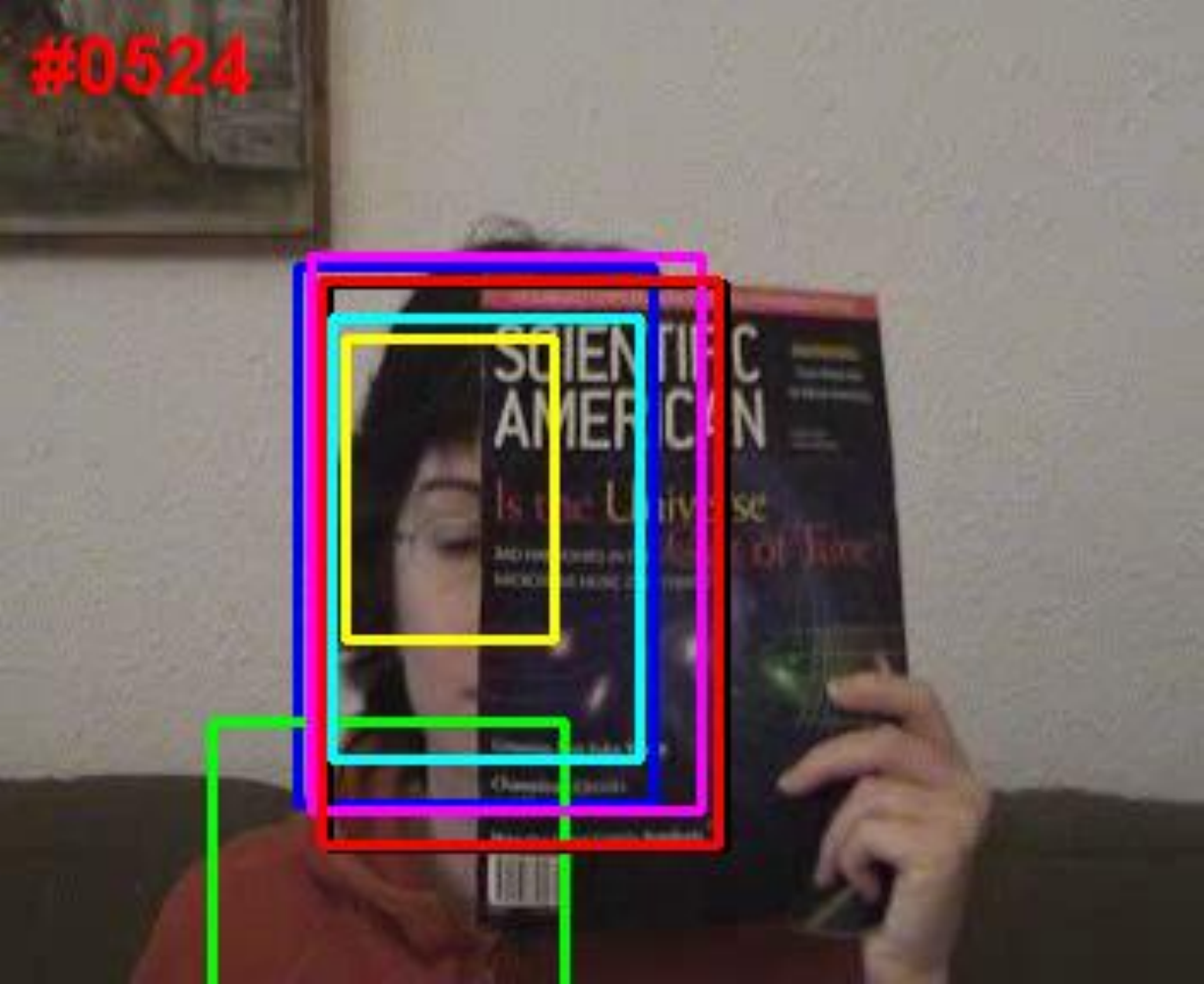}
&
\includegraphics[width=0.15\linewidth, height=0.08\linewidth]{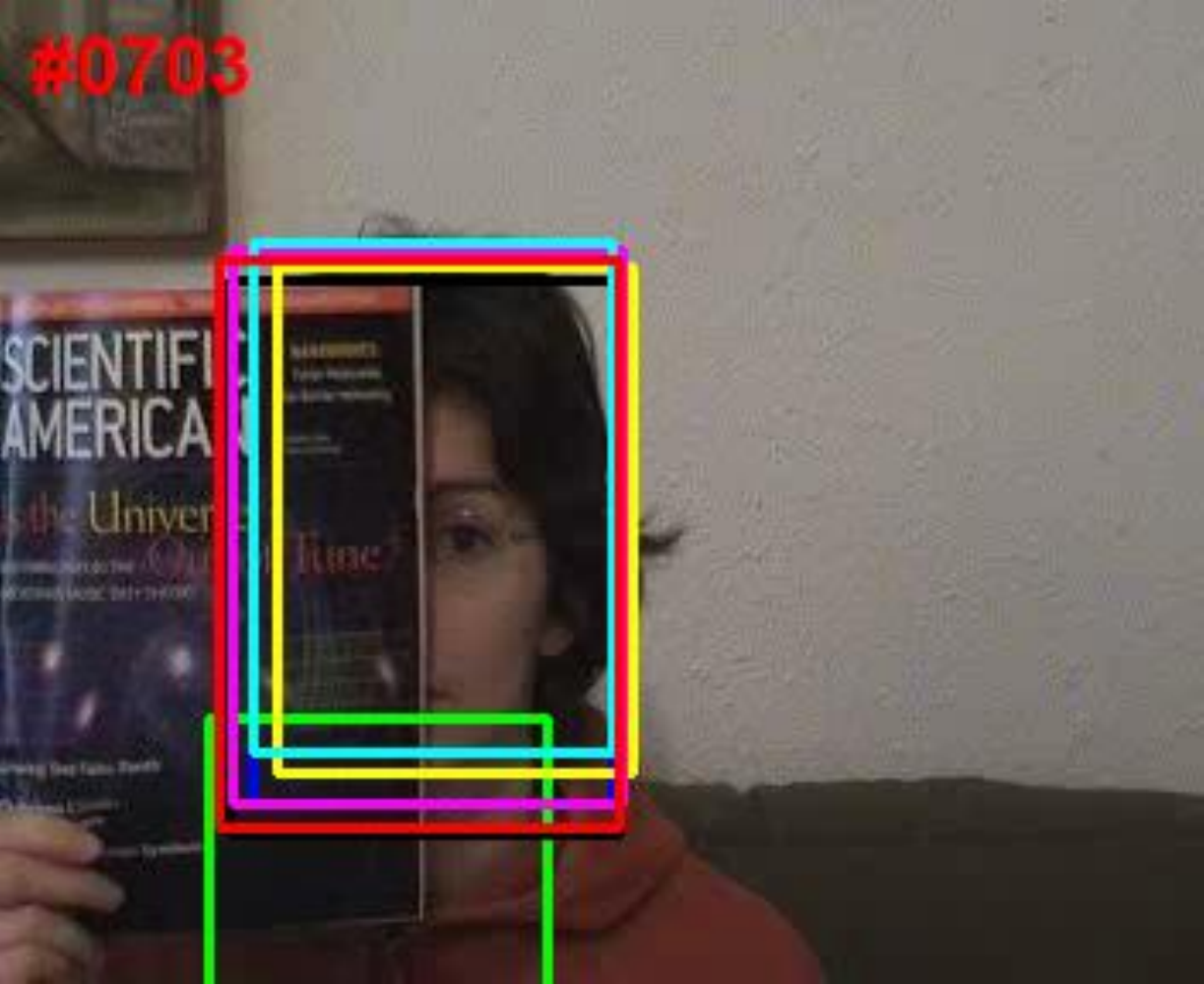}
&
\includegraphics[width=0.15\linewidth, height=0.08\linewidth]{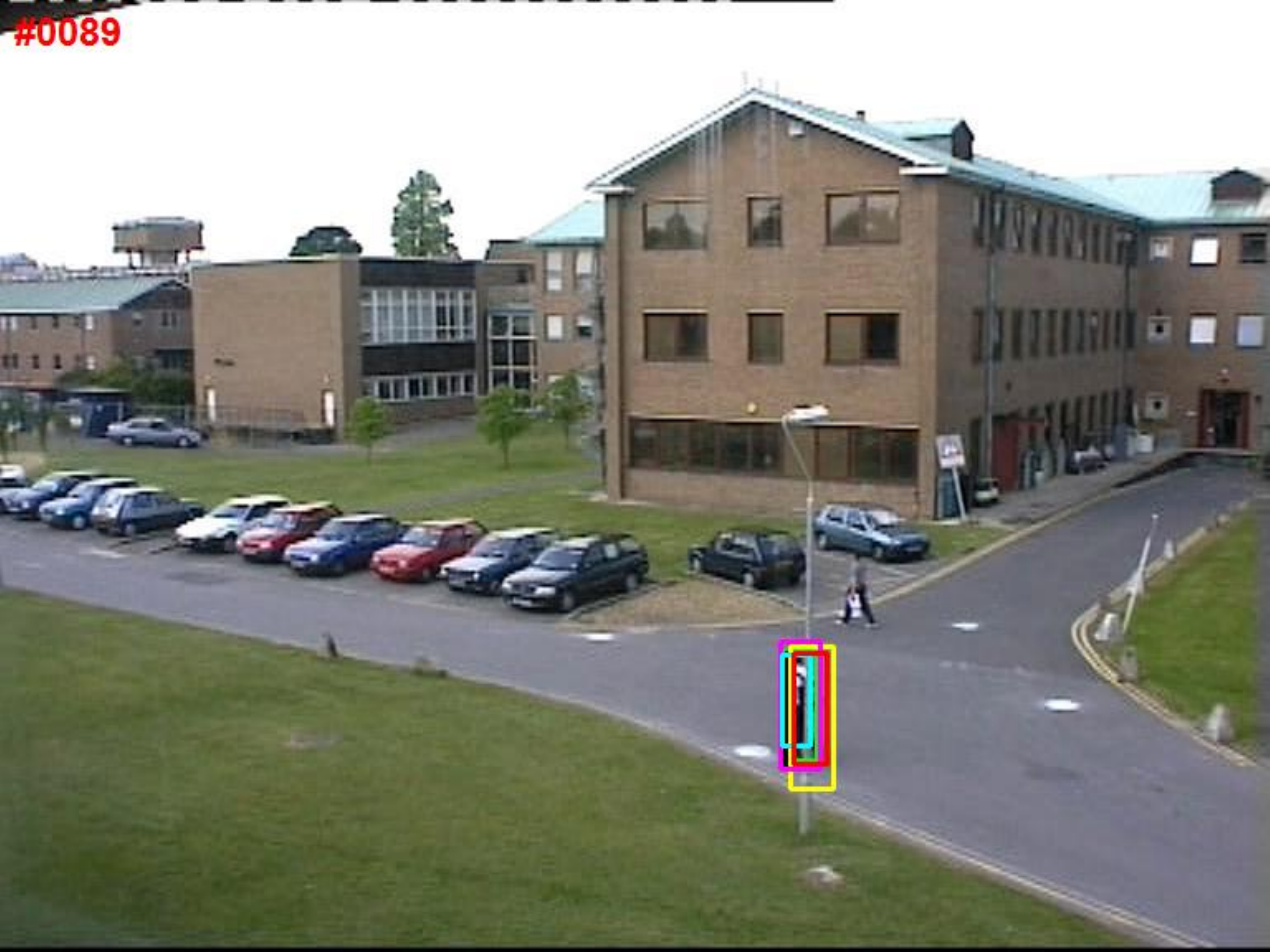}
&
\includegraphics[width=0.15\linewidth,height=0.08\linewidth]{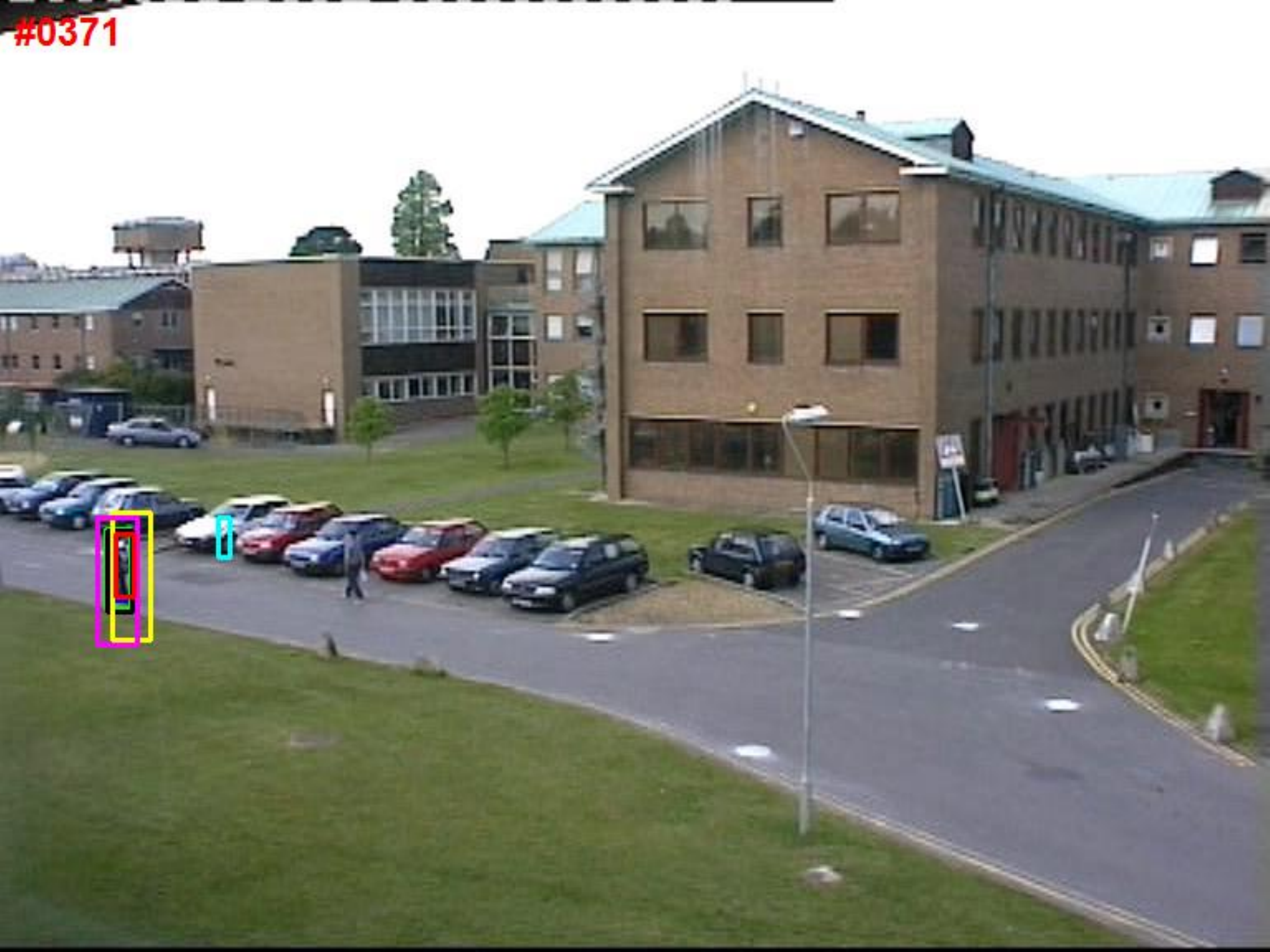}
&
\includegraphics[width=0.15\linewidth,height=0.08\linewidth]{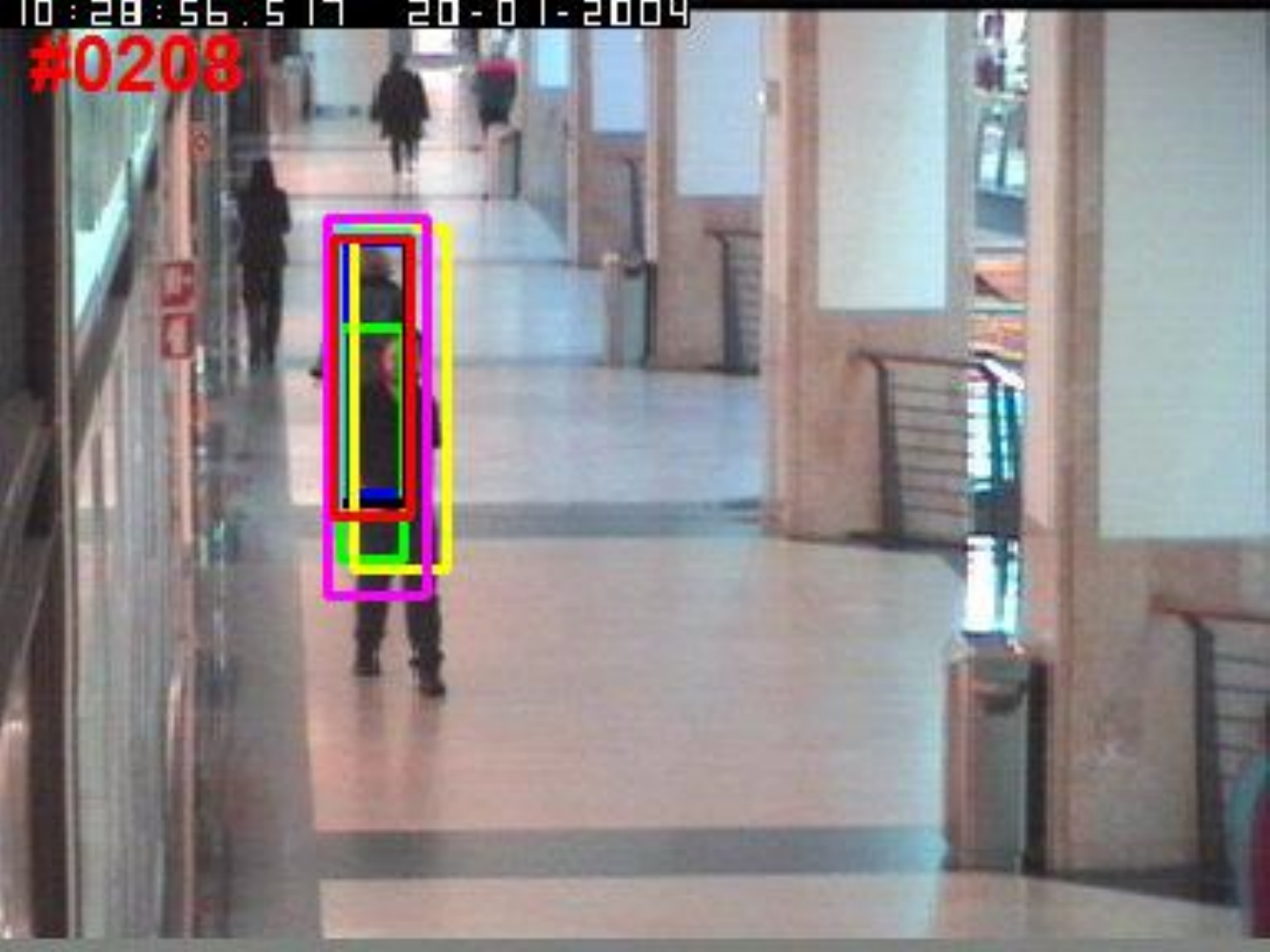}
&
\includegraphics[width=0.15\linewidth,height=0.08\linewidth]{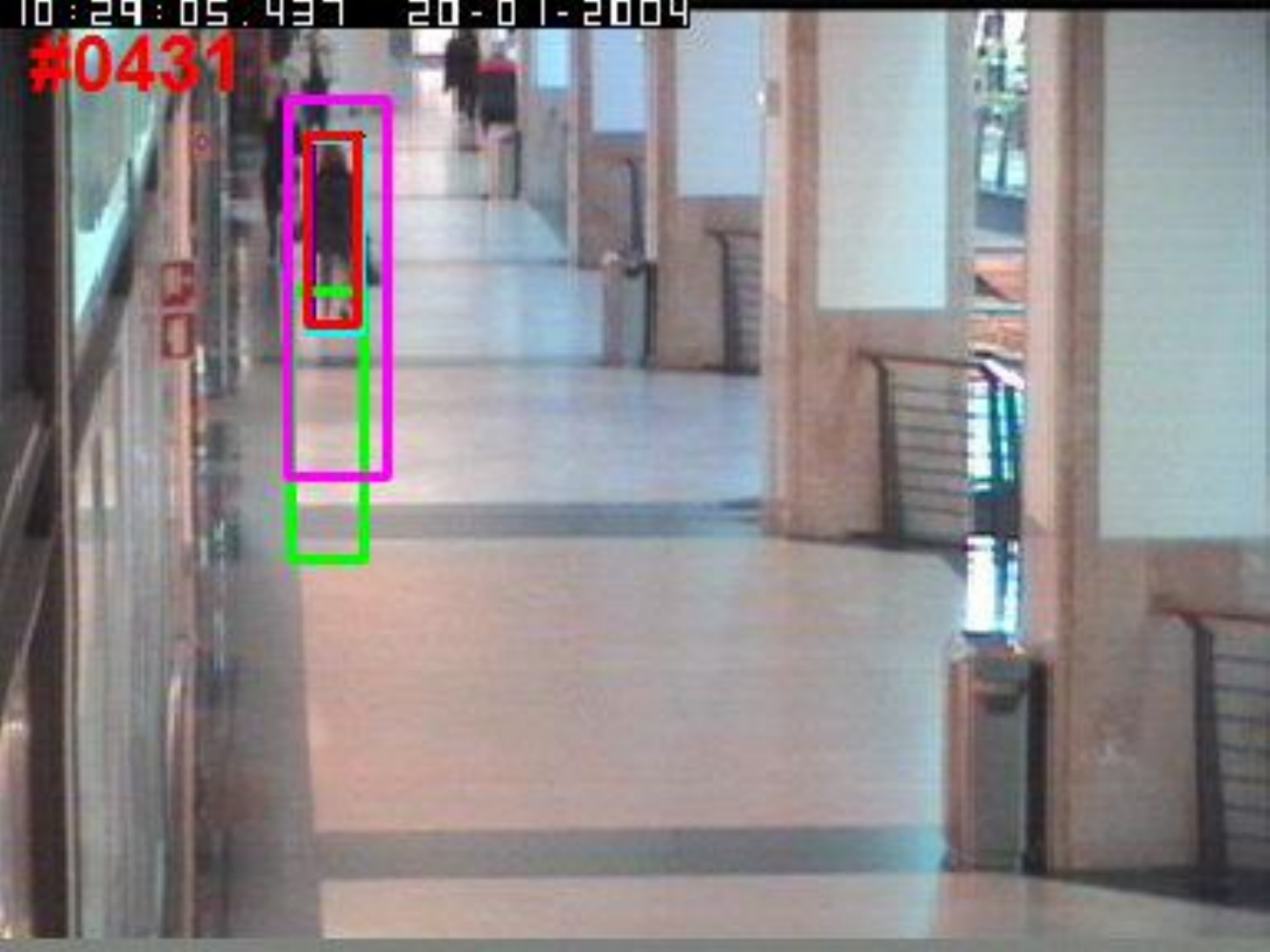}
\\
\end{tabular}

(a) \emph{faceocc1}, \emph{walking} and \emph{walking2} with partial occlusion.
\begin{tabular}{c@{}c@{}c@{}c@{}c@{}c}
\includegraphics[width=0.15\linewidth, height=0.08\linewidth]{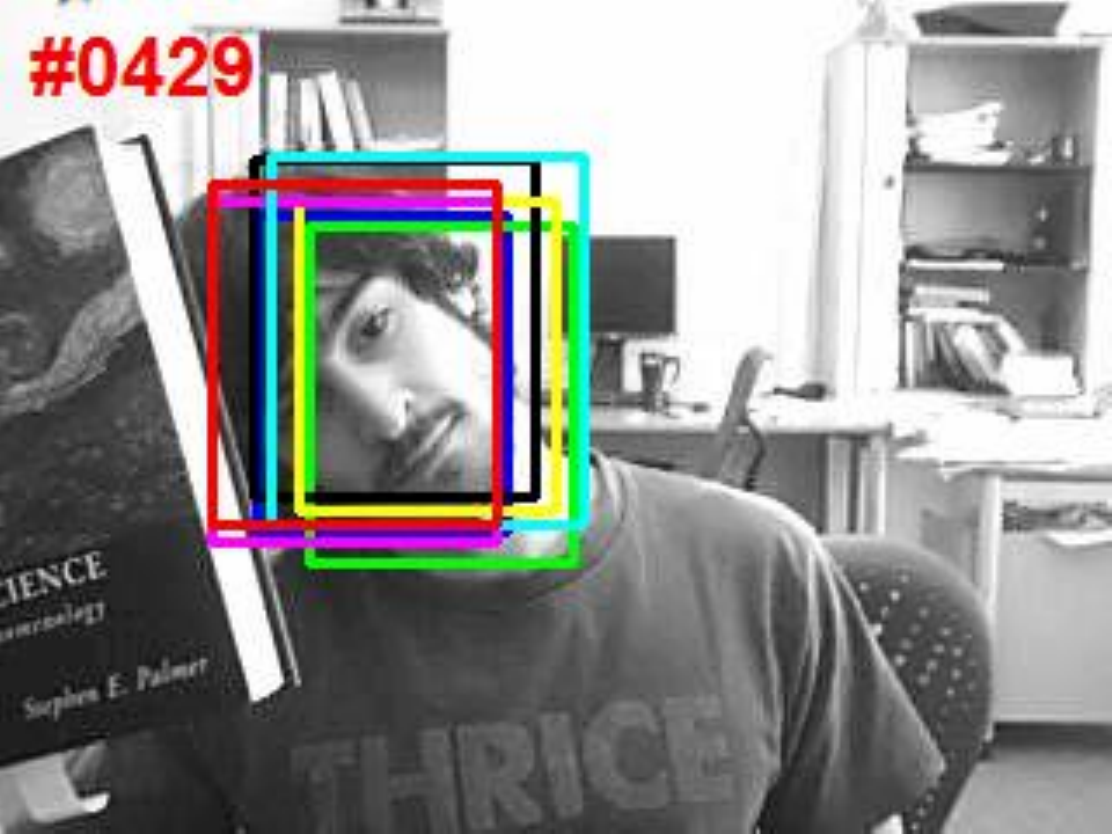}
&
\includegraphics[width=0.15\linewidth, height=0.08\linewidth]{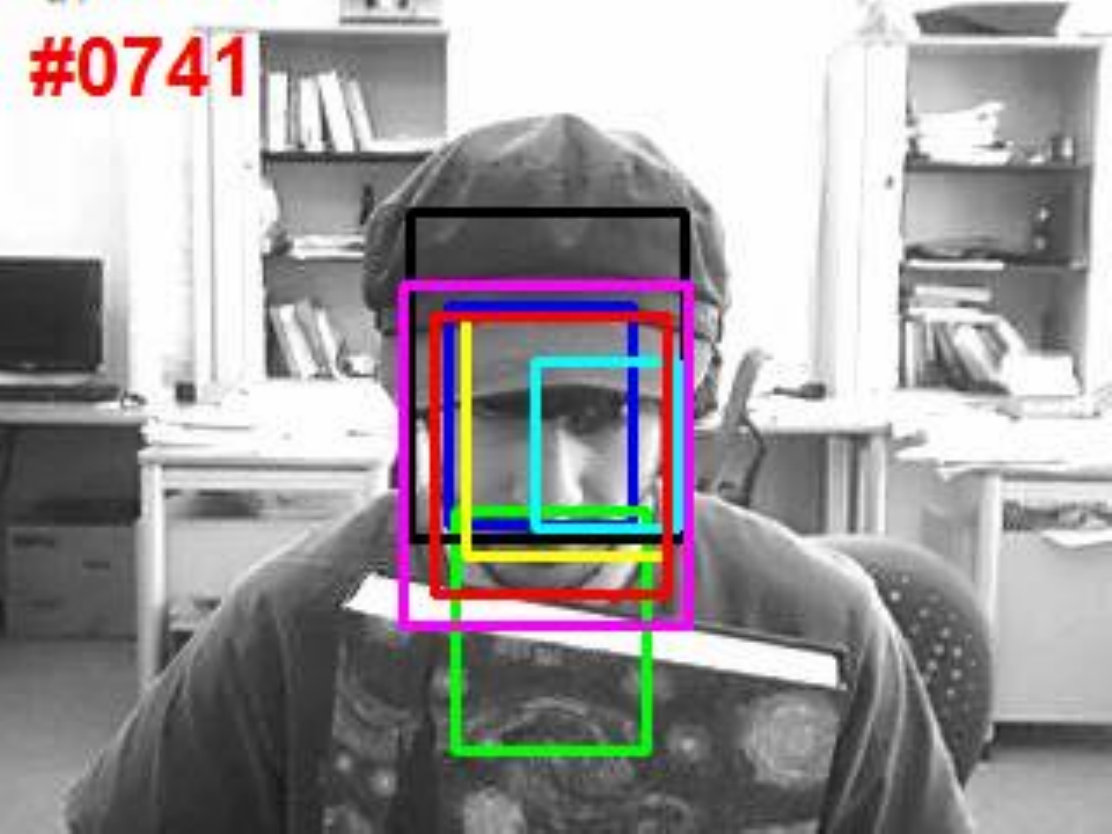}
&
\includegraphics[width=0.15\linewidth, height=0.08\linewidth]{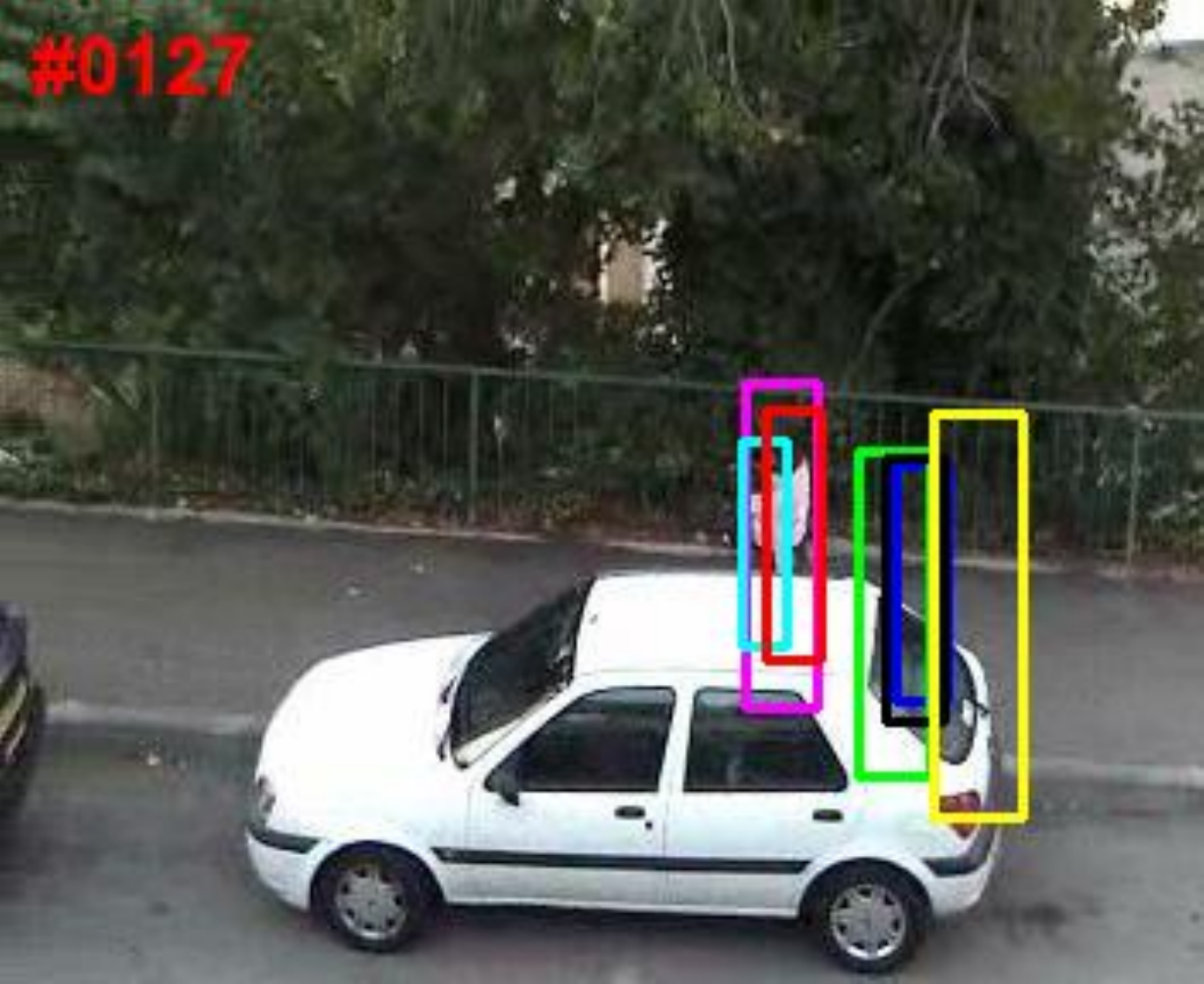}
&
\includegraphics[width=0.15\linewidth, height=0.08\linewidth]{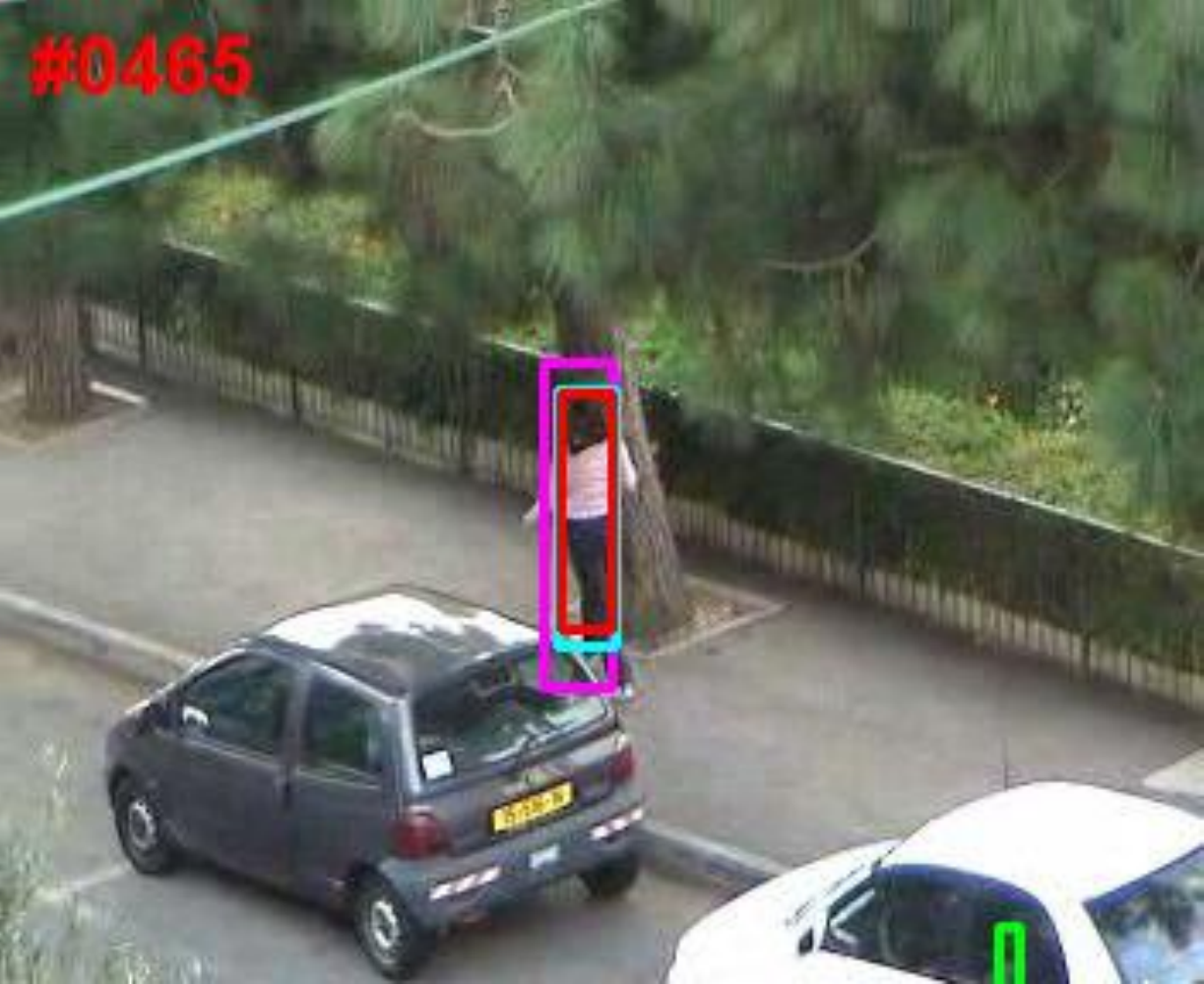}
&
\includegraphics[width=0.15\linewidth, height=0.08\linewidth]{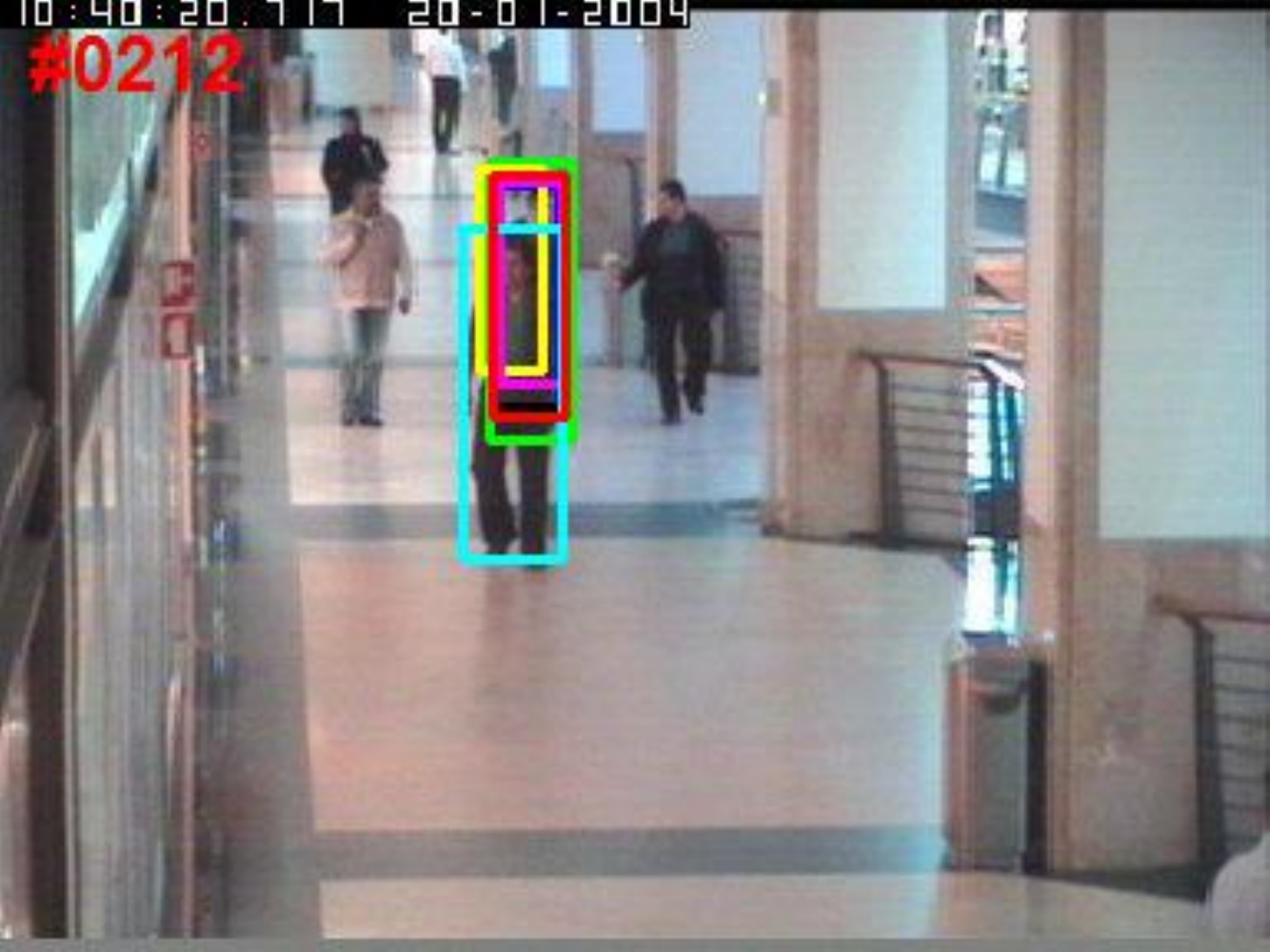}
&
\includegraphics[width=0.15\linewidth, height=0.08\linewidth]{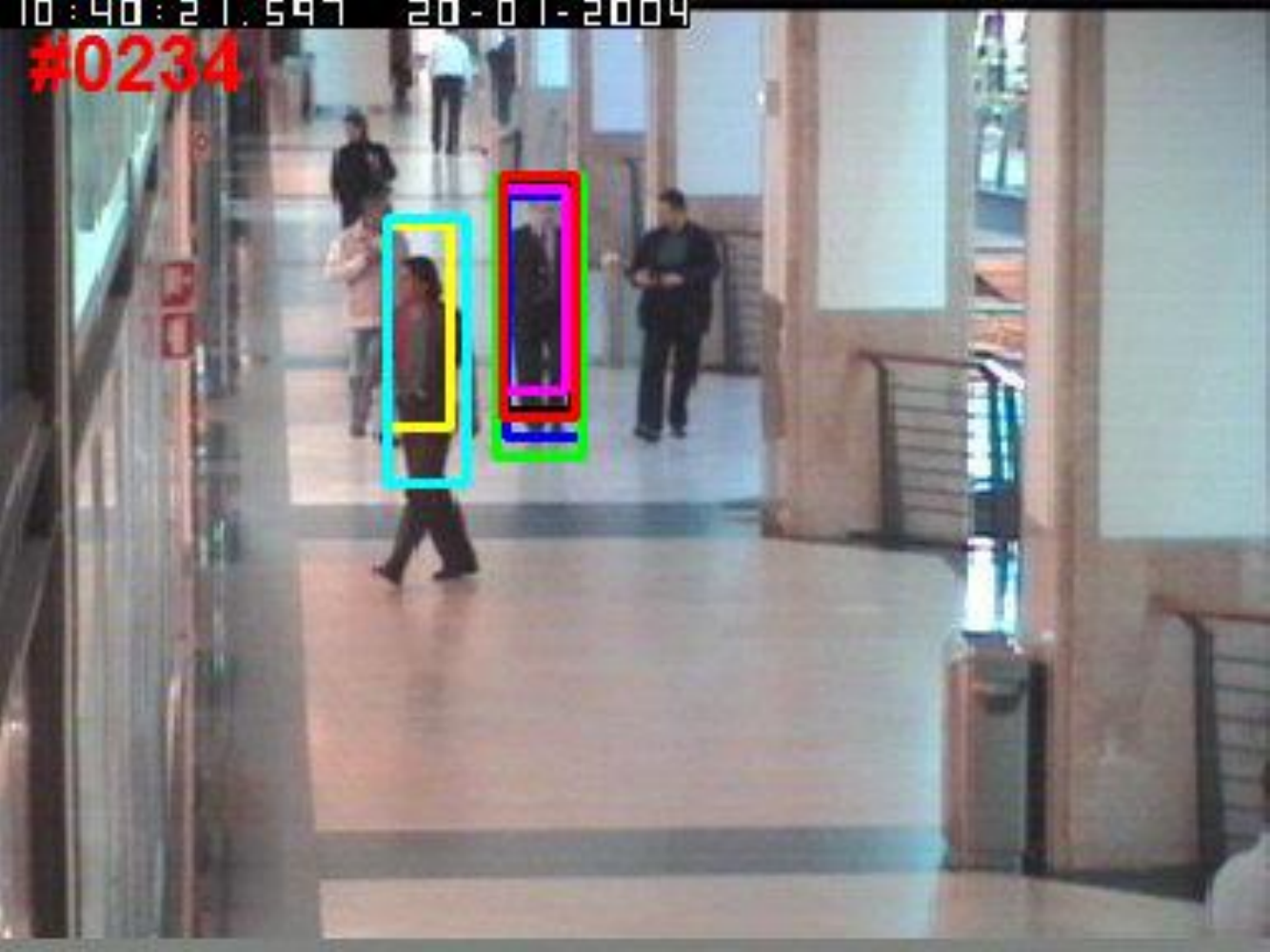}
\\
\end{tabular}

(b) \emph{faceocc2}, \emph{woman} and \emph{caviar2} with heavy occlusion.
\begin{tabular}{c@{}c@{}c@{}c@{}c@{}c}
\includegraphics[width=0.15\linewidth, height=0.08\linewidth]{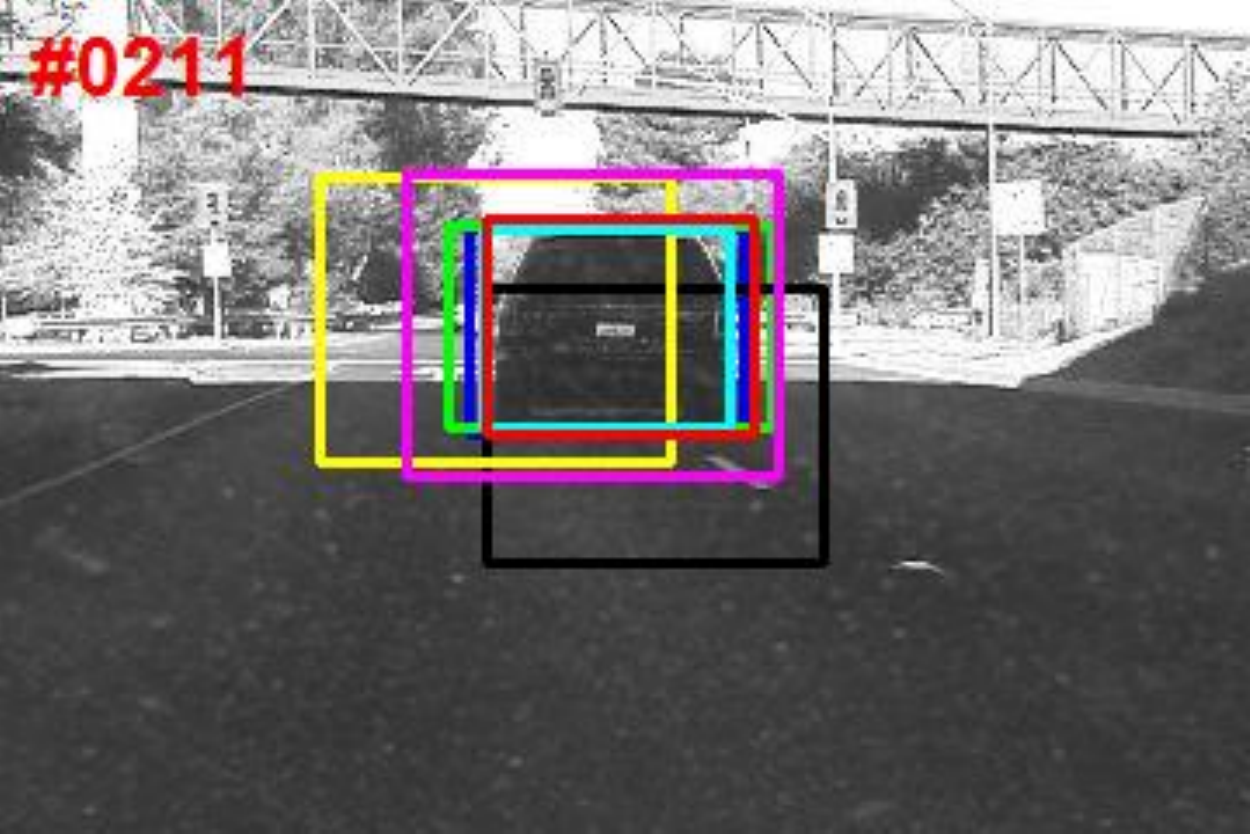}
&
\includegraphics[width=0.15\linewidth, height=0.08\linewidth]{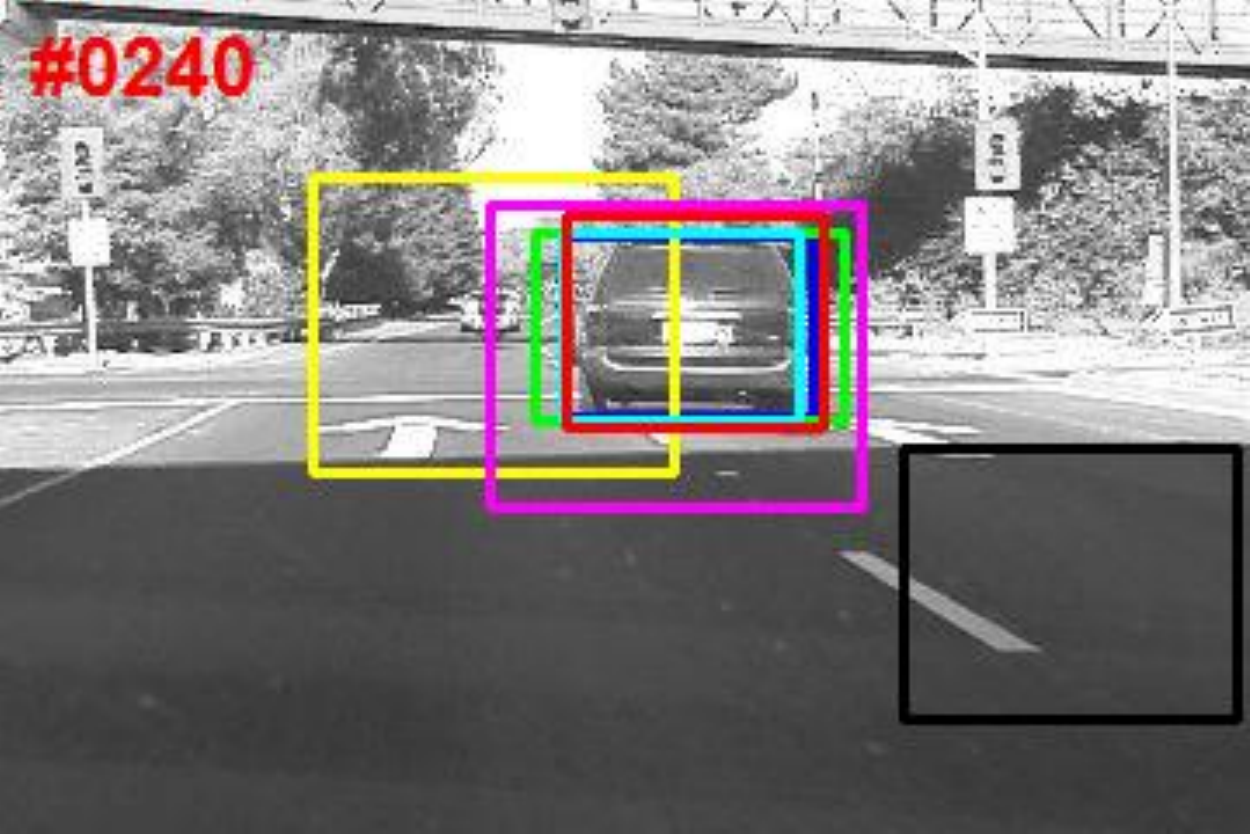}
&
\includegraphics[width=0.15\linewidth, height=0.08\linewidth]{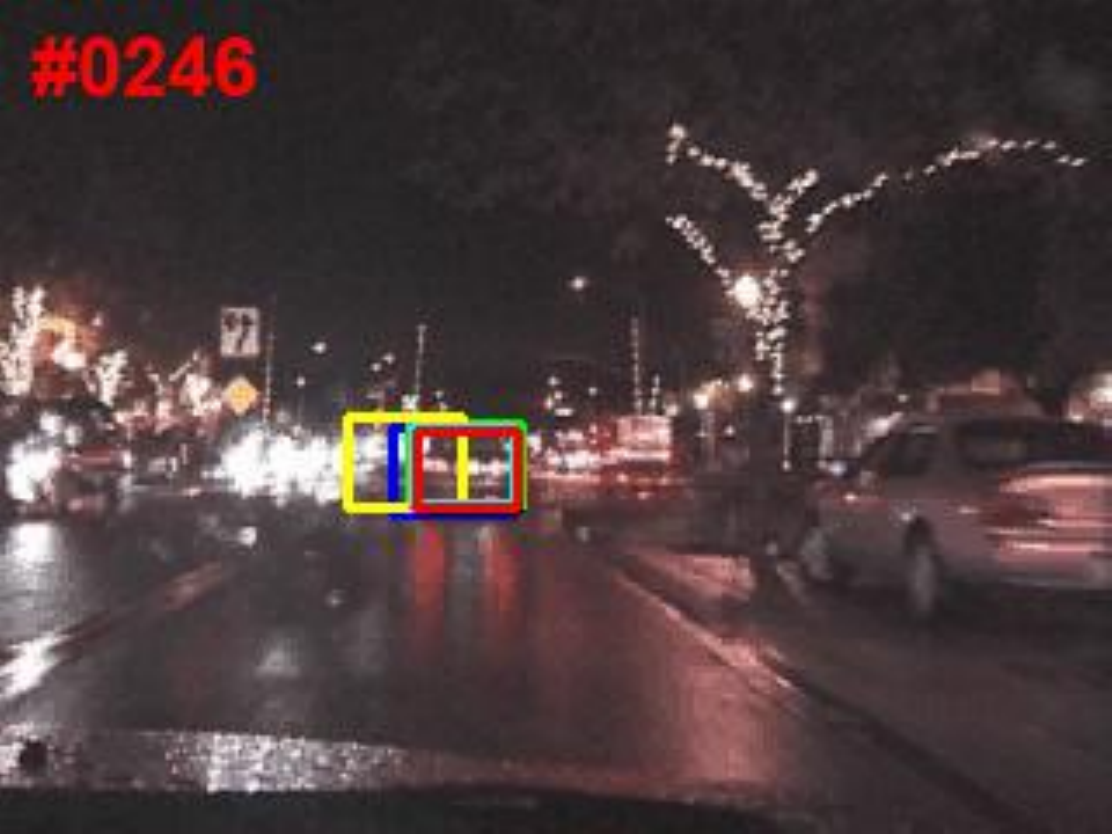}
&
\includegraphics[width=0.15\linewidth, height=0.08\linewidth]{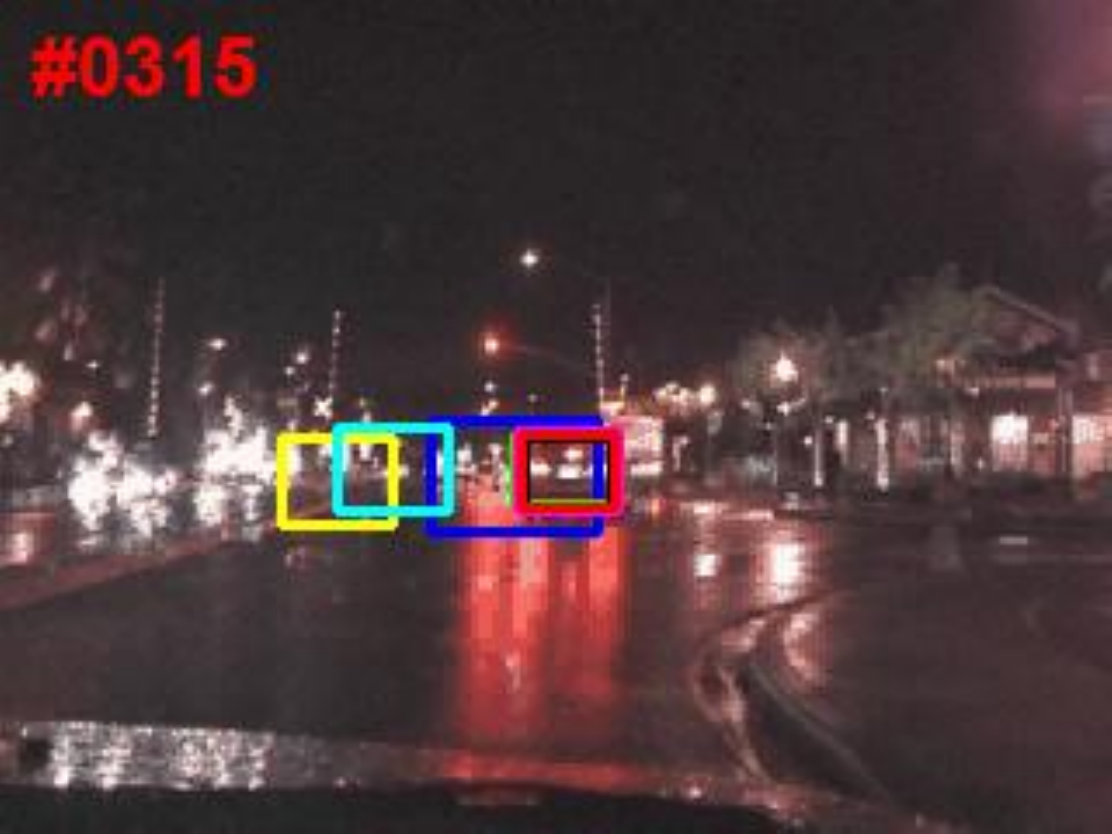}
&
\includegraphics[width=0.15\linewidth, height=0.08\linewidth]{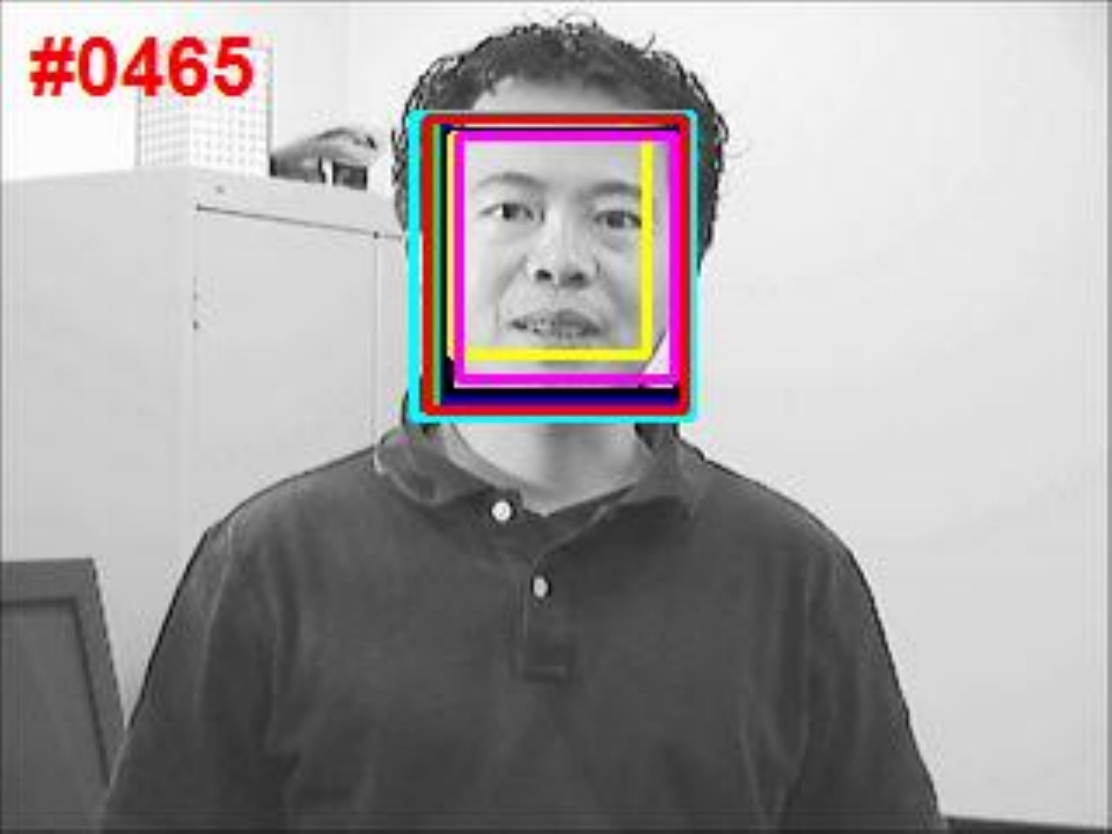}
&
\includegraphics[width=0.15\linewidth, height=0.08\linewidth]{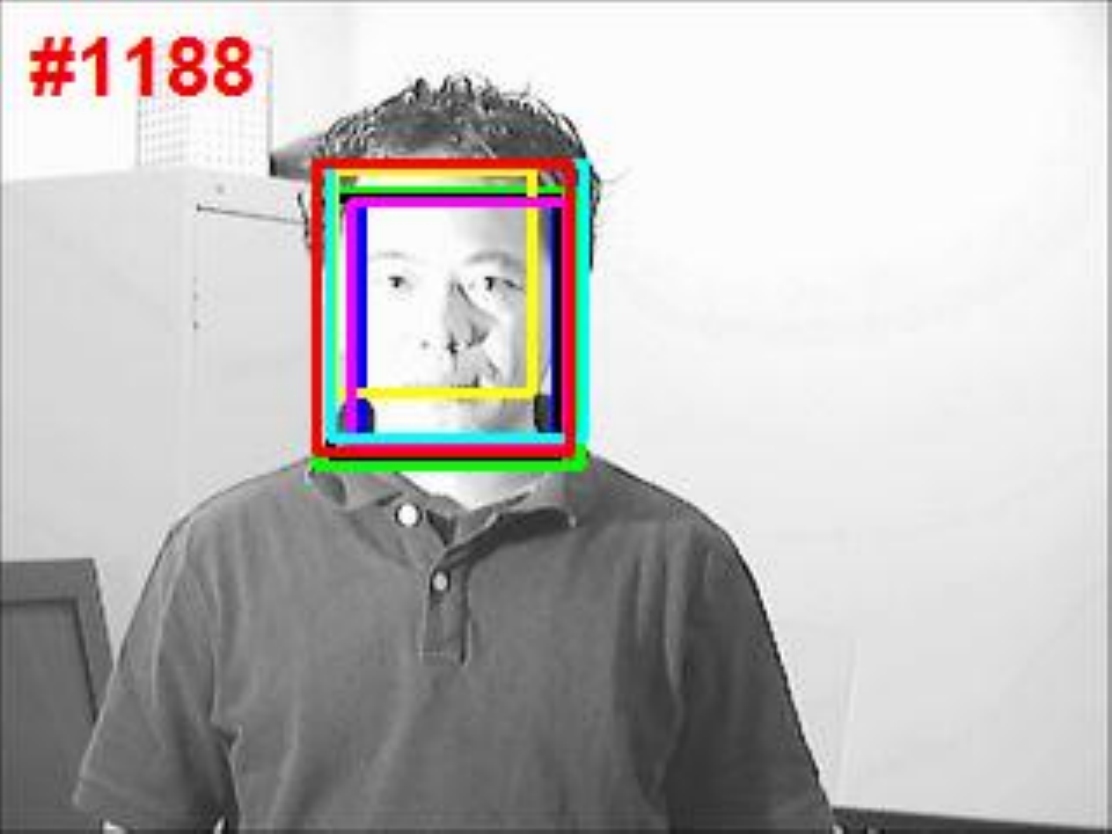}
\\
\end{tabular}

(c) \emph{car4}, \emph{carDark} and \emph{mhyang} with illumination variation.
\begin{tabular}{c@{}c@{}c@{}c@{}c@{}c}
\includegraphics[width=0.15\linewidth, height=0.08\linewidth]{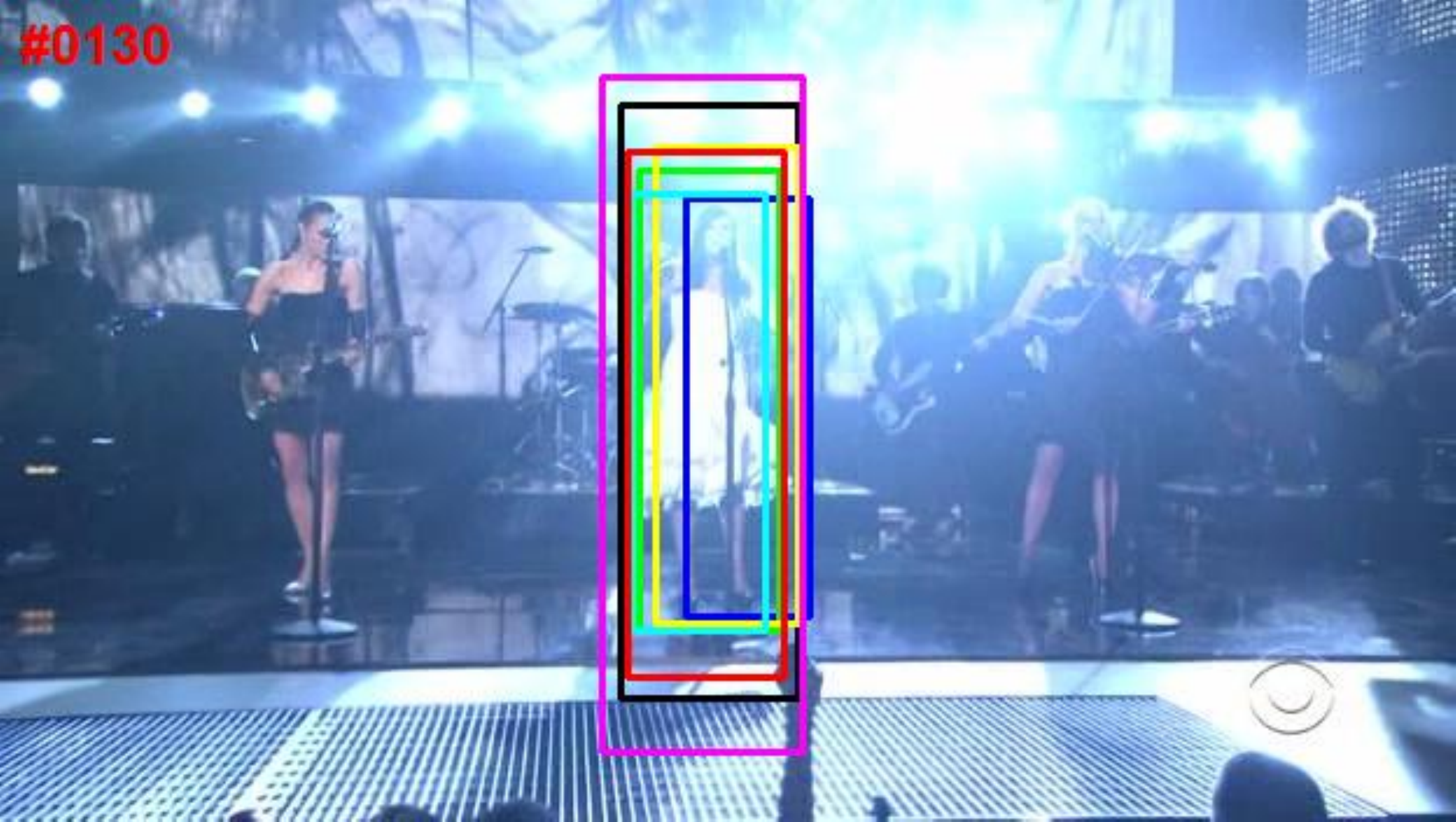}
&
\includegraphics[width=0.15\linewidth, height=0.08\linewidth]{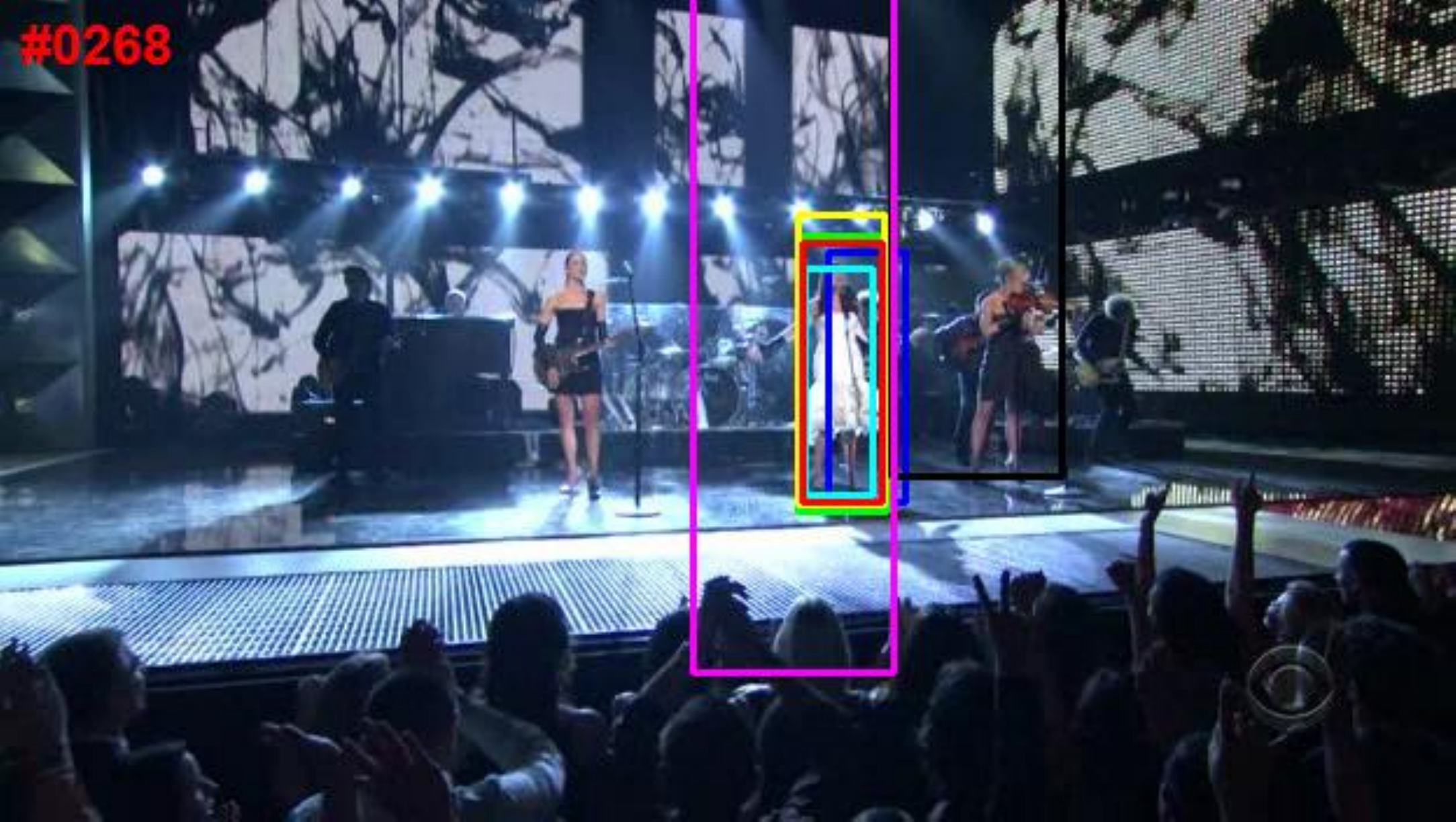}
&
\includegraphics[width=0.15\linewidth, height=0.08\linewidth]{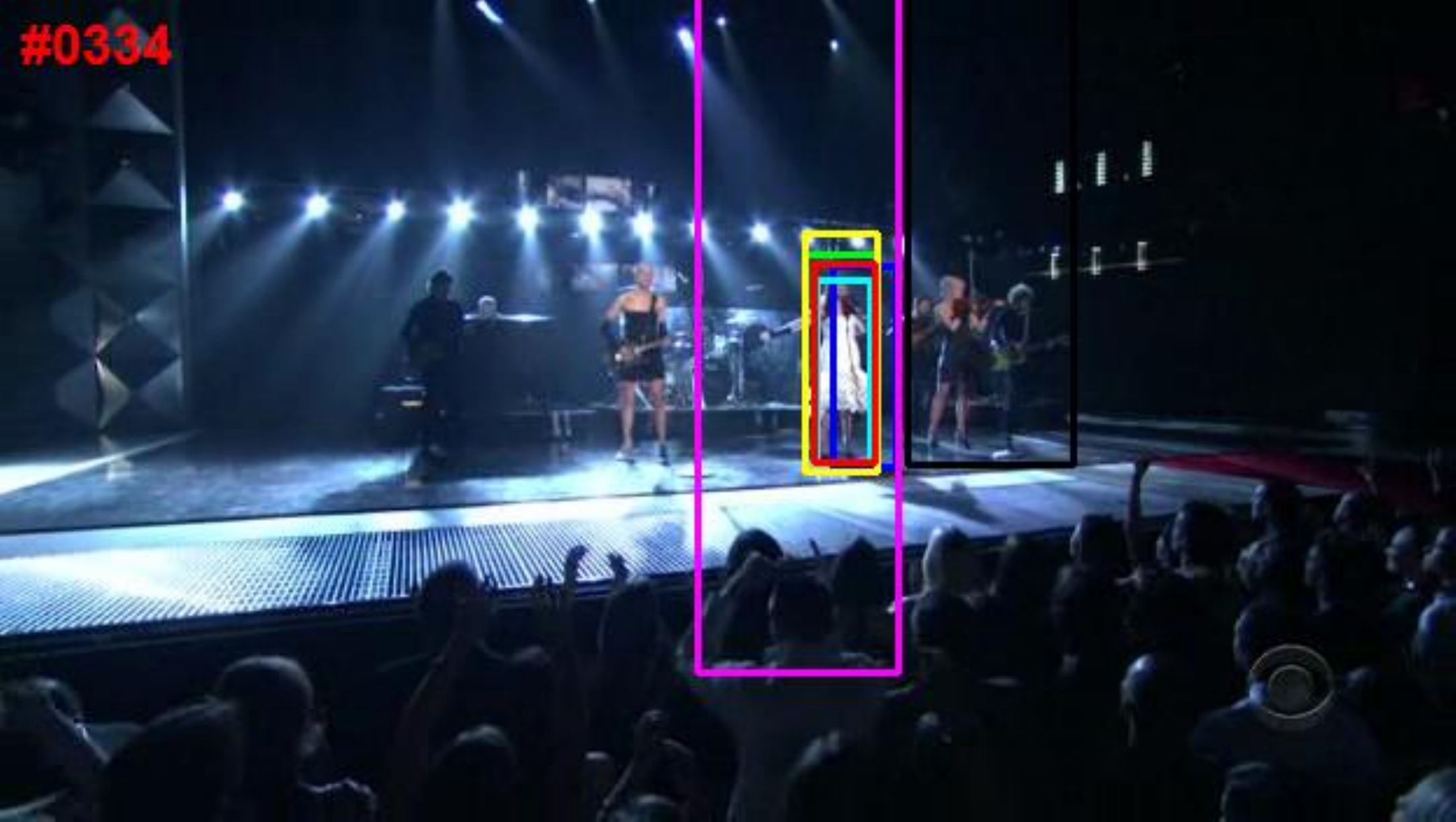}
&
\includegraphics[width=0.15\linewidth, height=0.08\linewidth]{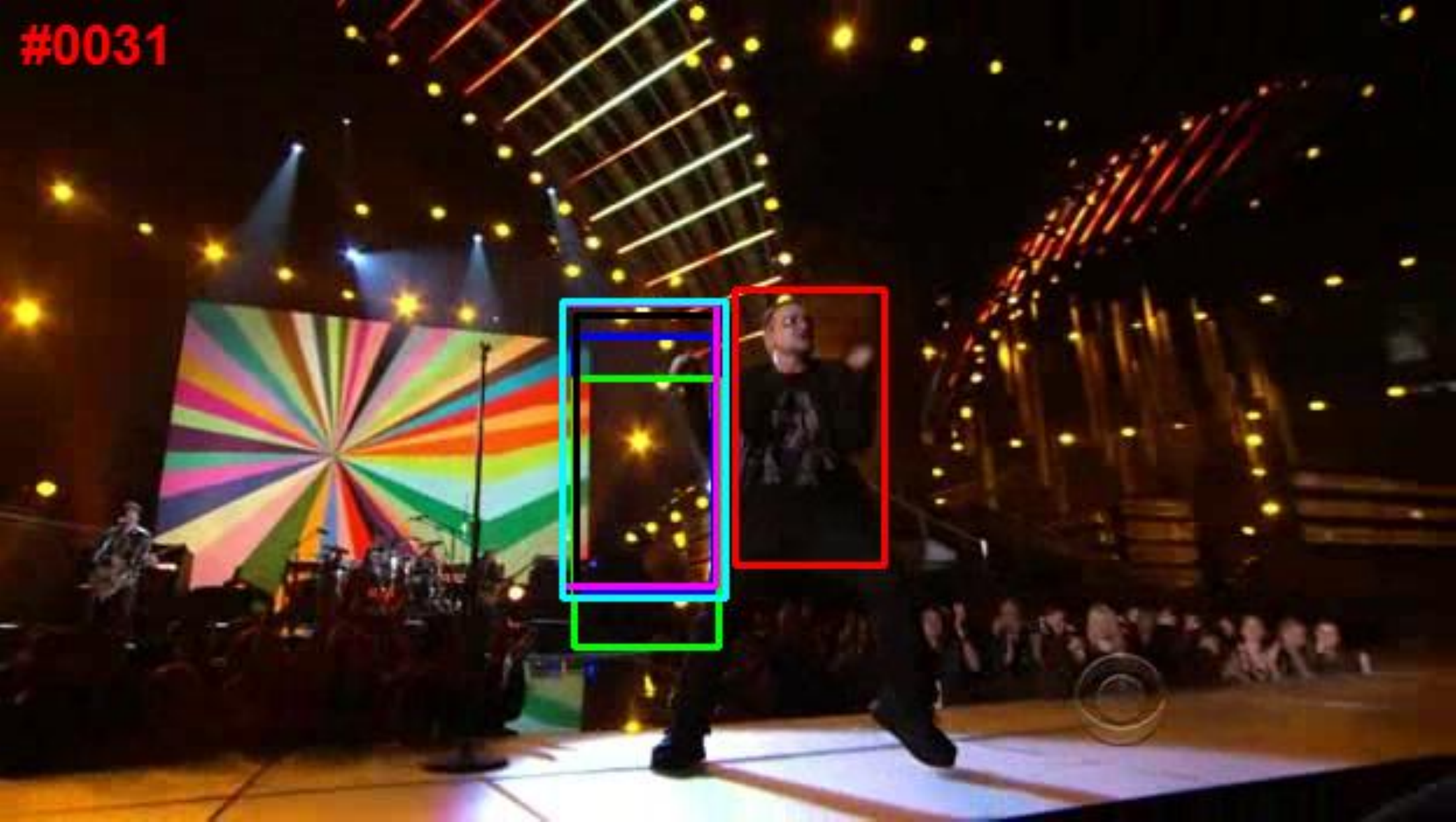}
&
\includegraphics[width=0.15\linewidth, height=0.08\linewidth]{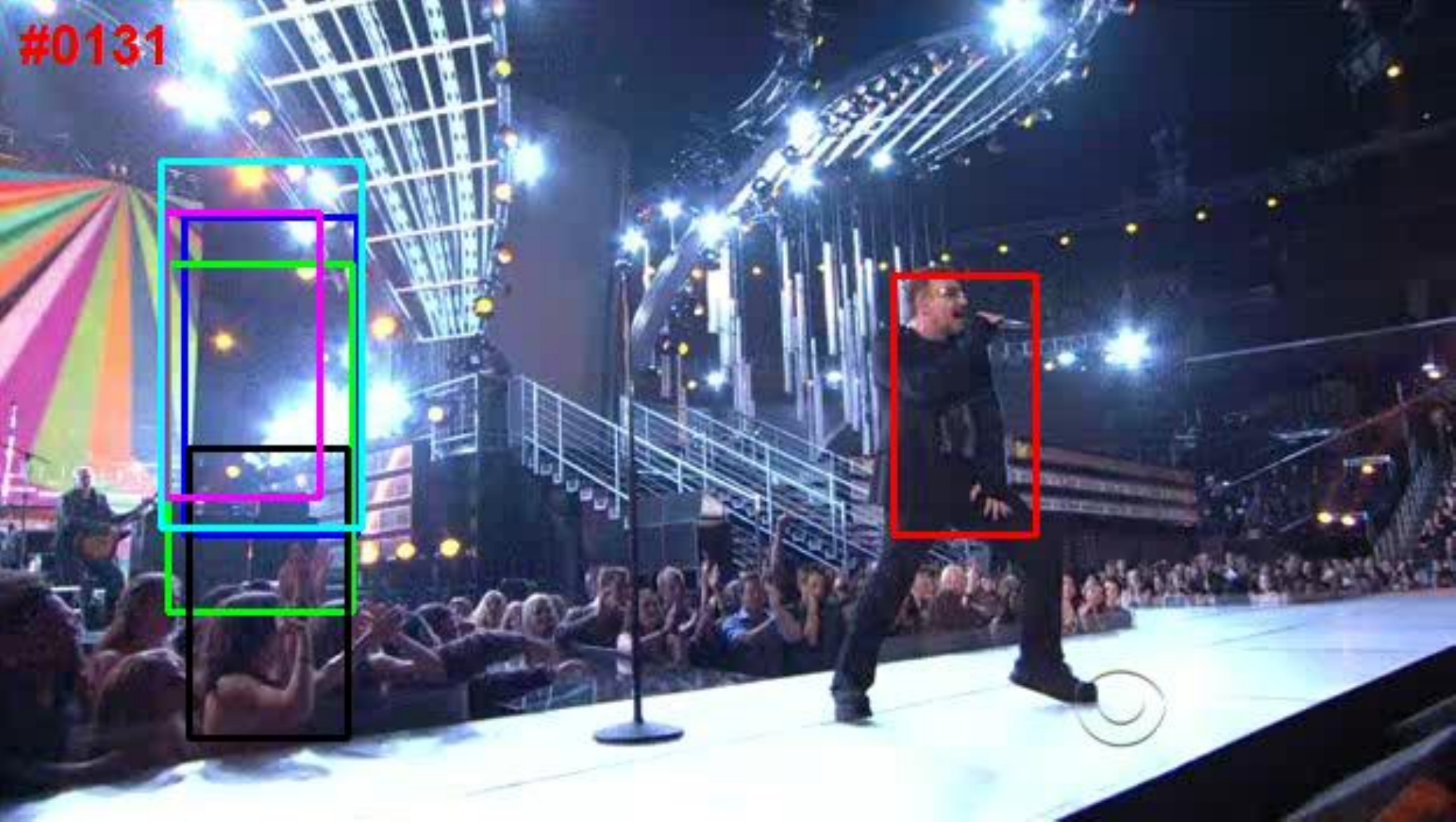}
&
\includegraphics[width=0.15\linewidth, height=0.08\linewidth]{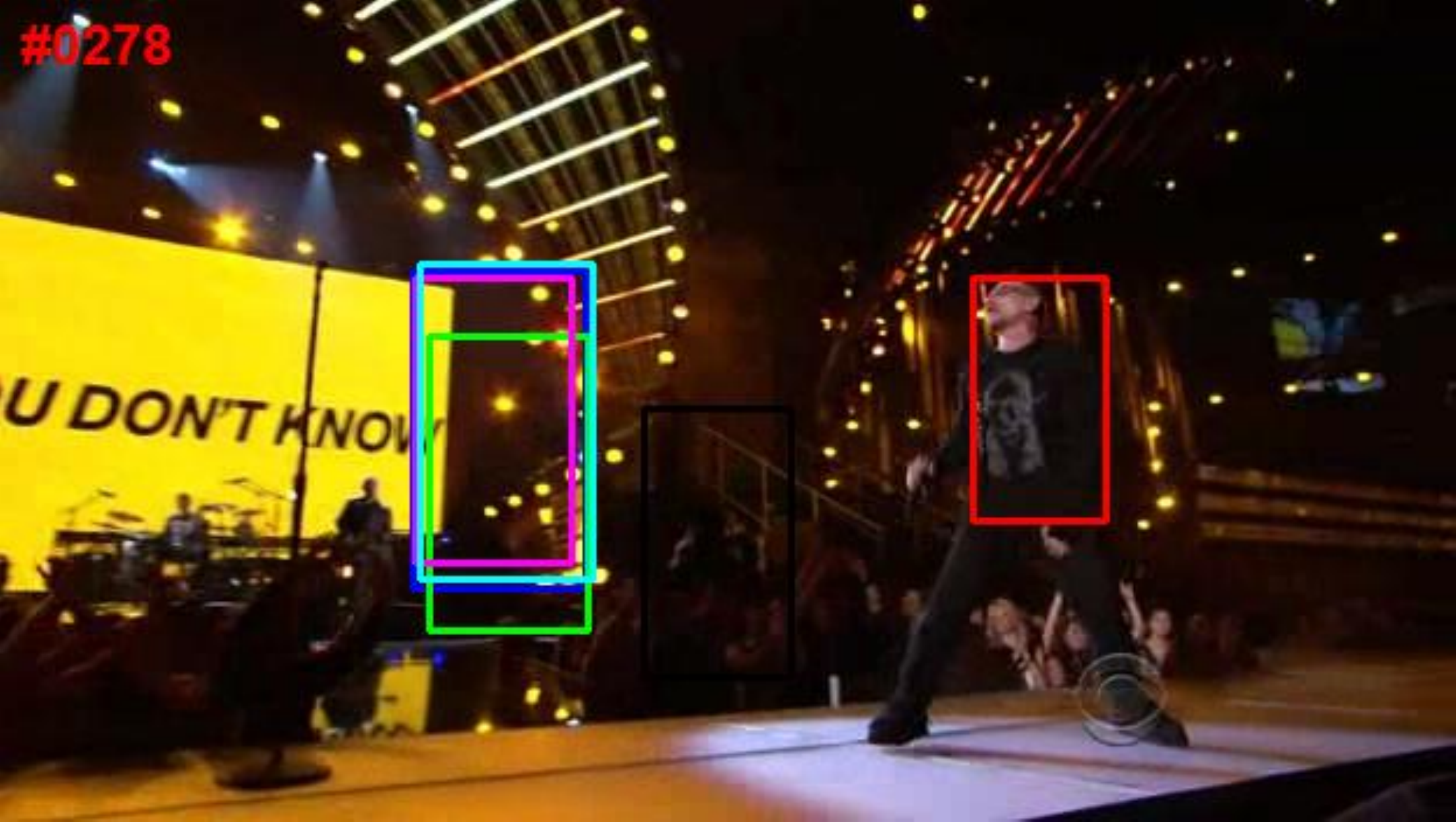}
\\
\end{tabular}

(d) \emph{singer1} and \emph{singer2} with light change.
\begin{tabular}{c@{}c@{}c@{}c@{}c@{}c}
\includegraphics[width=0.15\linewidth, height=0.08\linewidth]{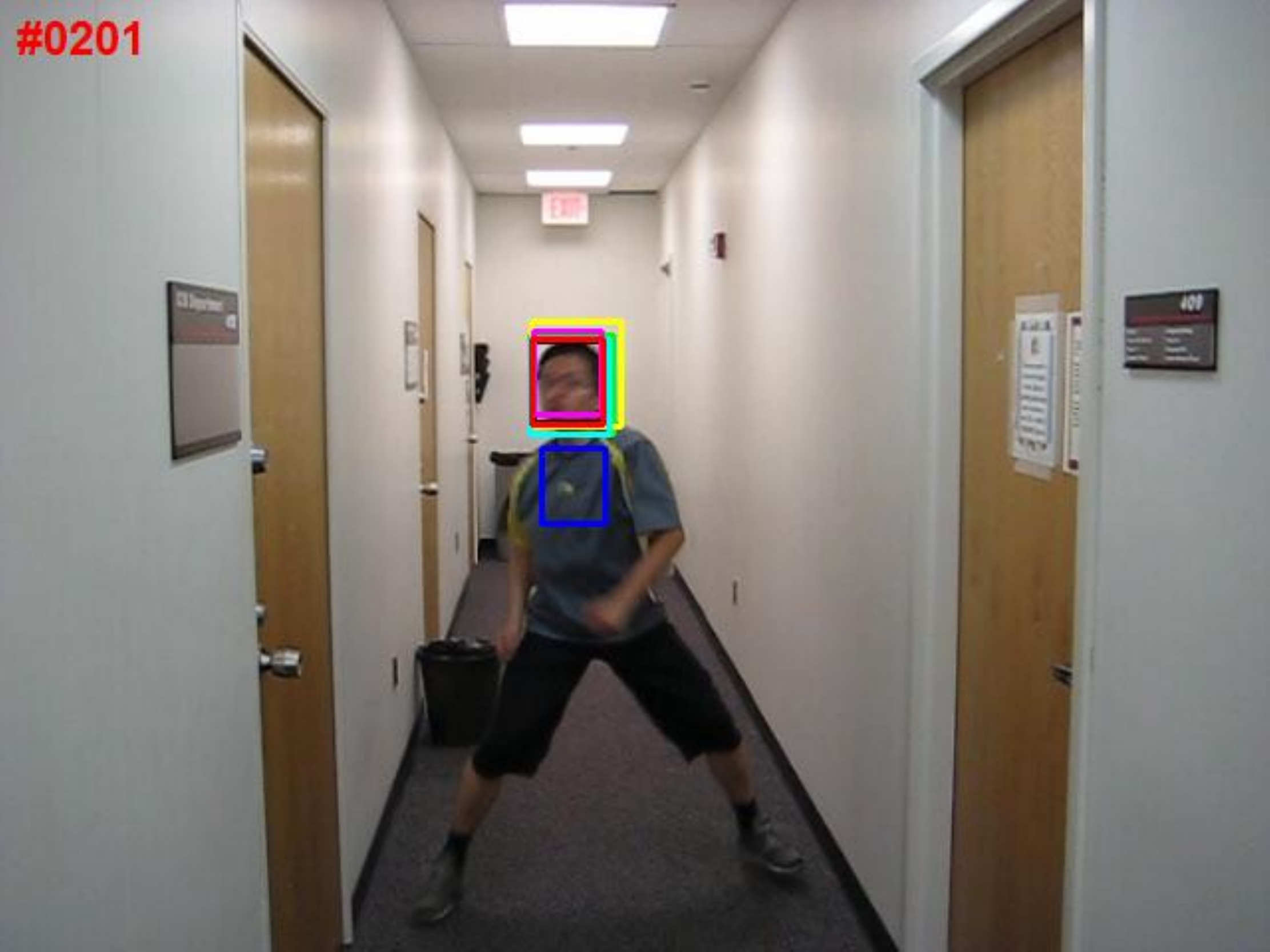}
&
\includegraphics[width=0.15\linewidth, height=0.08\linewidth]{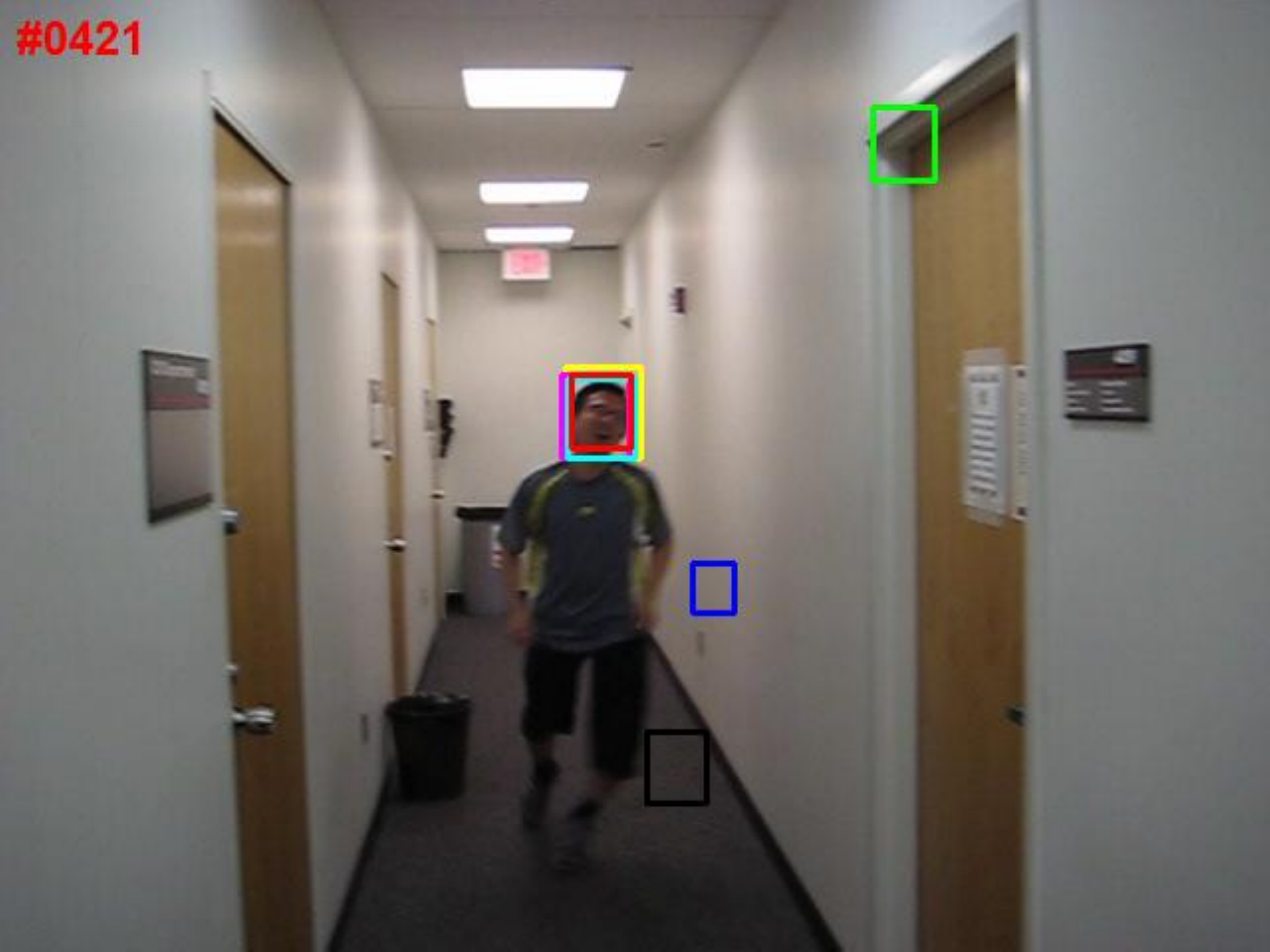}
&
\includegraphics[width=0.15\linewidth, height=0.08\linewidth]{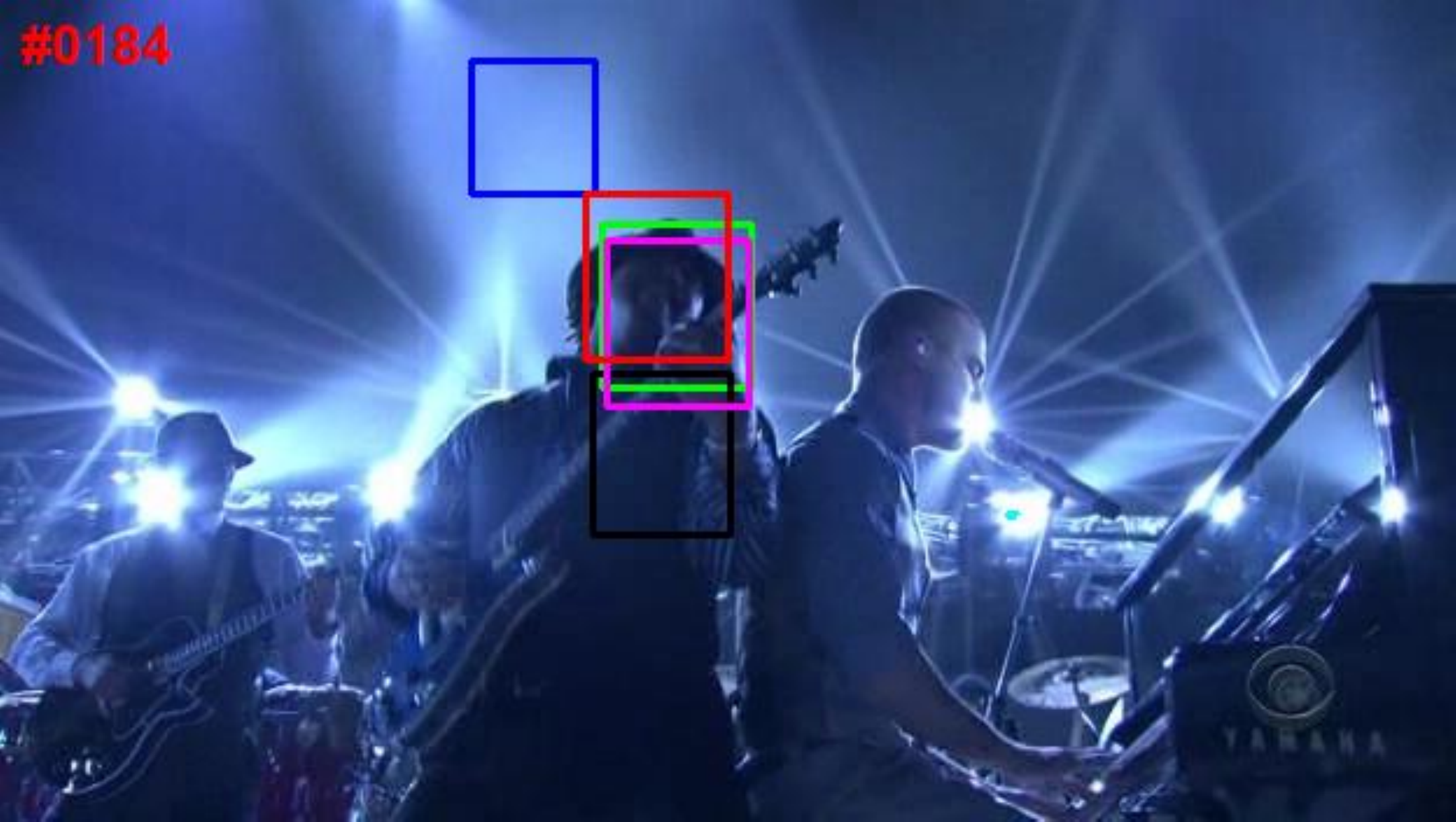}
&
\includegraphics[width=0.15\linewidth, height=0.08\linewidth]{picture/shaking_rs_0315.pdf}
&
\includegraphics[width=0.15\linewidth, height=0.08\linewidth]{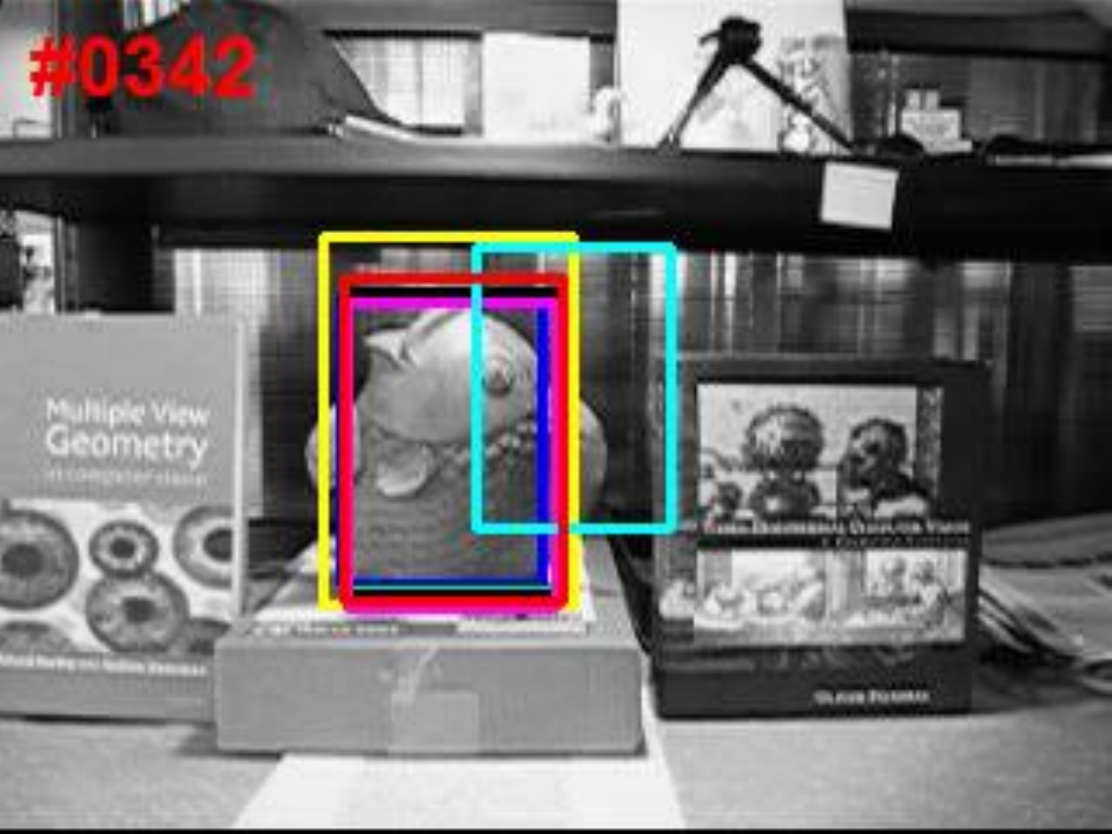}
&
\includegraphics[width=0.15\linewidth, height=0.08\linewidth]{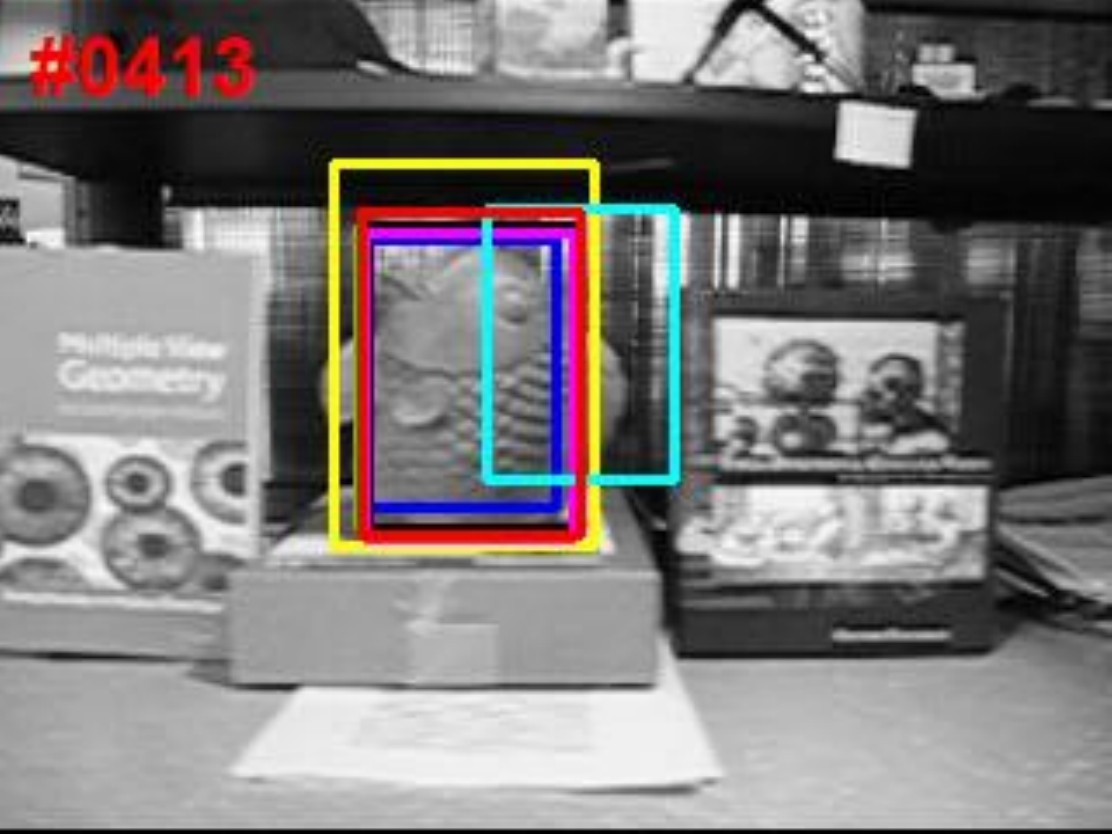}
\\
\end{tabular}

(e) \emph{boy}, \emph{shaking} and \emph{fish} with motion blur.
\begin{tabular} {c@{}c@{}c@{}c@{}c@{}c}
\includegraphics[width=0.15\linewidth, height=0.08\linewidth]{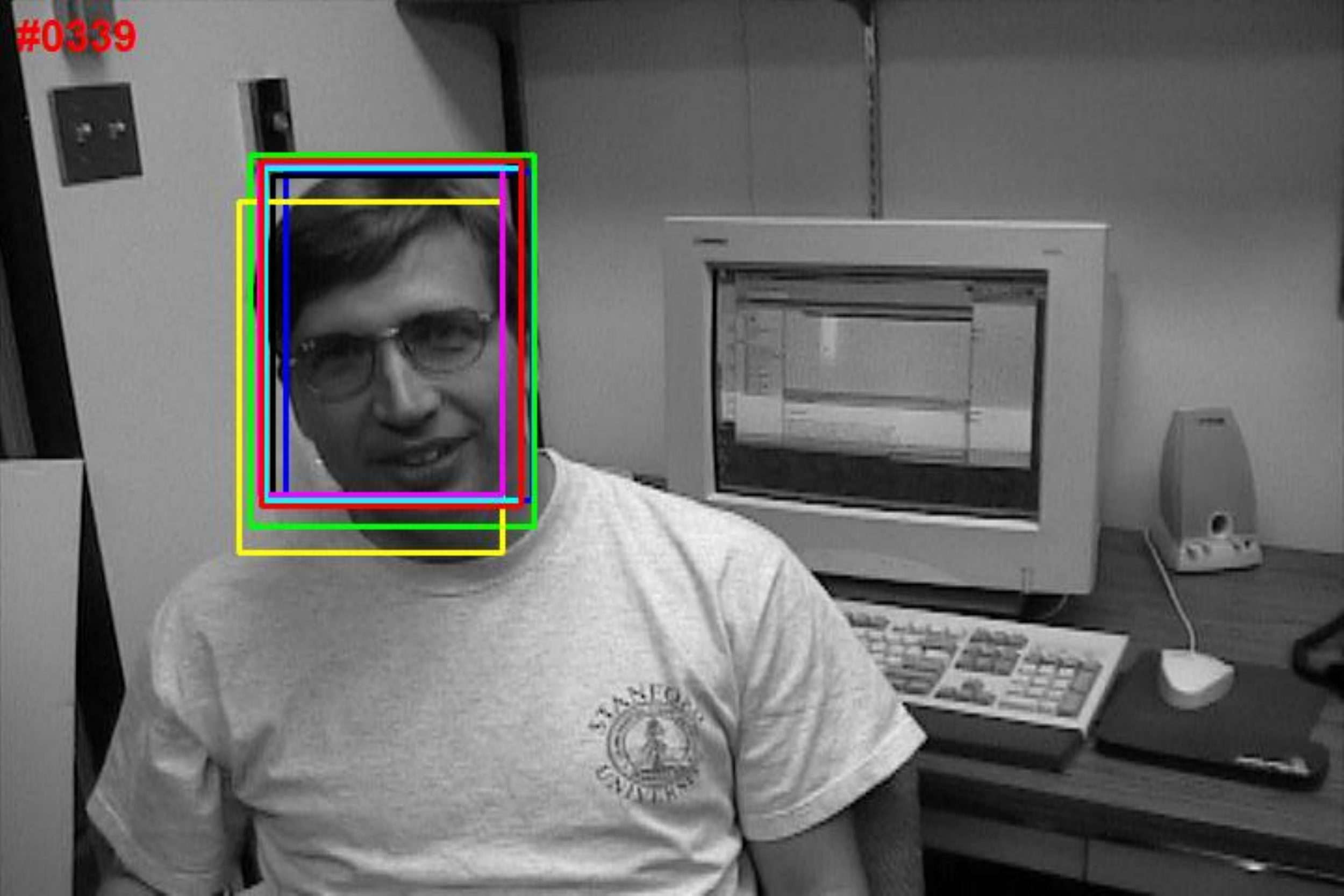}
&
\includegraphics[width=0.15\linewidth, height=0.08\linewidth]{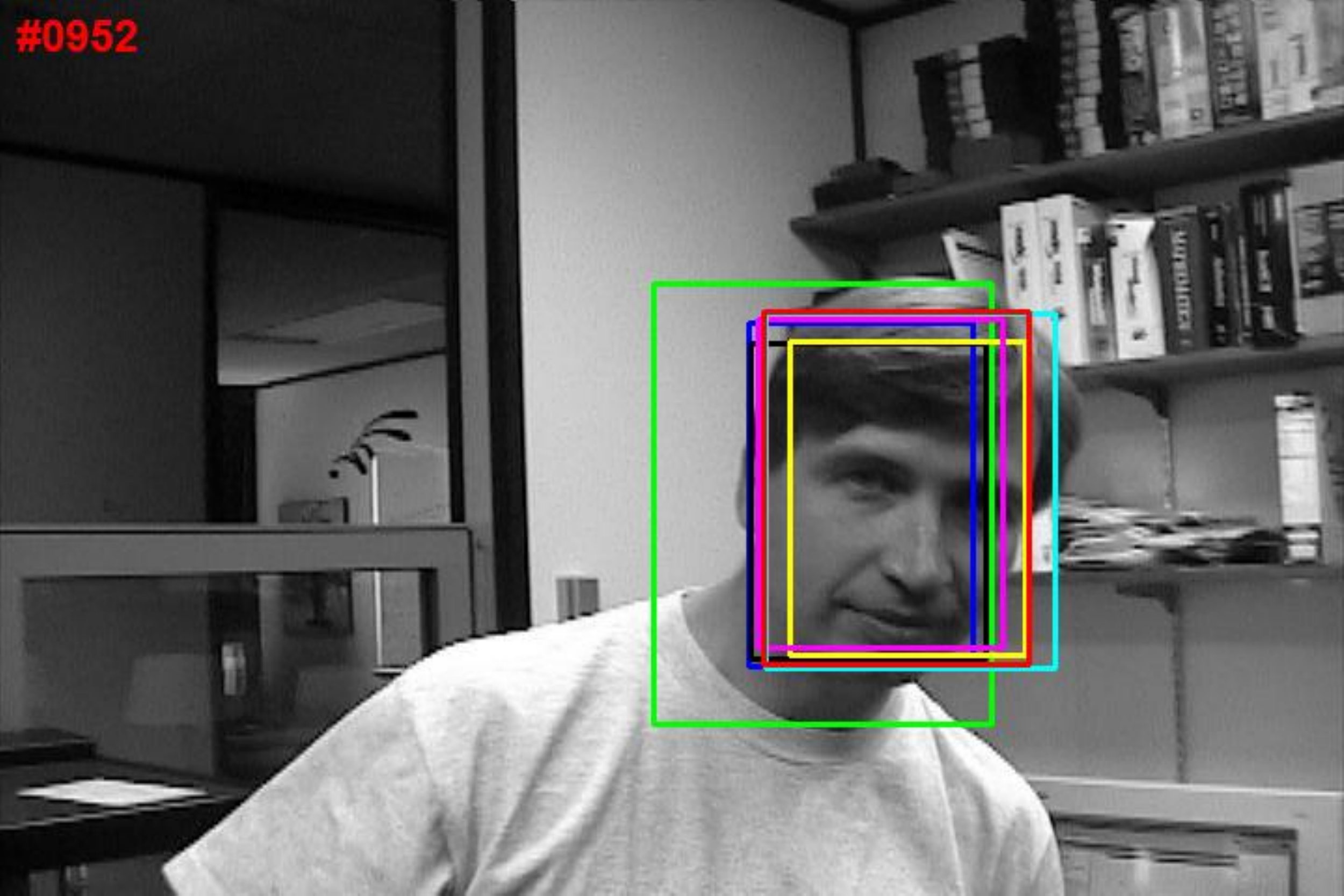}
&
\includegraphics[width=0.15\linewidth, height=0.08\linewidth]{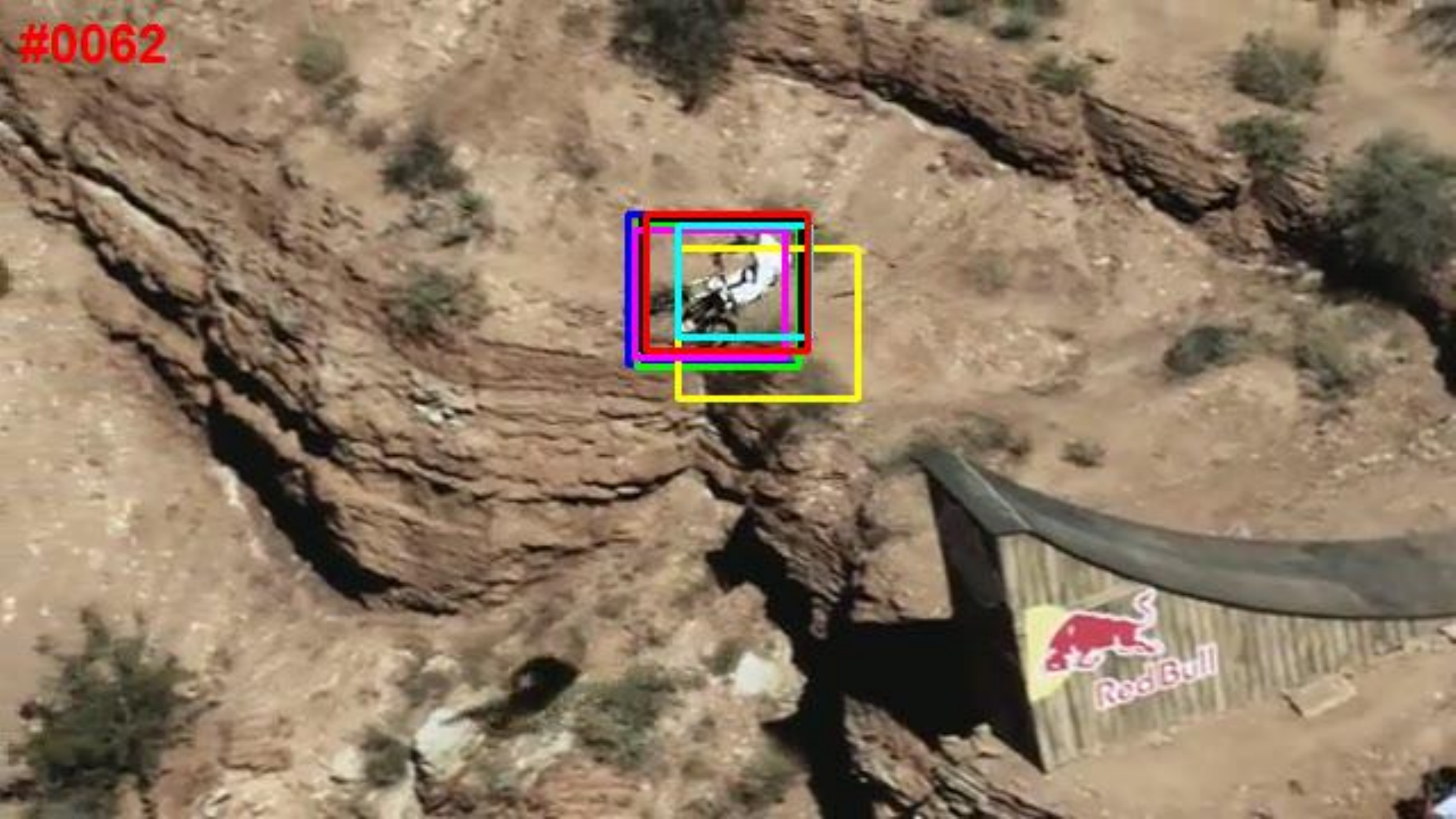}
&
\includegraphics[width=0.15\linewidth, height=0.08\linewidth]{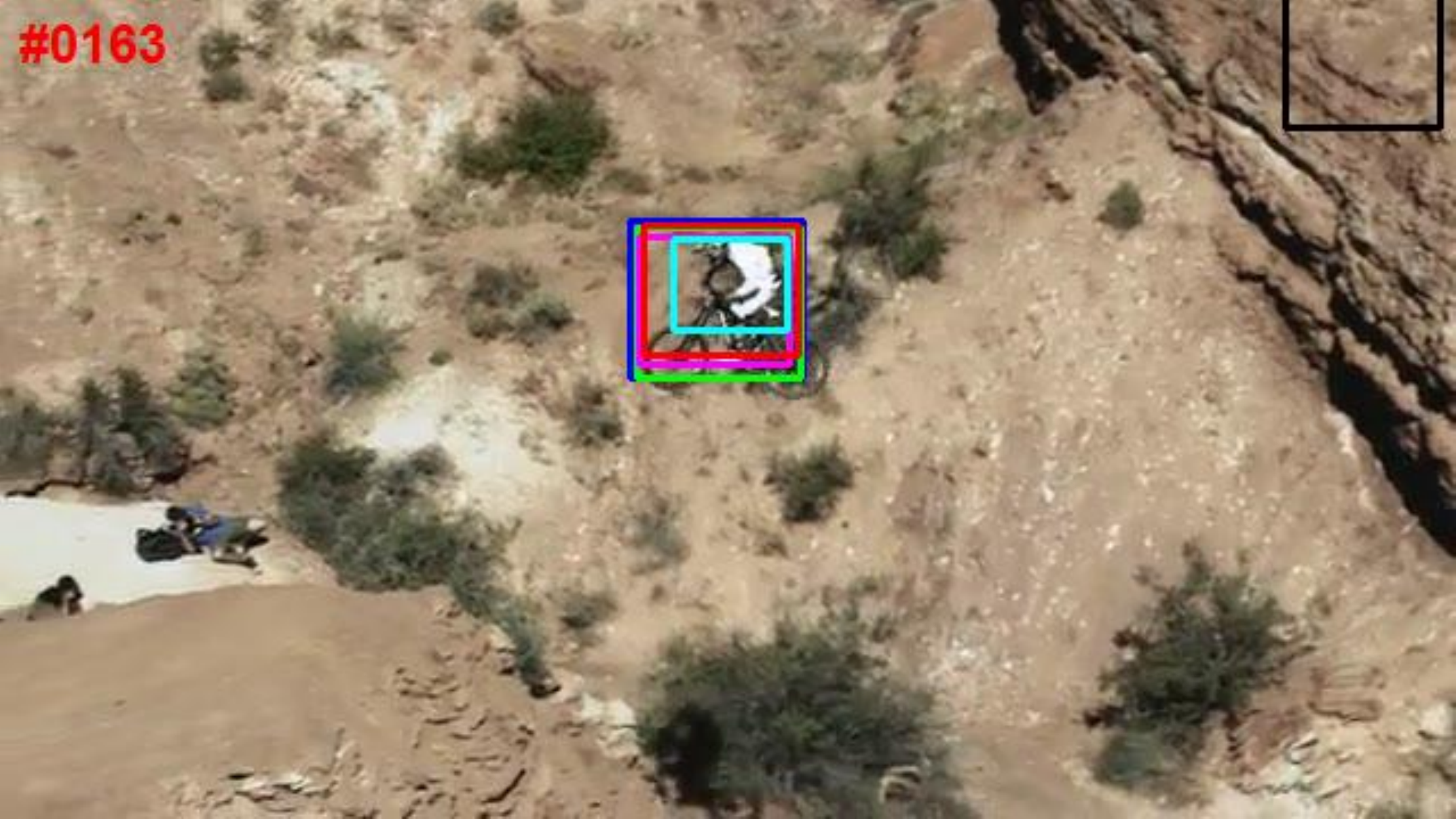}
&
\includegraphics[width=0.15\linewidth, height=0.08\linewidth]{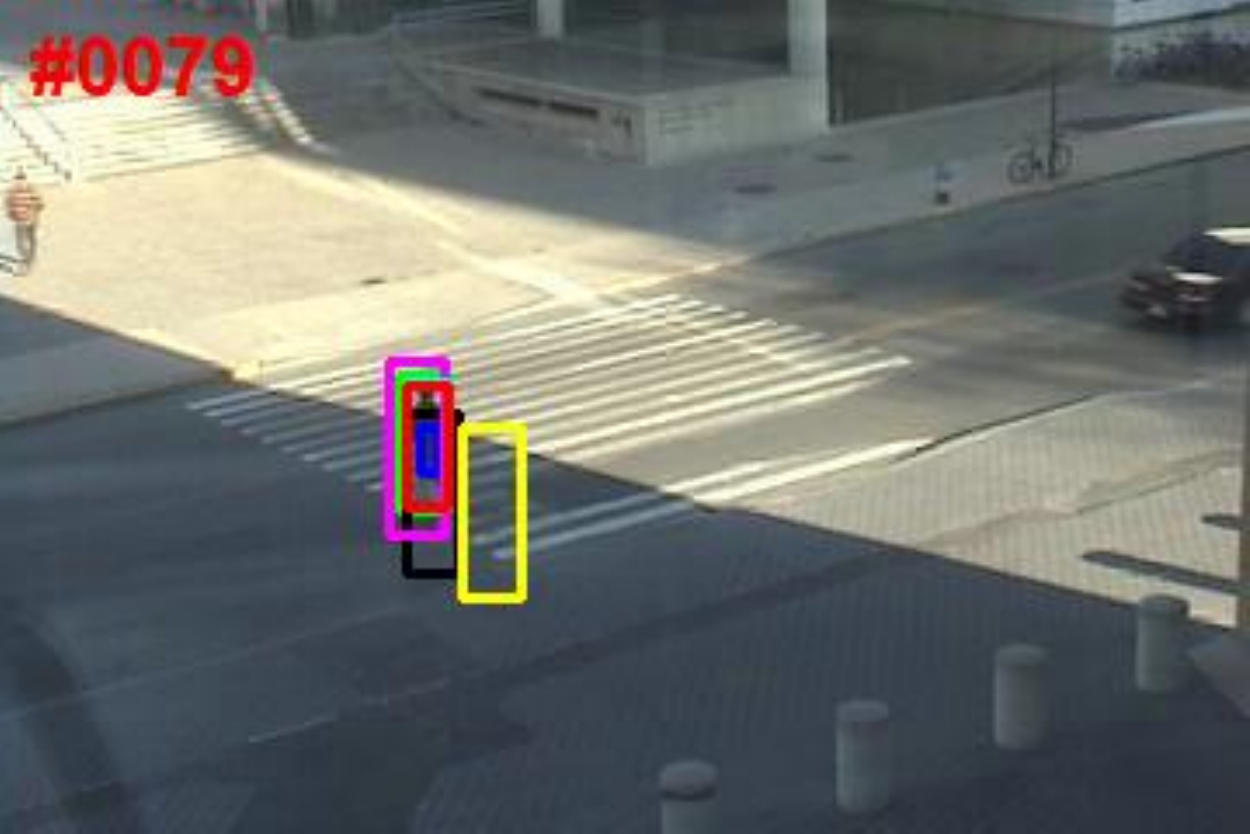}
&
\includegraphics[width=0.15\linewidth, height=0.08\linewidth]{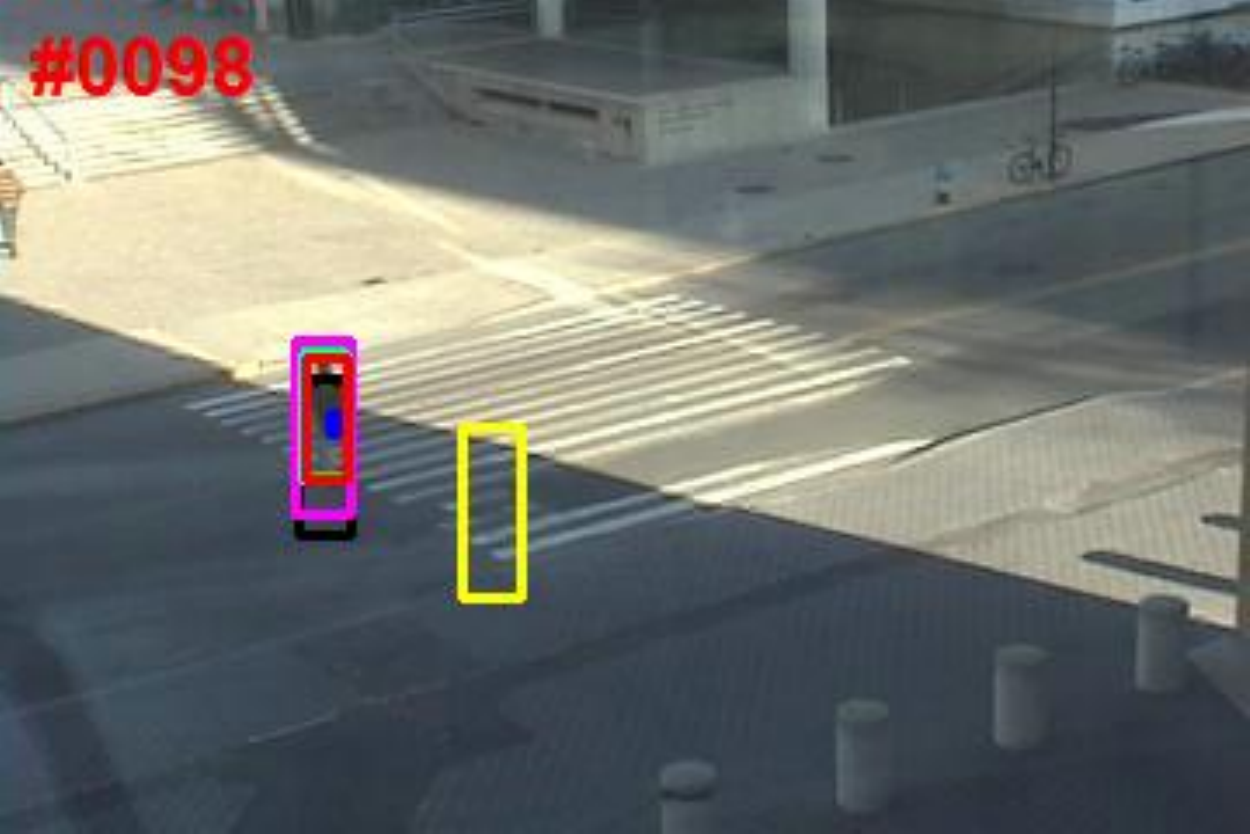}
\\
\end{tabular}

(f) \emph{dudek} , \emph{mountainBike} and \emph{crossing} with cluttered background.
\begin{tabular}{c@{}c@{}c@{}c@{}c@{}c}
\includegraphics[width=0.15\linewidth, height=0.08\linewidth]{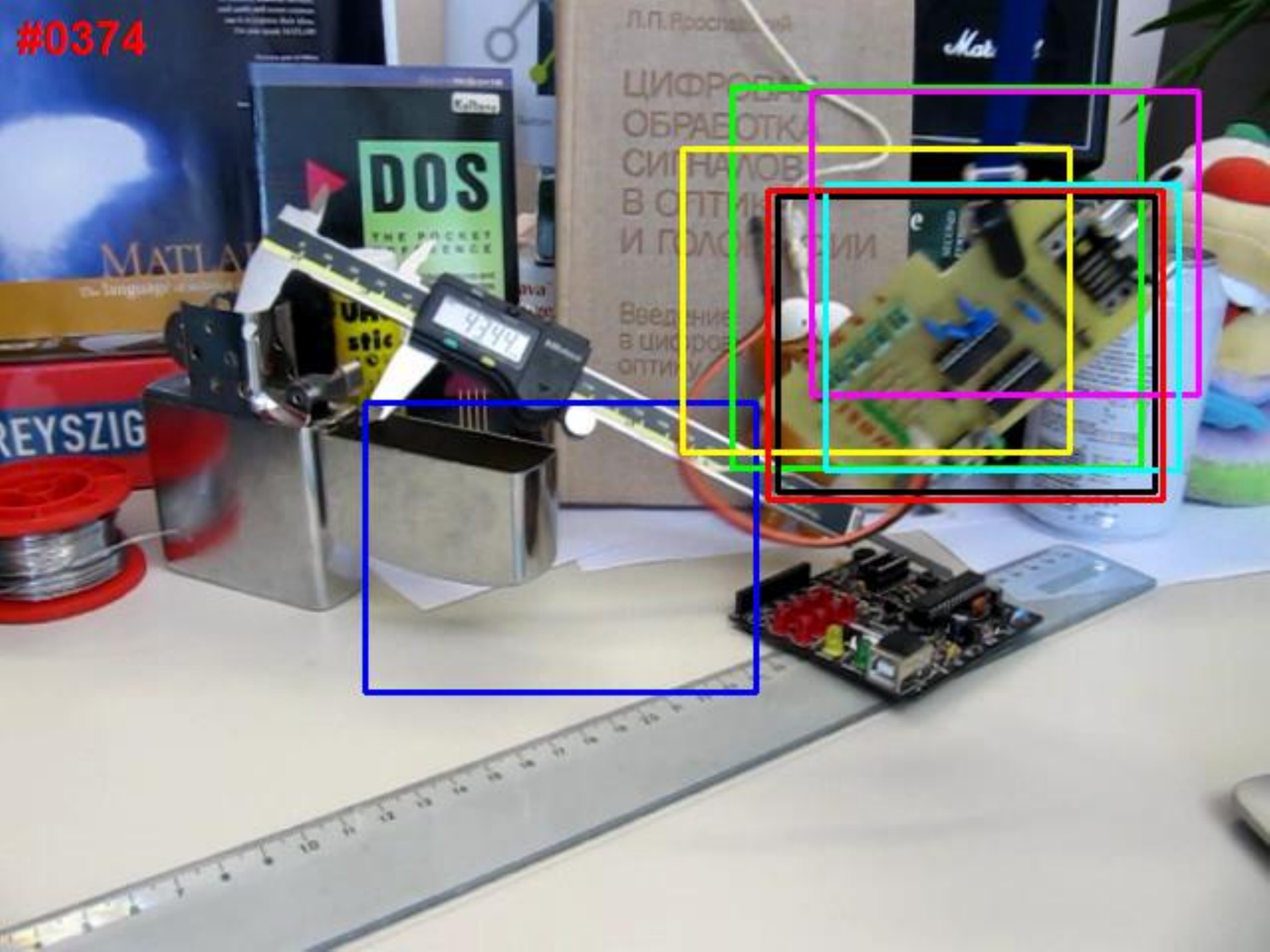}
&
\includegraphics[width=0.15\linewidth, height=0.08\linewidth]{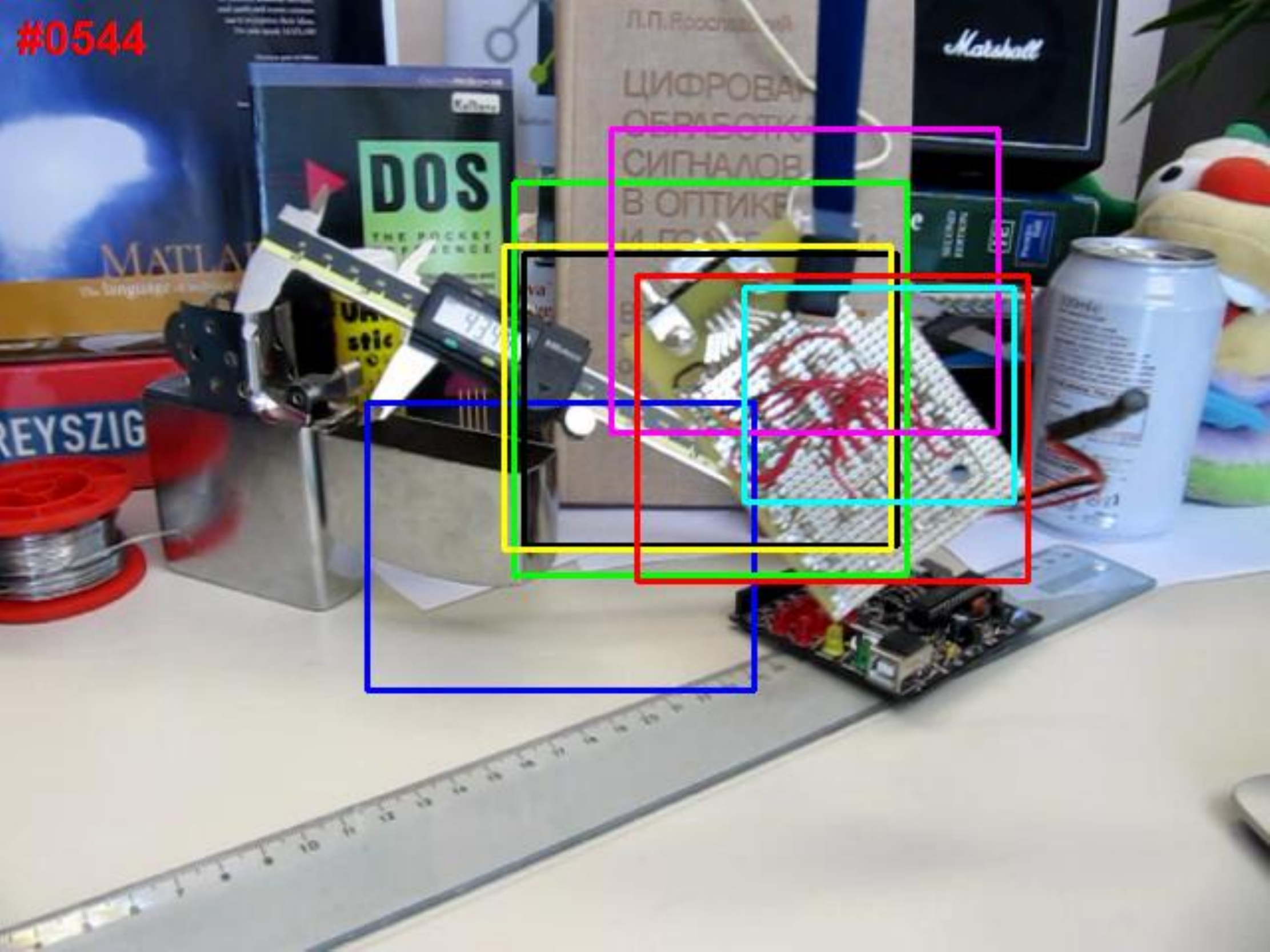}
&
\includegraphics[width=0.15\linewidth, height=0.08\linewidth]{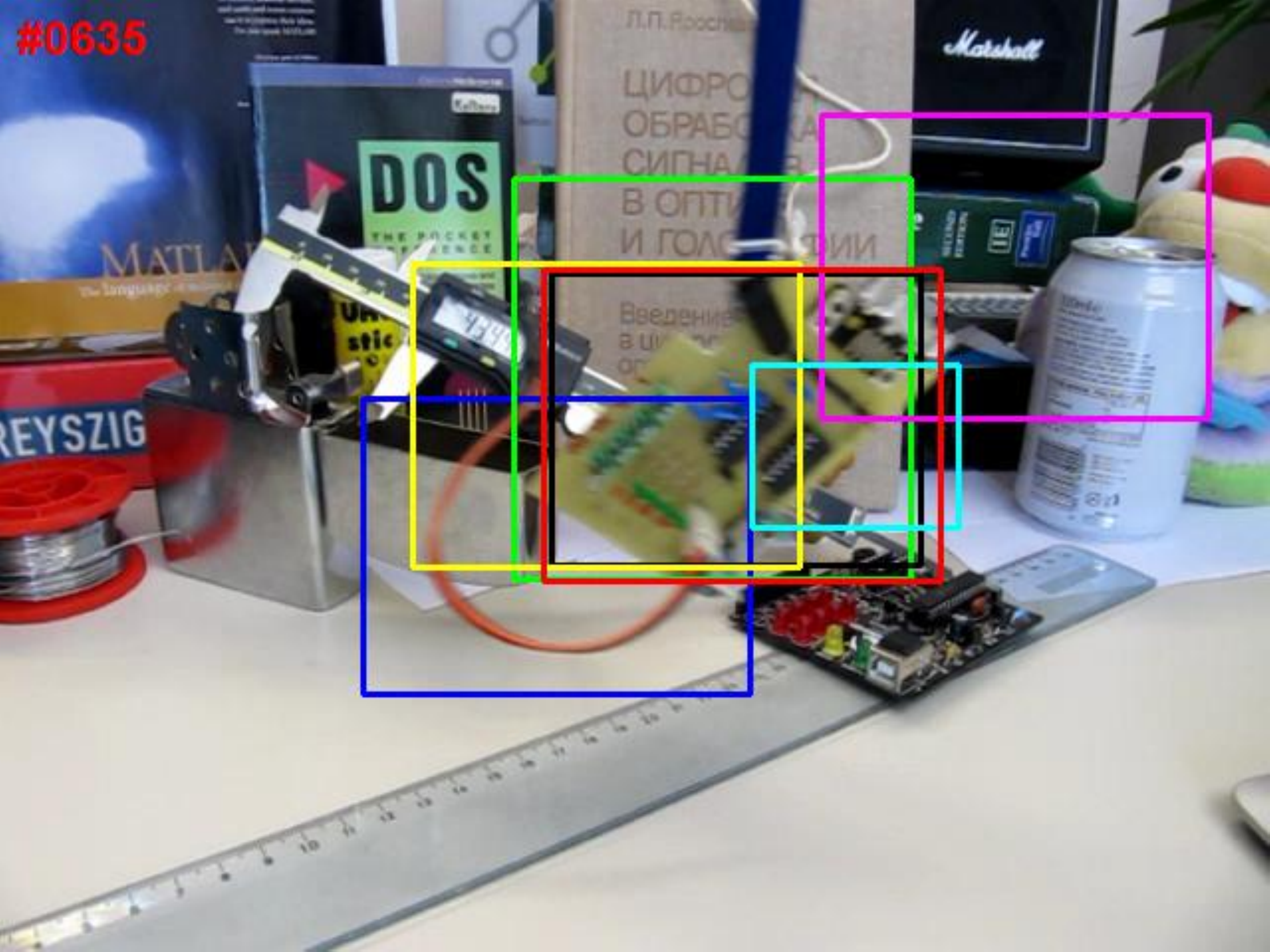}
&
\includegraphics[width=0.15\linewidth, height=0.08\linewidth]{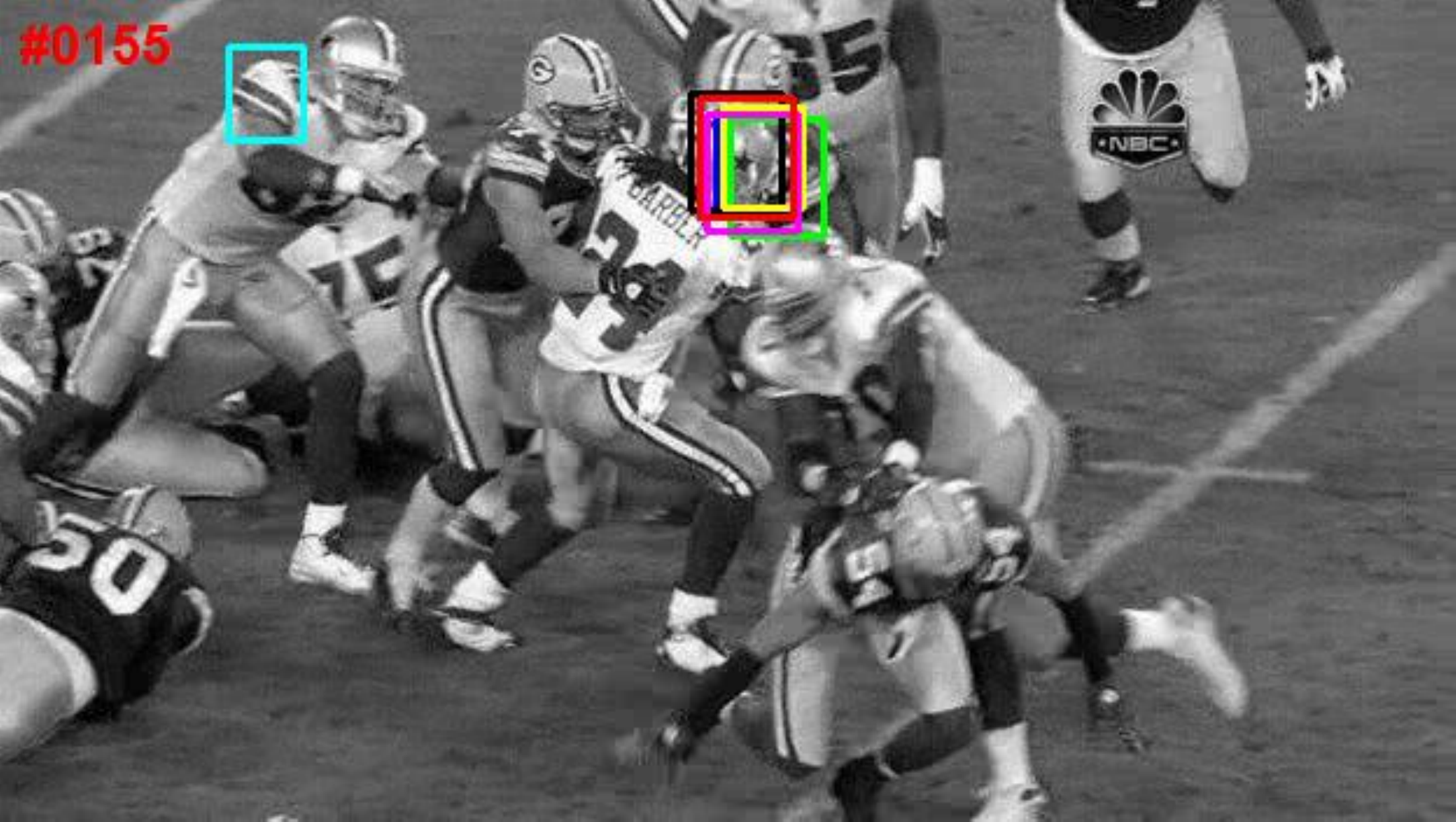}
&
\includegraphics[width=0.15\linewidth, height=0.08\linewidth]{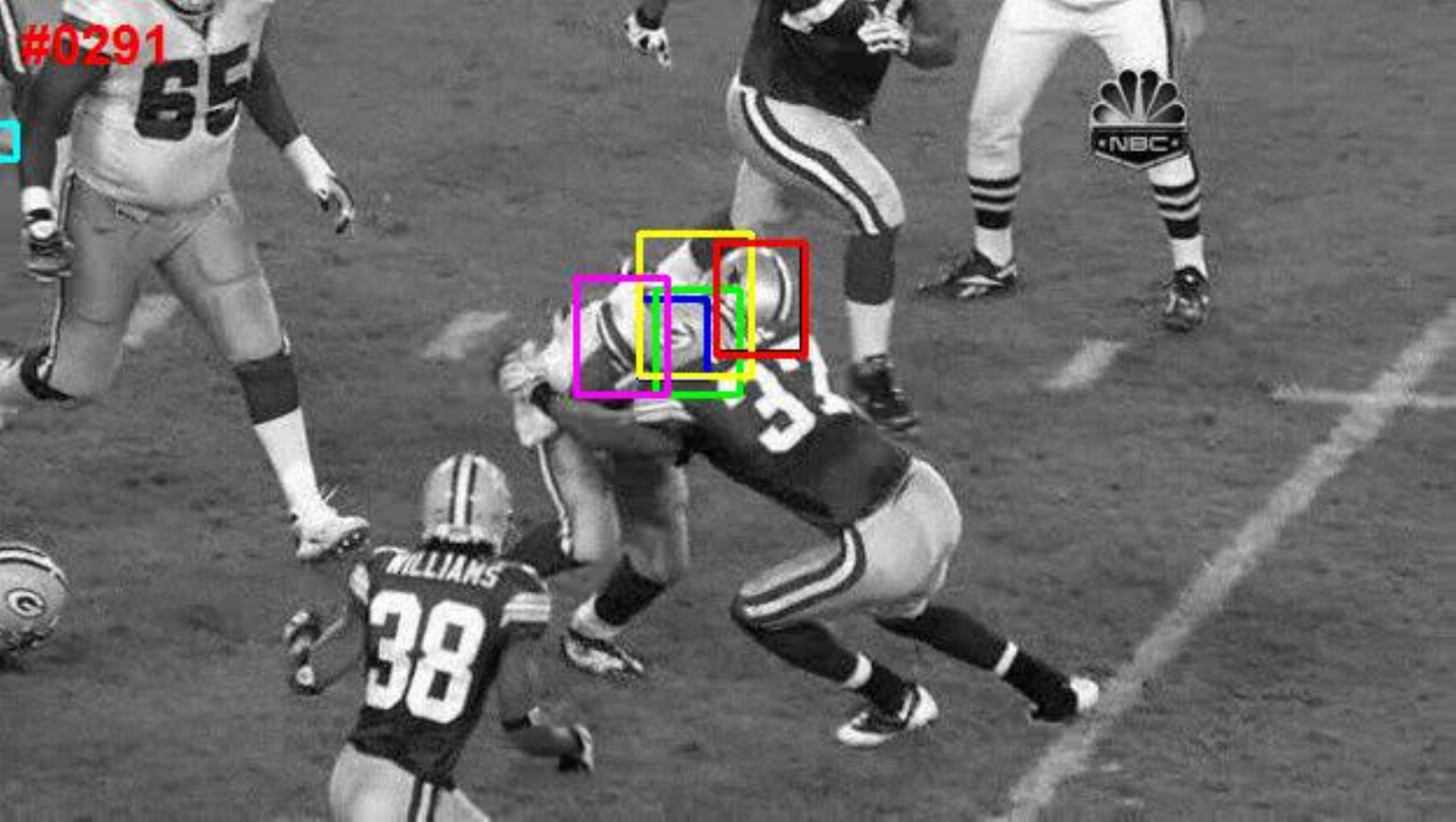}
&
\includegraphics[width=0.15\linewidth, height=0.08\linewidth]{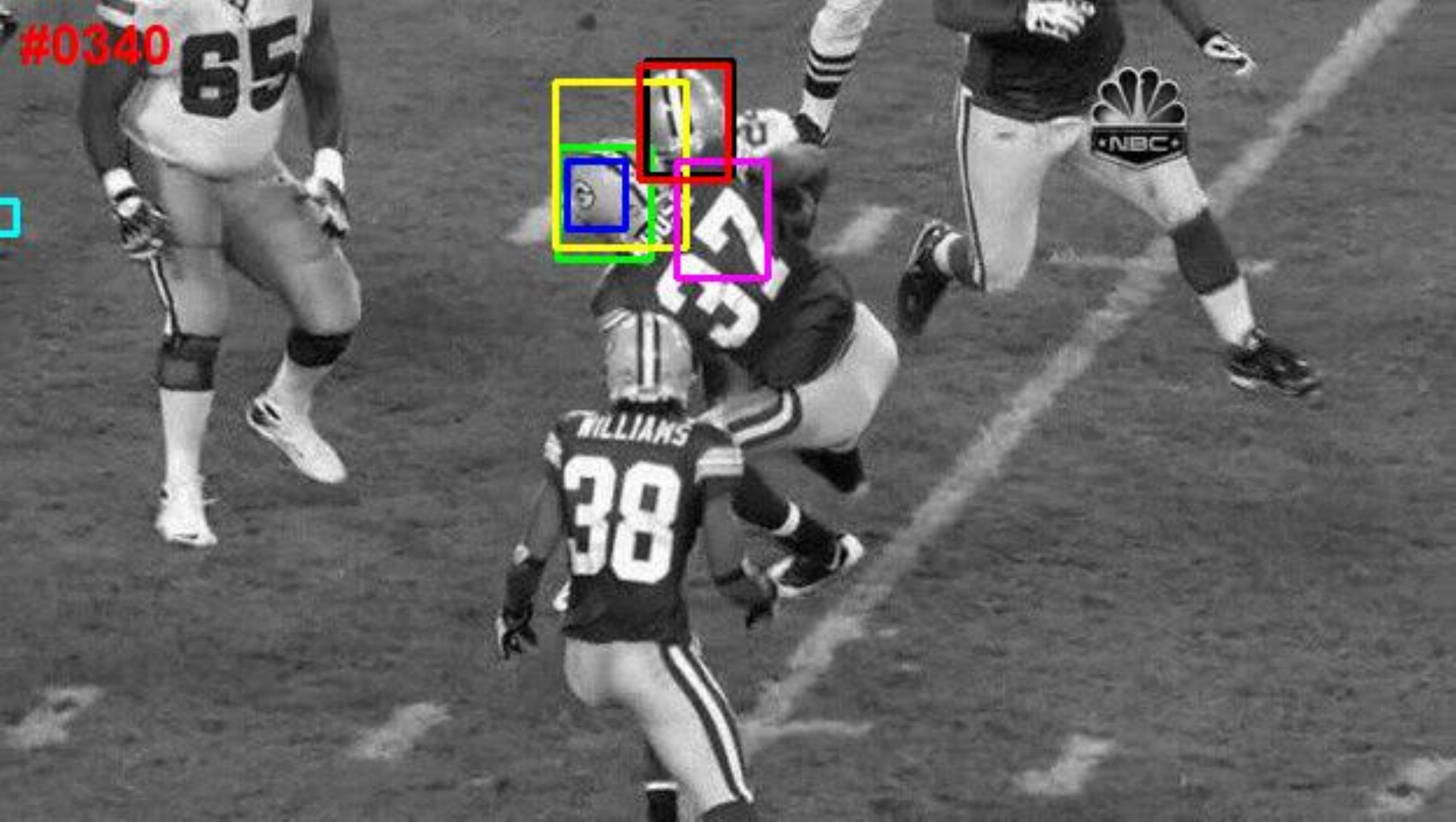}
\\
\end{tabular}

(g) \emph{board} and \emph{football}  with background clutter.
\begin{tabular}{c@{}c@{}c@{}c@{}c@{}c}
\includegraphics[width=0.15\linewidth, height=0.08\linewidth]{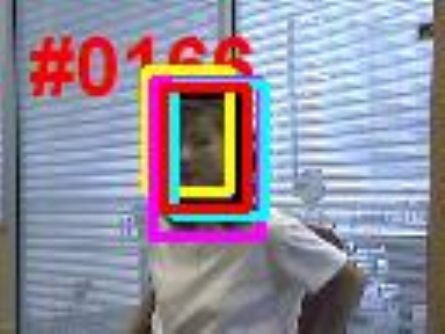}
&
\includegraphics[width=0.15\linewidth, height=0.08\linewidth]{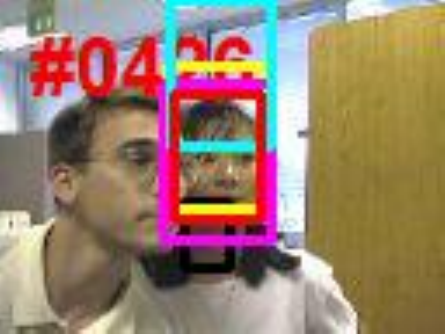}
&
\includegraphics[width=0.15\linewidth, height=0.08\linewidth]{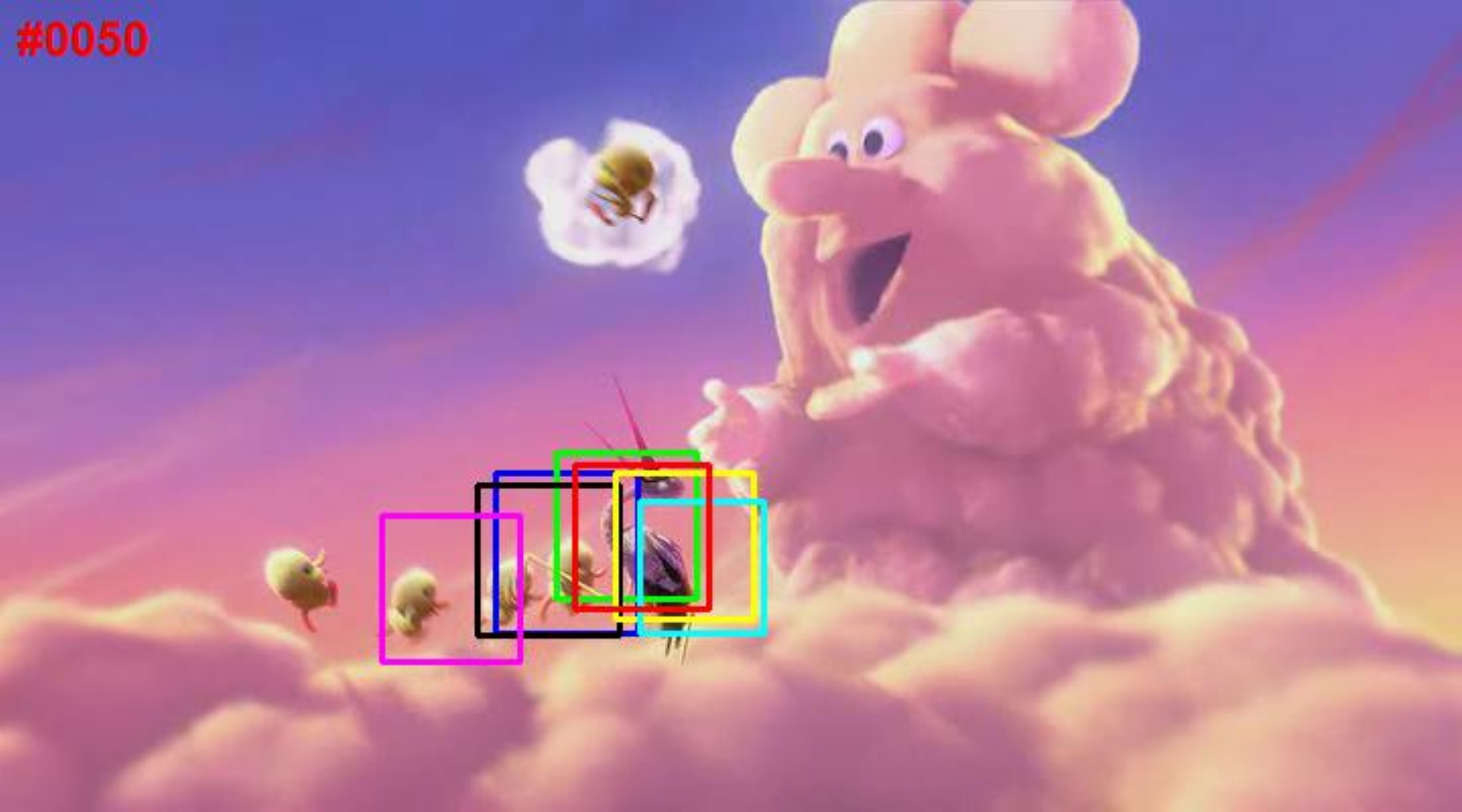}
&
\includegraphics[width=0.15\linewidth, height=0.08\linewidth]{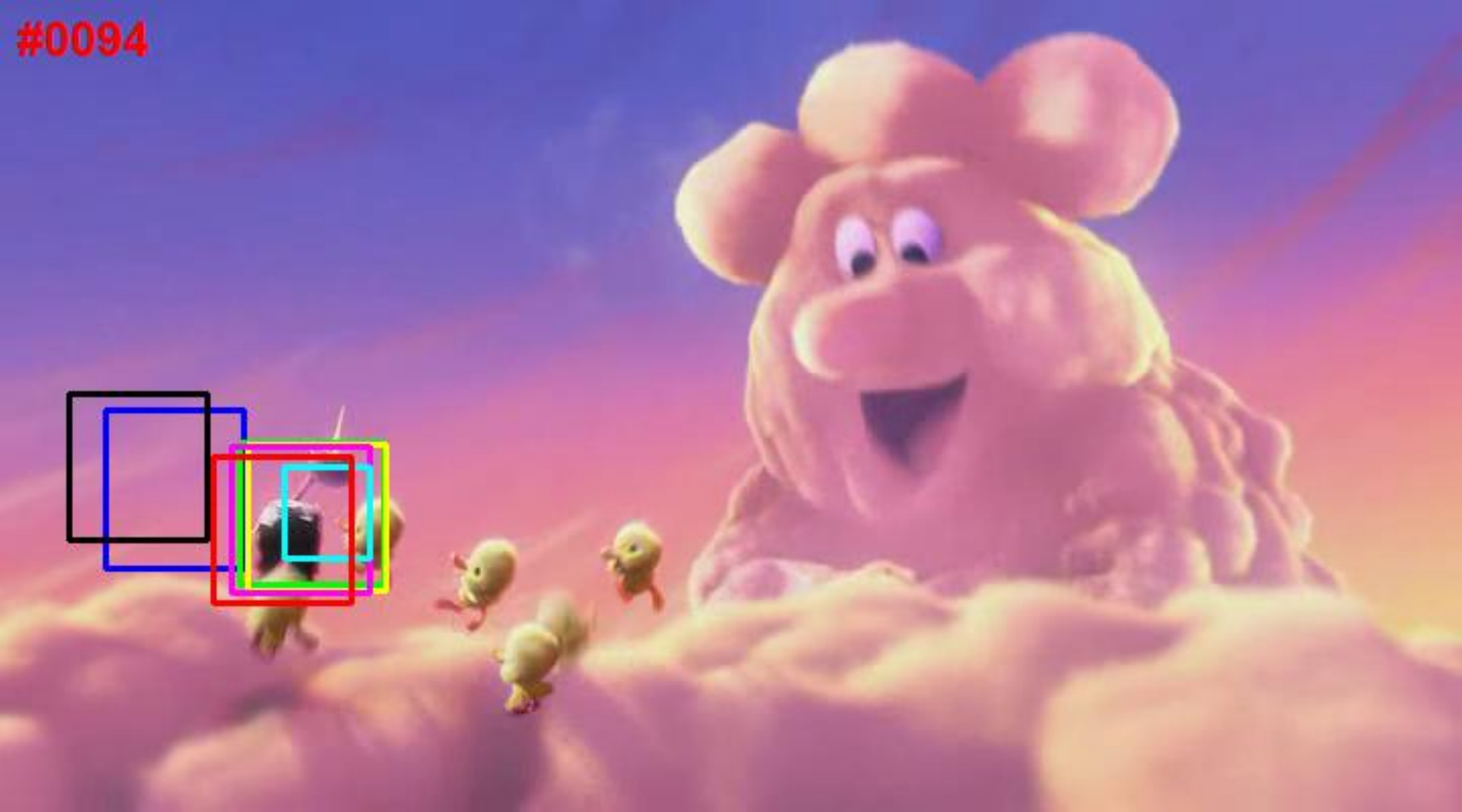}
&
\includegraphics[width=0.15\linewidth, height=0.08\linewidth]{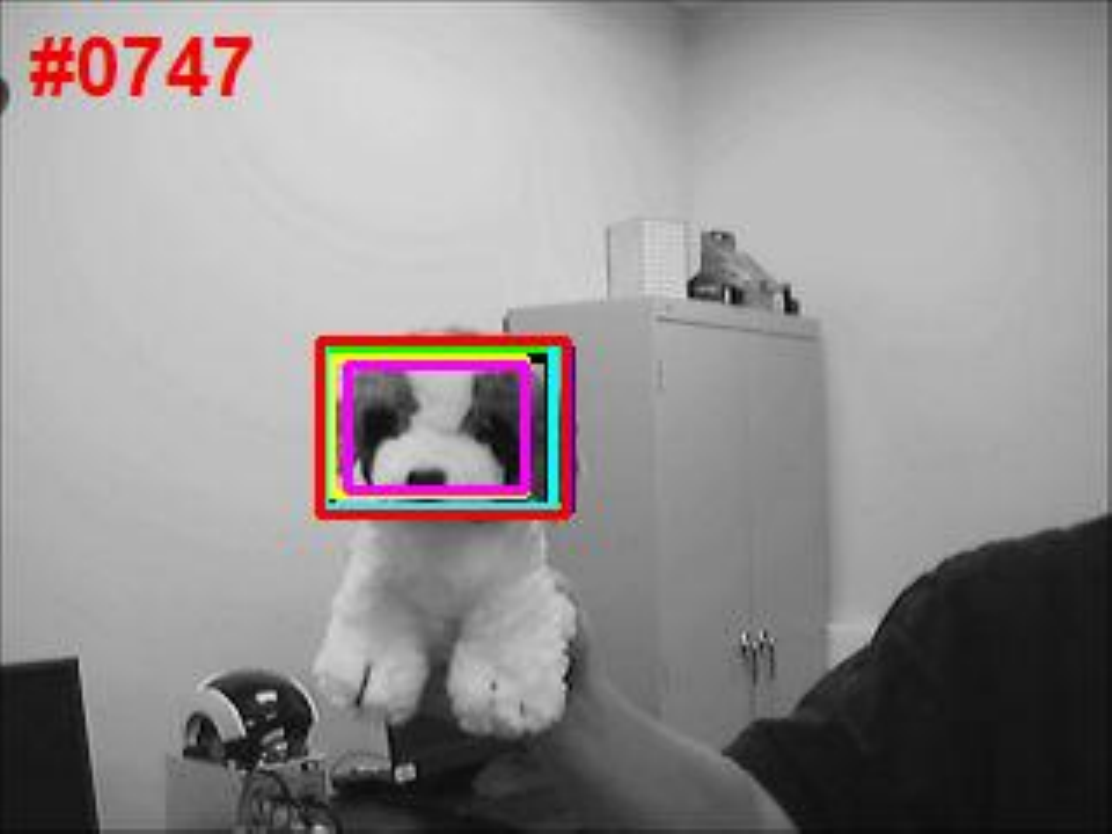}
&
\includegraphics[width=0.15\linewidth, height=0.08\linewidth]{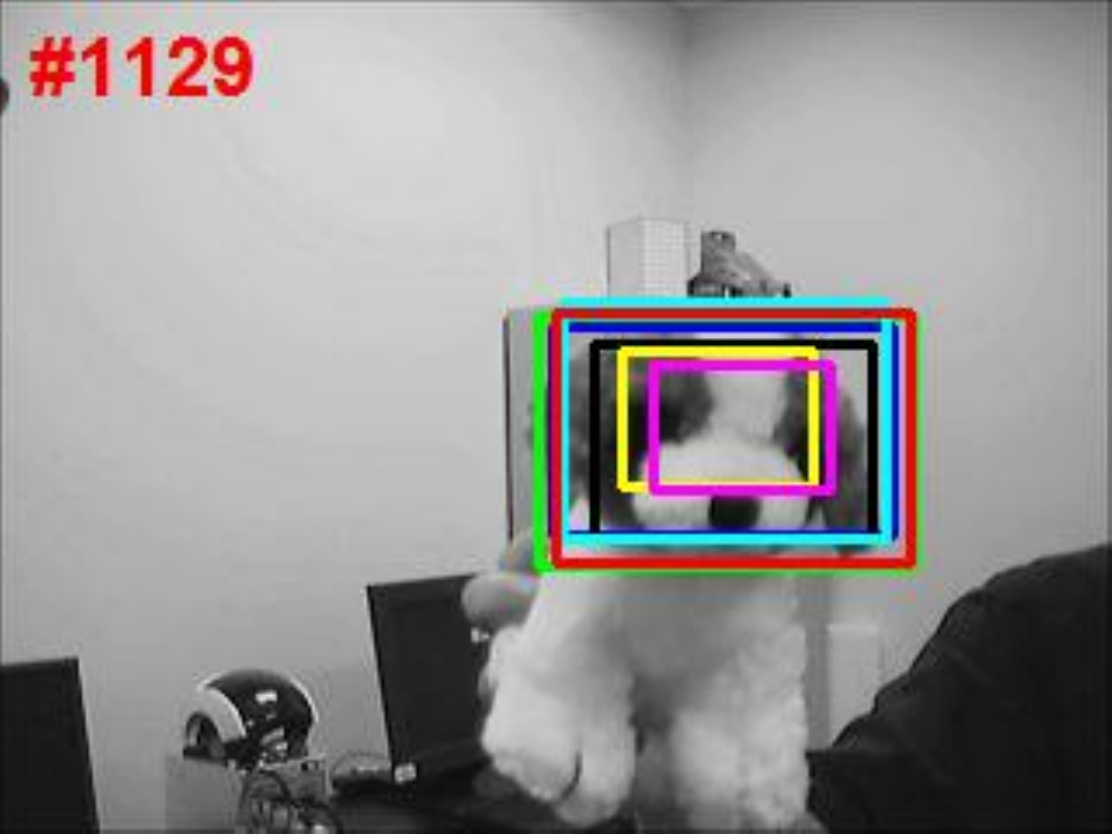}
\\
\end{tabular}

(h) \emph{girl}, \emph{bird} and \emph{dog1}  with in-plane or out-of-plane rotation.
\begin{tabular}{c@{}c@{}c@{}c@{}c@{}c}
\includegraphics[width=0.15\linewidth, height=0.08\linewidth]{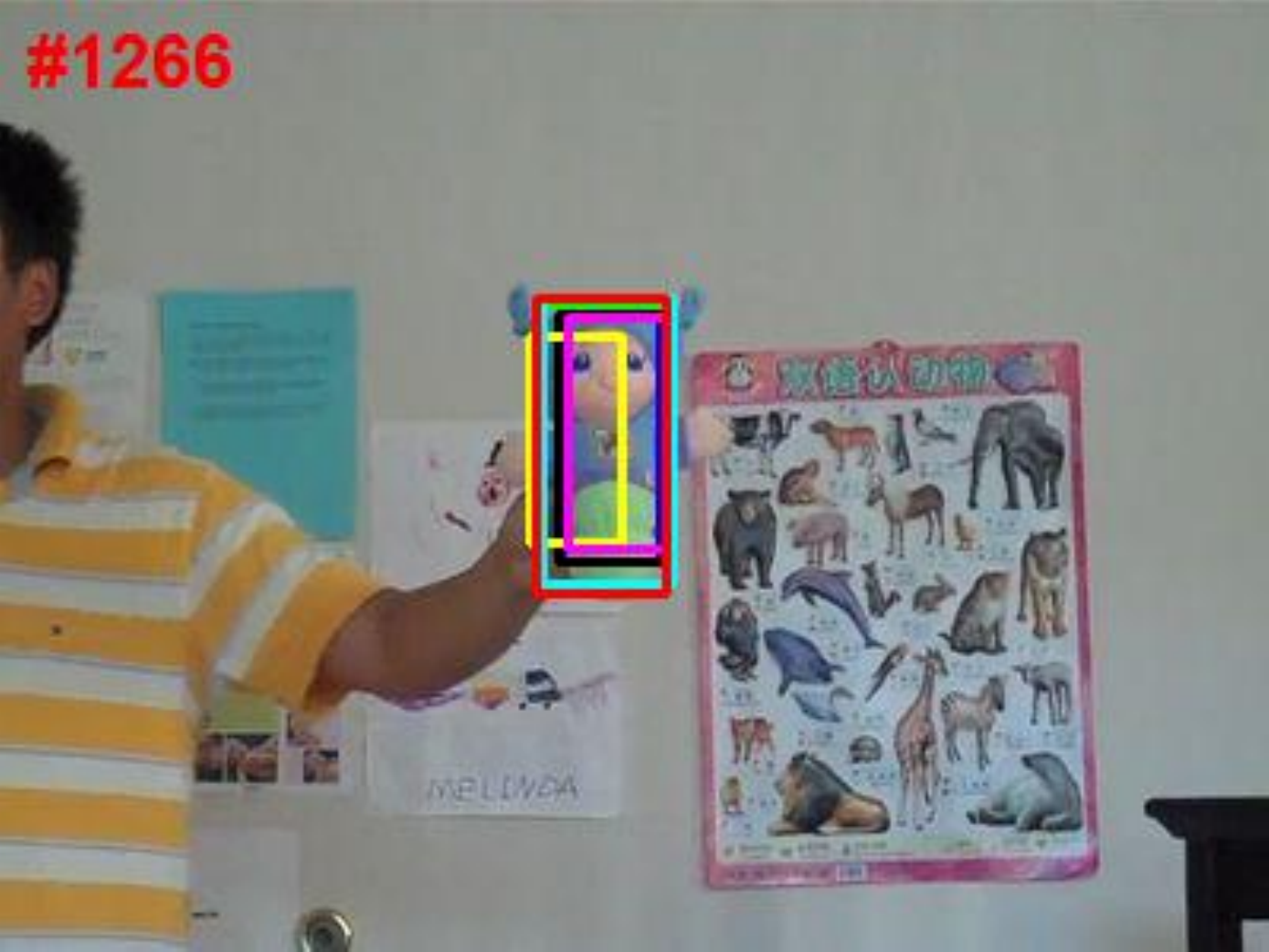}
&
\includegraphics[width=0.15\linewidth, height=0.08\linewidth]{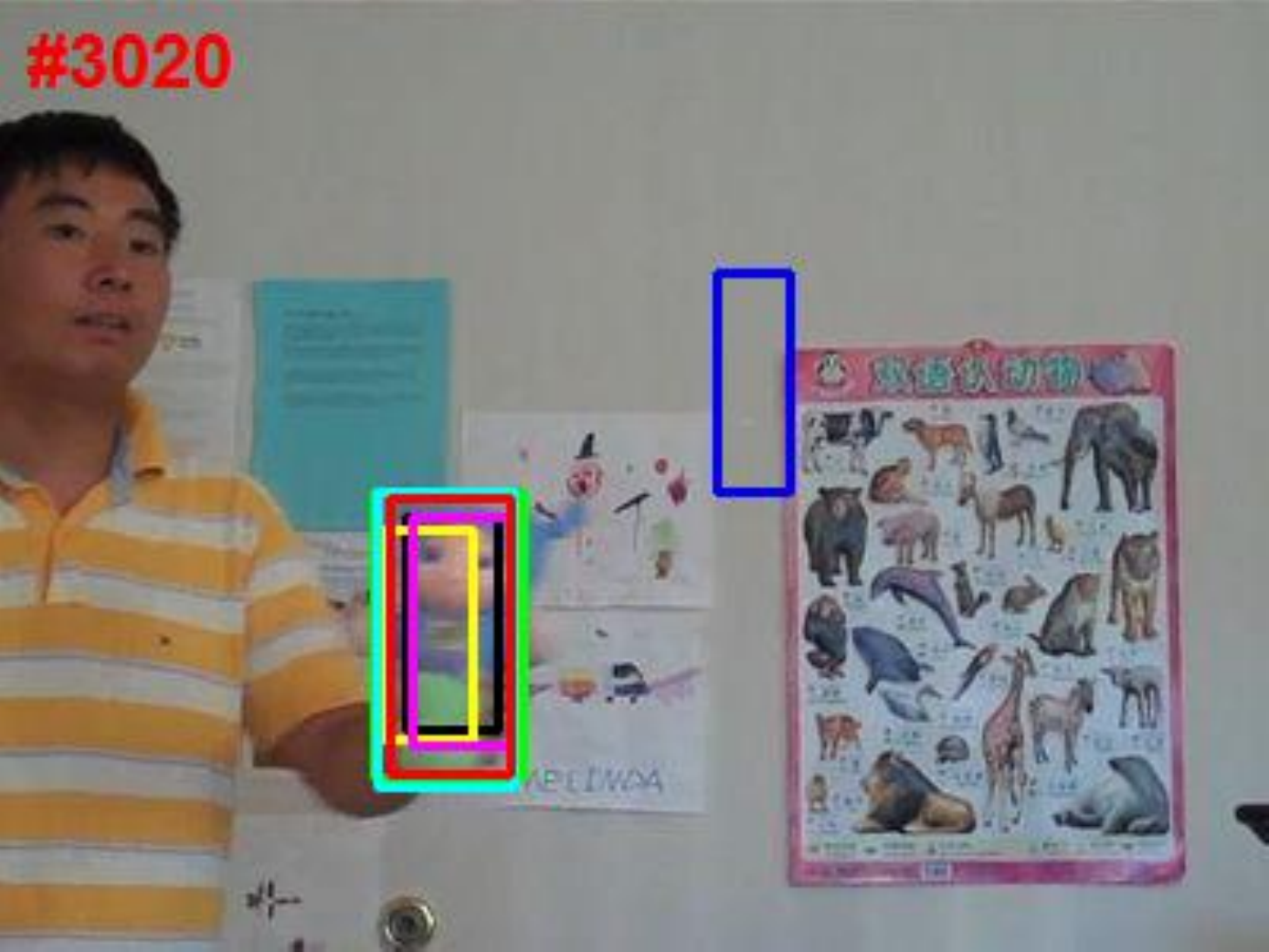}
&
\includegraphics[width=0.15\linewidth, height=0.08\linewidth]{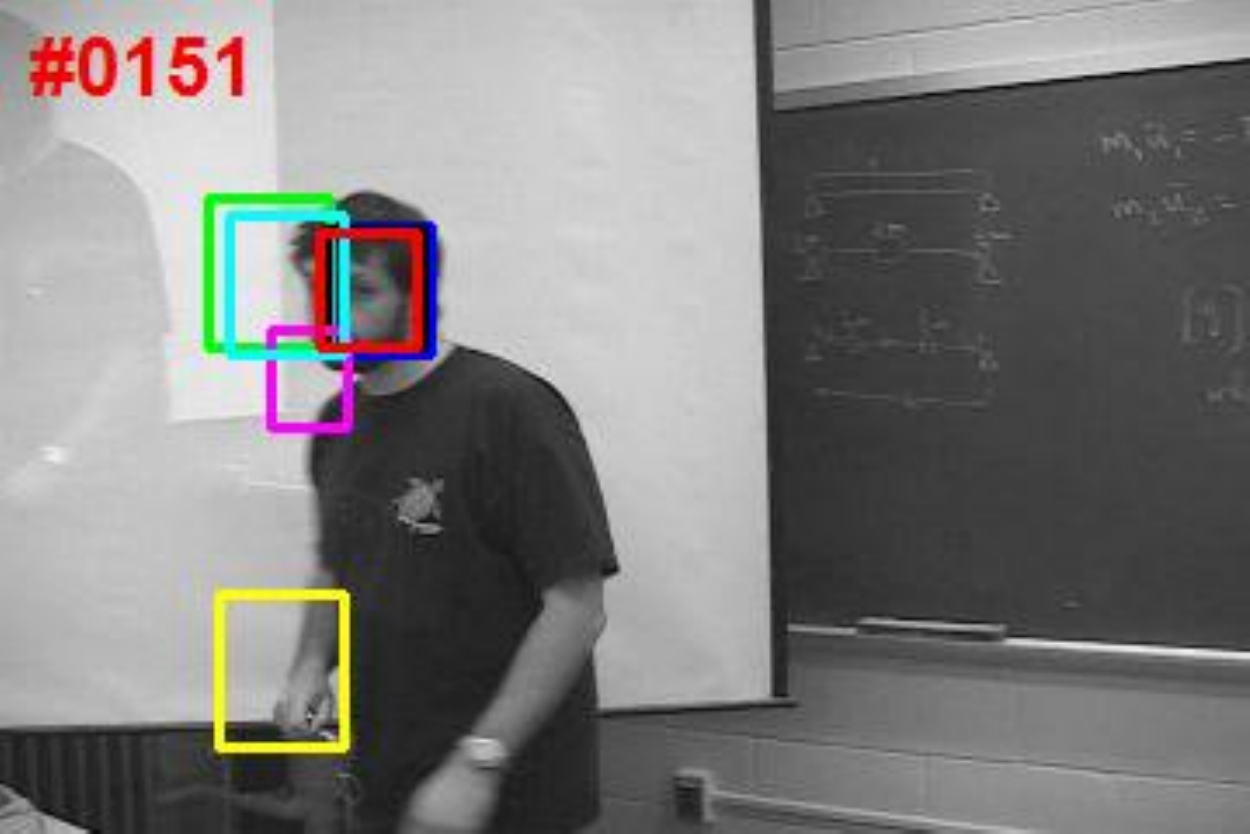}
&
\includegraphics[width=0.15\linewidth, height=0.08\linewidth]{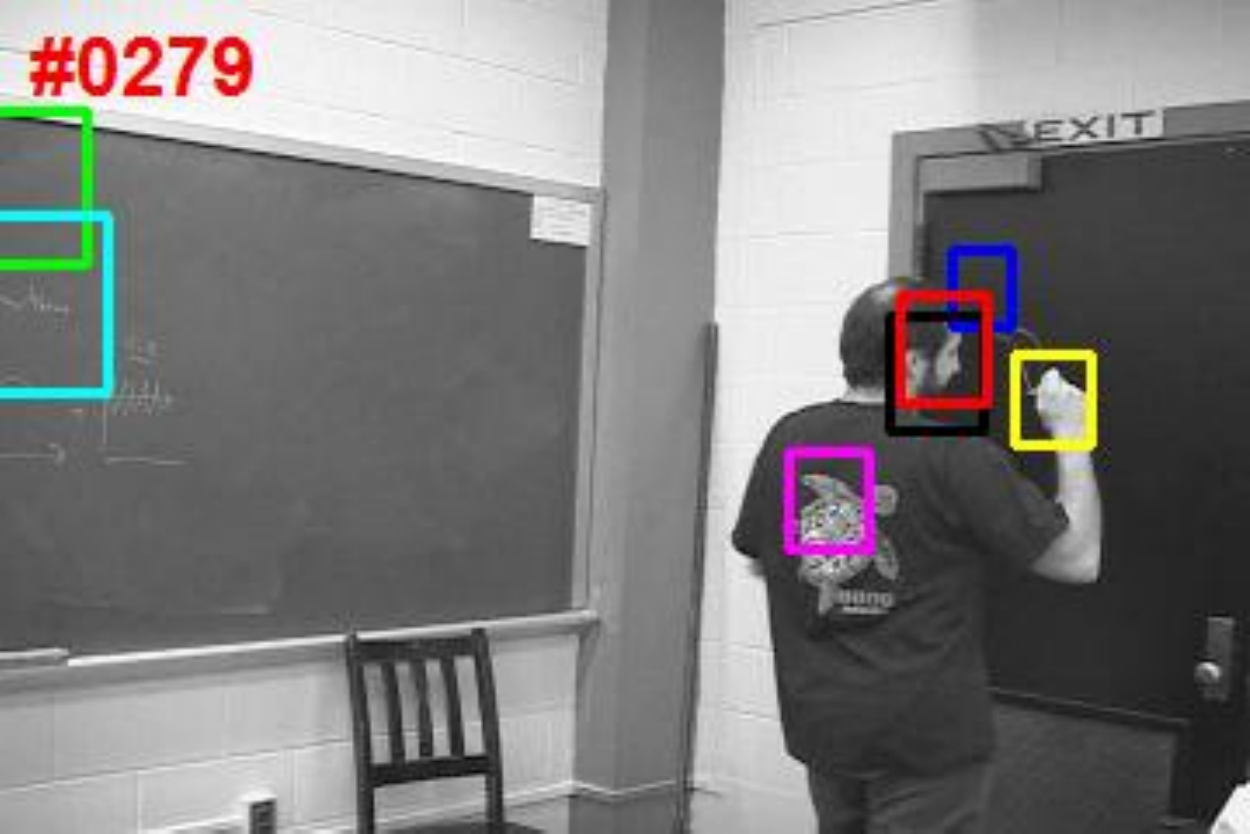}
&
\includegraphics[width=0.15\linewidth, height=0.08\linewidth]{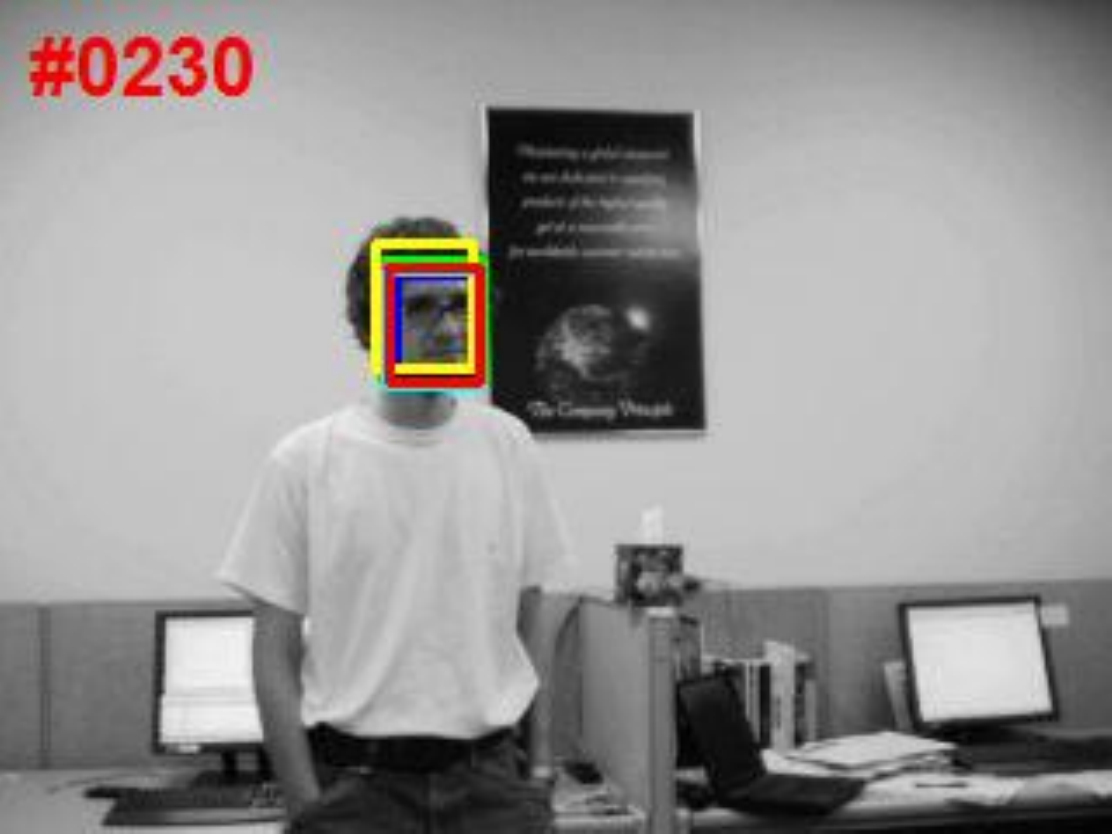}
&
\includegraphics[width=0.15\linewidth, height=0.08\linewidth]{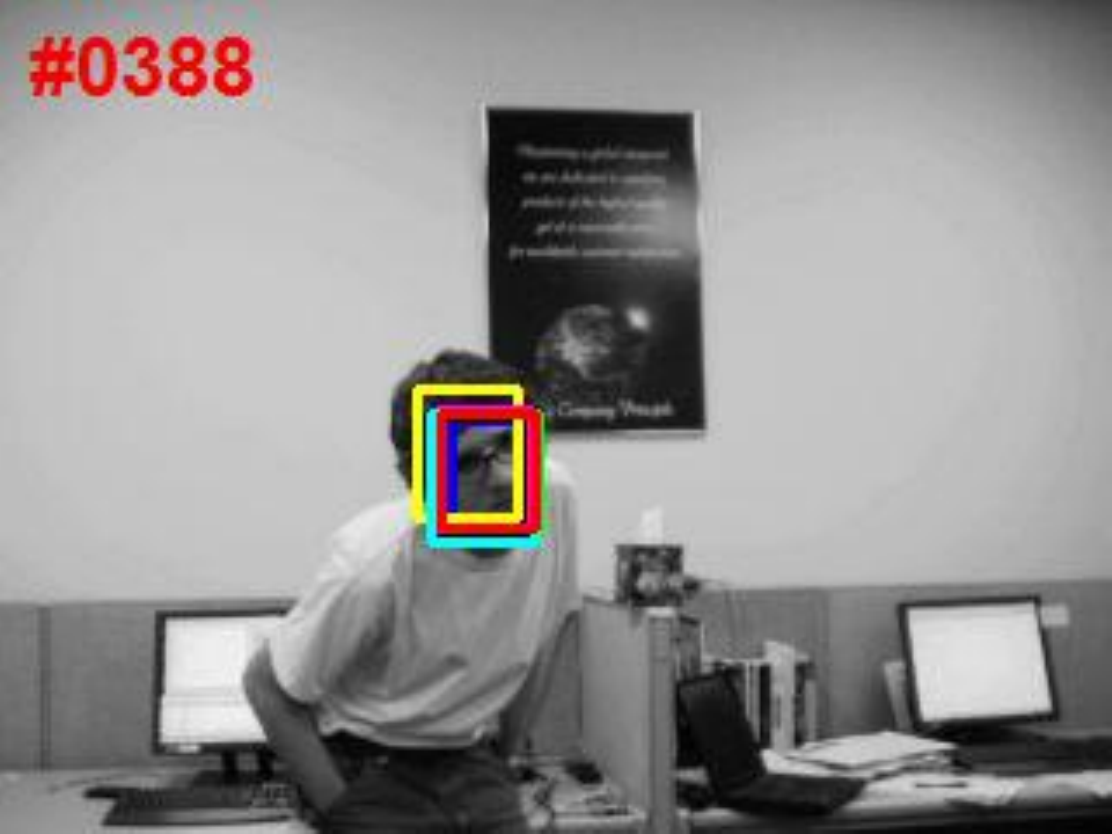}
\\
\end{tabular}

(i) \emph{doll}, \emph{freeman1} and \emph{david2}  with in-plane or out-of-plane rotation.
\includegraphics[width=0.70\linewidth,height=0.03\linewidth]{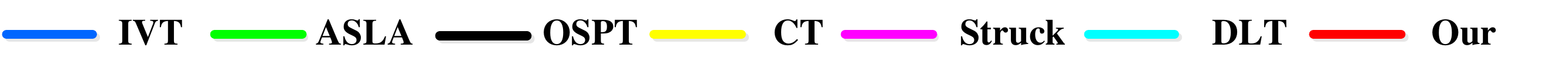}
\caption{Sample tracking results on twenty-five challenging sequences.}
\label{fig:trackingresults}
\end{figure*}

{\flushleft \bf{Occlusion:}}
We test six sequences (\emph{woman}, \emph{faceocc1}, \emph{faceocc2}, \emph{walking}, \emph{walking2}, \emph{caviar2}) characterizing in having either partial occlusion or heavy occlusion and the results are showen in Figure~\ref{fig:trackingresults} (a-b).
In the faceocc1 and walking2 sequences, the FRAG and the proposed tracker perform better because of the partial occlusion handling scheme they applied.
The FRAG method handle occlusion via fragments-based algorithm with an integral histogram tool.
As we employ the part-based incremental PCA method with an occlusion model, the appearance similarities of the occluded parts are penalized to be zeros, as a result, only parts that are not occluded account for the integral similarity between the candidate and its subspace.
For the woman and faceocc2 sequences, the target undergos pose variation under occlusion.
Since our deep classification network can adapt to appearance change through online fine-tuning, the proposed tracker can successfully track the target throughout the sequences.
{\flushleft \bf{Illumination Variation:}}
Figure~\ref{fig:trackingresults} (c-d) demonstrates the tracking results on challenging sequences \emph{car4}, \emph{carDark}, \emph{mhyang}, \emph{singer1} and \emph{singer2}.
The good performance of the proposed algorithm can be attributed to two aspects.
On one hand, in our generative model, the incremental subspace learning is robust to illumination change.
On the other hand, our deep discriminative model can extract discriminative features which can effectively separate the target from the background.
It is worth noticing that in the carDark sequence, the contrast is low between the foreground and the background.
The FRAG tracker fails because the part-based model cannot have the strong discriminative ability as the holistic representation.
However, our tracker can enjoy the advantage of both local representation and holistic representation by developing a collaborative model, which serve to enhance the robustness of our tracker.
{\flushleft \bf{Motion Blur:}}
We present tracking results on the sequences \emph{boy}, \emph{shaking}, \emph{fish} in Figure~\ref{fig:trackingresults} (e).
It is rather difficult to account for the blurred target appearance due to its fast motion.
As our deep discriminative model is trained to learn generic features for classification offline and adapted to the target appearance online, the classifier can exploit unchanged discriminative features extracted to better locate the target from the background.
{\flushleft \bf{Background Clutter:}}
The sequences, \emph{football}, \emph{mountainBike}, \emph{crossing}, \emph{board} and \emph{dudek}, which with respect to cluttered background, are reported in Figure~\ref{fig:trackingresults} (f-g).
The board sequence is challenging as the background is cluttered and the target experiences severe out-of-plane rotation.
Most holistic representation based trackers fail since they are not effective in handling objects with large appearance variations and the inevitable introduction of background pixels during update results in gradually drifting.
The FragTrack and the proposed algorithm are able to track the target better due to the use of local appearance models.
Also, as we use the discriminative and generic features extracted by the deep belief network for classification, the target can be differentiated from the cluttered background effectively.
{\flushleft \bf{Rotation and Deformation:}}
Figure~\ref{fig:trackingresults} (h-i) shows the tracking results in the sequences with in-plane or out-of-plane rotation, including \emph{girl}, \emph{bird}, \emph{dog1}, \emph{doll}, \emph{freeman1} and \emph{david2}.
The \emph{girl} sequence includes the challenge in terms of in-plane and out-of-plane rotation.
To overcome this challenge, both the positive training set and the subspace are updated online in order to capture the appearance change of the target.
Also, our discriminative model can be automatically retrained when it covers large appearance change, which serve to adapt our tracking scheme to the changing target.
\section{Conclusion}
In this paper, we propose a collaborative model consisting of both deep and shallow modules to account for major challenging factors in visual tracking, i e. appearance change and partial occlusion.
For the deep discriminative model, we use the DBN with a classification layer as output.
An auxiliary training set containing $100$ thousand training samples is built to train the DBN offline.
In the tracking process, the offline trained DBN is further fine-tuned to adapt to the specific target.
The high-level features learned in this way is characterized by the generic appearance of different objects as well as the specific appearance of the tracked target, thus can facilitate better handling appearance variation.
%
%
%
For the generative model, a block-based local model is proposed which exploits the classic principal component analysis and explicitly takes occlusion into consideration with an occlusion mask.
By measuring the appearance similarity between the candidate and the subspace block-wisely, our local model can alleviate the negative influence of the partial occlusion well.
Plenty experimental evaluations on challenging image sequences demonstrate that the proposed tracking algorithm performs favorably against the state-of-the-art methods.
\bibliographystyle{IEEEtran}
\bibliography{manuscript}
\end{document}